\documentclass{article}

\PassOptionsToPackage{square,numbers,sort}{natbib}
\usepackage{ml_arxiv}
\usepackage{ml_defs}

\usepackage{siunitx}

\usepackage{footmisc}  

\usepackage{graphicx}

\usepackage{xkcdcolors}
\usepackage[pdfusetitle]{hyperref}
\usepackage{bookmark}
\hypersetup{
  colorlinks=true,
  linkcolor=xkcdCrimson,
  urlcolor=xkcdBluish,
  citecolor=xkcdLightNavy,
}
\usepackage{xcolor}
\usepackage{subcaption}
\usepackage{multirow} 
\usepackage{booktabs}
\usepackage{wrapfig}

\usepackage{pifont} 
\newcommand{\xmark}{\ding{55}}
\newcommand{\cmark}{\ding{51}}

\newcommand{\st}[2][]{_{\text{#2}{#1}}} 
\newcommand{\spt}[1]{^{\text{#1}}}
\newcommand{\bvec}[1]{{\boldsymbol{#1}}}
\renewcommand*{\thefootnote}{\fnsymbol{footnote}}
\newcommand\nnfootnote[1]{%
  \begin{NoHyper}
  \renewcommand\thefootnote{}\footnote{#1}%
  \addtocounter{footnote}{-1}%
  \end{NoHyper}
}

\title{NeuralDEM -- Real-time Simulation of \\ Industrial Particulate Flows}
\author{
Benedikt Alkin$^{\dagger,*,1,2}$ \quad
Tobias Kronlachner$^{\dagger,*,1,3}$ \quad
Samuele Papa$^{\dagger,\ddagger, 1,4,5}$ \quad \\ \vspace{0.2cm} \bf
Stefan Pirker$^{3}$ \quad 
Thomas Lichtenegger$^{1,3}$ \quad 
Johannes Brandstetter$^{@, 1,2}$\\ \\
$~^{1}$Emmi AI GmbH, Linz, Austria\\
$~^{2}$ELLIS Unit Linz, Institute for Machine Learning, JKU Linz, Austria\\
$~^{3}$Department of Particulate Flow Modelling, JKU Linz, Austria \\
$~^{4}$University of Amsterdam, Amsterdam, Netherlands \\
$~^{5}$The Netherlands Cancer Institute, Amsterdam, Netherlands \\
}
\date{March 2024}

\begin{document}
\maketitle

\begin{abstract}
\nnfootnote{This work is a preprint.}
\nnfootnote{$~^{\dagger}$ core contributor, $~^{*}$ equal contribution, $~^{\ddagger}$ work done during internship.}
\nnfootnote{@ Correspondence to: \texttt{brandstetter@ml.jku.at} , \ \ \texttt{johannes@emmi.ai}}\\
Advancements in computing power have made it possible to numerically simulate large-scale fluid-mechanical and/or particulate systems, many of which are integral to core industrial processes. Among the different numerical methods available, the discrete element method (DEM) provides one of the most accurate representations of a wide range of physical systems involving granular and discontinuous materials. Consequently, DEM has become a widely accepted approach for tackling engineering problems connected to granular flows and powder mechanics. Additionally, DEM can be integrated with grid-based computational fluid dynamics (CFD) methods, enabling the simulation of chemical processes taking place, e.g.,~in fluidized beds.
However, DEM is computationally intensive because of the intrinsic multiscale nature of particulate systems, restricting either the duration of simulations or the number of particles that can be simulated. Moreover, the non-trivial relationship between microscopic DEM and macroscopic material parameters necessitates extensive calibration procedures.
Towards this end, NeuralDEM presents a first end-to-end approach to replace slow and computationally demanding numerical DEM routines with fast, adaptable deep learning surrogates. NeuralDEM is capable of picturing long-term transport processes across different regimes using macroscopic observables without any reference to microscopic model parameters.
First, NeuralDEM treats the Lagrangian discretization of DEM as an underlying continuous field, while simultaneously modeling macroscopic behavior directly as additional auxiliary fields.
Second, NeuralDEM introduces multi-branch neural operators scalable to real-time modeling of industrially-sized scenarios -- from slow and pseudo-steady to fast and transient. Such scenarios have previously posed insurmountable challenges for deep learning models.
Notably, our largest NeuralDEM model is able to faithfully model coupled CFD-DEM fluidized bed reactors of $160$k CFD cells and $500$k DEM particles for trajectories of \texorpdfstring{$\SI{28}{s}$}, which amounts to $2800$ machine learning timesteps.   
NeuralDEM will open many new doors to advanced engineering and much faster process cycles. \\
Project page: \url{https://emmi-ai.github.io/NeuralDEM/}.

\end{abstract}

\newpage
\tableofcontents

\clearpage

\section{Introduction}

\begin{figure}
    \centering
    \includegraphics[width=0.99\linewidth ,trim={100cm 100cm 110cm 40cm},clip]{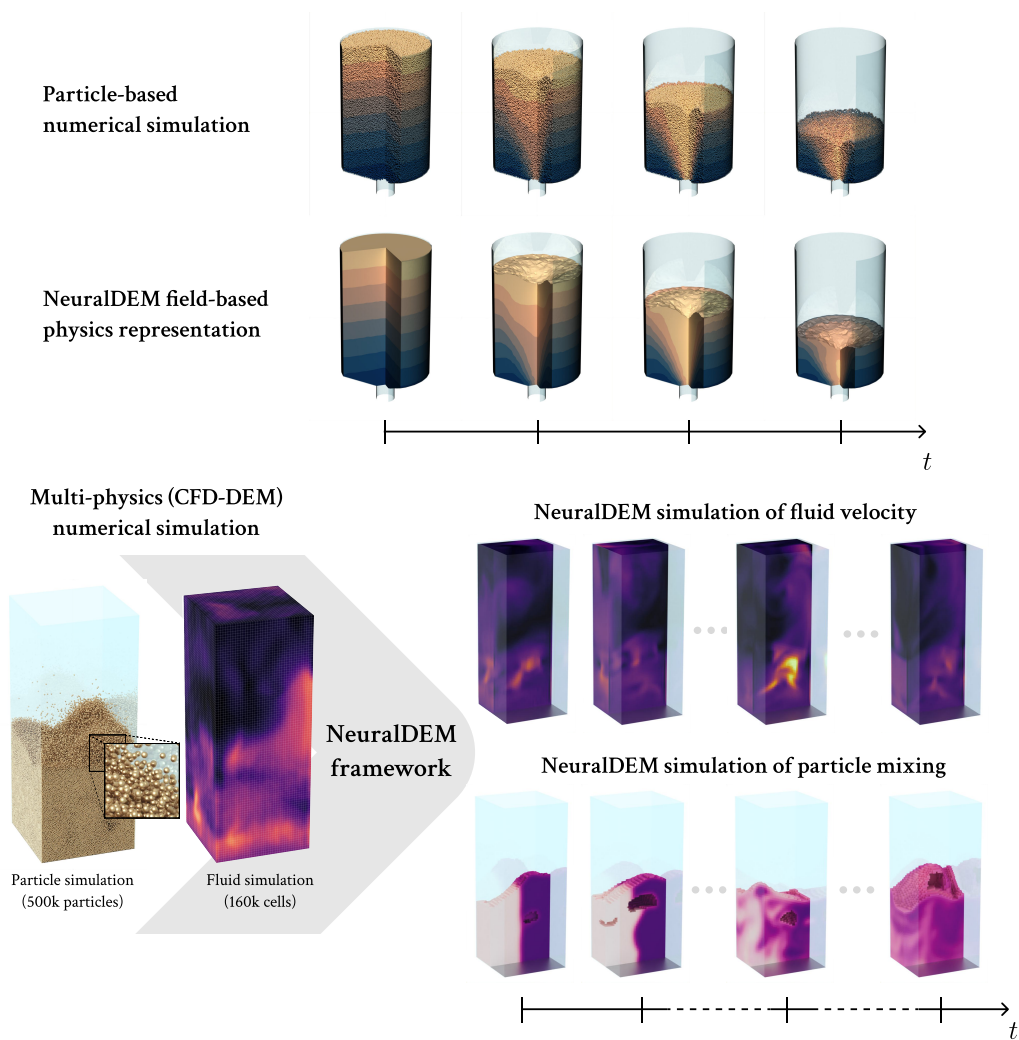}
    \caption{\textbf{NeuralDEM} presents an end-to-end approach to replace discrete element method (DEM) routines and coupled multiphysics simulations with deep learning surrogates. \textit{Top}: Hopper simulations. NeuralDEM treats inputs and outputs as continuous fields, while modeling macroscopic behavior directly as additional auxiliary fields. \textit{Bottom}: Fluidized bed reactors. NeuralDEM is built to model complex multiphysics simulations, i.e., scenarios which necessitate the interaction of DEM and computational fluid dynamics (CFD). For fluidized bed reactors, air enters the domain from the bottom plane (CFD problem) and pushes the particles up (DEM problem). Both parts and the interaction thereof is modeled via the new multi-branch neural operator approach of NeuralDEM. 
    }
    \label{fig:enter-label}
\end{figure}

In recent years, real-time numerical simulations \cite{Chen2020Realtime, Zeeshan2024Realtime,  IEEEPower2015Realtime,Yan2023Realtime, Javaid2023Realtime}, have emerged as new modeling paradigm, enabling immediate analysis and decision-making based on live data and conditions. Unlike traditional simulations, which may take hours or even days to run, real-time simulations provide instantaneous feedback, allowing users to interact with and adjust parameters on the fly. Moreover, in engineering, fast simulations are driving the design of safer and more efficient structures and machines by accurately predicting their behavior under different conditions, thereby allowing extensive scans of vast parameter spaces, and eliminating the need for expensive physical prototypes.

Especially for numerically expensive problems as found, e.g.,\ in computational particle mechanics and/or fluid dynamics (CFD), selecting the appropriate numerical simulation tool requires weighing accuracy against speed. Among the different numerical methods available, the discrete element method (DEM)~\cite{Cundall1979} provides one of the most accurate representations of a wide range of physical systems involving particulate matter, by tracking and computing the behavior of each grain. Consequently, DEM has become a widely accepted approach for tackling engineering problems connected to disperse and discontinuous materials, particularly in granular flows and powder mechanics. Typical target areas comprise mining and mineral processing~\cite{andre2020,mittal2023,zhang2024}, steelmaking~\citep{aminnia2023,amani2024,lichtenegger2024}, pharmaceutics~\cite{benque2024,giannis2021,grohn2020}, agriculture and food processing~\citep{wang2022,yan2022,zhao2021}, and
additive manufacturing and powder bed fusion~\cite{chen2022,nasato2020,zhang2022}.

However, the inherent multiscale nature of particulate systems makes DEM computationally costly in several well-known regards. (i) Large-scale granular flows consist of a huge number of particles, each interacting with the surrounding ones. For every grain, its equation of motion (EOM) needs to be solved in a coupled fashion. 
Current DEM studies dealing with process-relevant problems use many 100k~\cite{Demunck2023,Li2023LargeDEM} or even a few million~\cite{diez2019fb} particles, while, e.g., an industrial shaft furnace or fluidized bed reactor contains several orders of magnitude more. 
Although GPU-based codes~\cite{Govender2015,Gan2020} can handle more grains than those on CPUs and are significantly faster for the same problem size, they are still far away from convenient runtimes for demanding industrial processes, especially if a fluid phase is involved~\cite{Lu2022,Yu2024}.
(ii) The high material stiffness of solid particles severely limits the numerical timestep that can be used in the solution procedure of the EOMs. Often, its value is in the range of microseconds, whereas process-relevant durations may be minutes or hours. (iii) There is no straightforward relationship between the microscopic DEM parameters and macroscopically observed behavior. Instead, optimization techniques need to be used to find a set of DEM parameters that reproduces certain characterization measurements (e.g., angle of repose and shear cell measurements). Such a calibration routine~\cite{coetzee2017} needs to be performed for any type of material before the actual simulation of interest can be approached. Any change in the material properties (e.g., due to different size distribution, degradation, moisture uptake) requires a new calibration.

Issue (i) is usually mitigated by employing coarse-graining techniques that replace many small particles with a large parcel~\cite{bierwisch2009,sakai2009}. If the interaction parameters of these parcels are chosen appropriately -- either using scaling rules or a calibration routine -- the accuracy impairments compared to the fine-grained ground truth are often acceptable. However, the limitation of small timesteps and the need for parameter calibration persist and make DEM slow and sometimes too cumbersome for a quick application within engineering workflows.

We present \textbf{NeuralDEM}, the first end-to-end deep learning alternative for modeling industrial processes. NeuralDEM introduces \textit{multi-branch neural operators} inspired by multi-modal diffusion transformers (MMDiT)~\citep{esser2024stablediffusion3} and is scalable to real-time modeling of industrially-sized relevant scenarios. 
In NeuralDEM, we introduce two key components. The first is modeling the Lagrangian discretization given by DEM simulations directly from a compressed Eulerian perspective, i.e., a \textit{field-based point of view}. Our method -- through model conditioning -- can generalize across macroscopic quantities such as inflow velocity or internal friction angle, addressing issue (iii). Interestingly, this new modeling point of view aligns well with recent findings that a DEM-simulated system's effective degrees of freedom are orders of magnitude less than the microscopic degrees of freedom~\cite{Lichtenegger2018b}. This has the benefit of allowing direct modeling of macroscopic processes, e.g., mixing or transport processes, via additional auxiliary continuous fields. The second key component is \textit{multi-physics modeling} via repeated interactions between physics phases.
Multi-physics is prevalent when modeling the interaction of fluid dynamics and particulate systems in fluidized bed simulations. NeuralDEM, additionally, learns to stably simulate the system using longer timesteps, addressing issue (ii). Together, these properties allow tackling all common issues with DEM, making systems with large numbers of particles computationally feasible, allowing the use of longer timesteps, and enabling direct conditioning on macroscopic characterization measurements, without the need for fine-tuning the microscopic DEM parameters.

We test NeuralDEM and demonstrate its capability to picture various transport processes, e.g., mass, species, residence time, or mixing, in two scenarios, both of which are simulated with several 100k DEM particles: (i) slow and pseudo-steady hoppers with varying hopper angles, internal friction angles and flow regimes, and (ii) fast and transient fluidized bed reactors with varying inflow velocities. In all scenarios, we investigate the correct physics modeling of, e.g., outflow ratios, residence time, mixing ratio, and others. We observe that NeuralDEM generalizes to unseen parameter choices, and, that NeuralDEM produces faithful physics simulations for long time horizons. 
Most notably, our largest NeuralDEM model is able to physically-correctly model coupled CFD-DEM fluidized bed reactors of $160$k CFD cells and $500$k DEM particles for trajectories of $\SI{28}{s}$, which amounts to $2800$ machine learning timesteps \footnote{See \url{https://emmi-ai.github.io/NeuralDEM/} for qualitative comparisons of full sequences.}.
These findings will open many new doors to advanced engineering and much faster process cycles.

\section{Background}

In this section, we introduce the relevant core phenomena of DEM. A more in-depth explanation can be found, e.g., in the review article form ~\citet{blais2019} or the textbook from ~\citet{norouzi2016}. Further, we introduce neural operators and discuss their applicability to model particulate systems.

\subsection{Discrete element method}

\begin{figure}[h]
    \centering
    \includegraphics[width=0.5\linewidth]{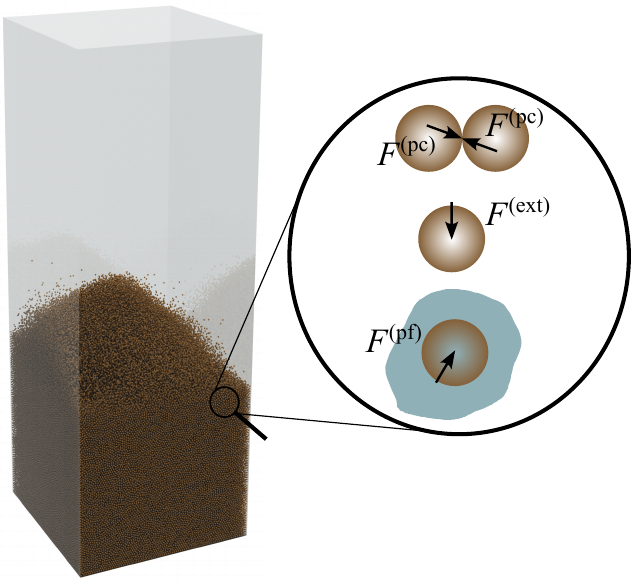}
    \caption{\textit{Discrete element method}. The force on a particle consists of particle-particle contacts $\bvec{F}_i\spt{(pc)}$, the external force $\bvec{F}_i\spt{(ext)}$, and the interaction with a surrounding fluid phase $\bvec{F}_i\spt{(pf)}$.}
    \label{fig:DEM_sketch}
\end{figure}
In a system of solid particles with masses $m_i$, radii $r_i$, positions $\bvec{r}_i$, and velocities $\bvec{v}_i$, each of them has to obey Newton's second law
\begin{equation}
    \frac{\text{d}}{\text{d}t}m_i\bvec{v}_i = \bvec{F}_i\spt{(ext)} + \bvec{F}_i\spt{(pc)} + \bvec{F}_i\spt{(pf)}.
    \label{eq:demeom}
\end{equation}
Particle $i$ experiences forces of external origin, most important gravity $\bvec{F}_i\spt{(ext)} \approx m_i \bvec{g}$, contact forces with the nearby grains and walls $\bvec{F}_i\spt{(pc)} = \sum_{j\neq i} \bvec{F}_{i,j}$, and the influence of a surrounding fluid phase $\bvec{F}_i\spt{(pf)}$ if present and relevant. 

The contact force between solid particles $i$ and $j$ is commonly approximated with spring-dashpot models for both the normal $\bvec{F}\spt{(n)}_{i,j}$ and tangential component $\bvec{F}\spt{(t)}_{i,j}$,
\begin{align}
    &\bvec{F}\spt{(n)}_{i,j} = -k\spt{(n)}\delta\spt{(n)}_{i,j}\bvec{n}_{i,j} + \gamma\spt{(n)}\bvec{v}\spt{(n)}_{i,j} \\
    &\bvec{F}\spt{(t)}_{i,j} = \text{min}\Big[-k\spt{(t)}\delta\spt{(t)}_{i,j}\bvec{t}_{i,j} + \gamma\spt{(t)}\bvec{v}\spt{(t)}_{i,j},\mu |\bvec{F}\spt{(n)}_{i,j}|\bvec{t}_{i,j} \Big],
\end{align}
where the tangential force is limited by Coulomb's friction law. Material properties enter these expressions in terms of the spring stiffnesses $k\spt{(n,t)}$, damping coefficients $\gamma\spt{(n,t)}$ and sliding friction $\mu$. $k\spt{(n,t)}$ give rise to the reaction against normal and tangential overlap $\delta\spt{(n,t)}_{i,j}$ between two grains in the respective directions $\bvec{n}_{i,j}$ and $\bvec{t}_{i,j}$, and $\gamma\spt{(n,t)}$ account for viscous dissipation caused by the normal and tangential relative velocities $\bvec{v}\spt{(n,t)}_{i,j}$ during contact.
While simple geometric shapes such as perfect spheres allow computing values for the material parameters from measurable properties like Young's and shear modulus~\cite{johnson1985}, actual, imperfect grains necessitate calibration towards characterization experiments~\cite{coetzee2017}.

The magnitude of the numerical timestep to solve Equation~\eqref{eq:demeom} is limited by the requirement to properly resolve contacts between grains. More specifically, the timestep needs to be significantly smaller than the duration of contact of colliding particles (``Hertz time'') and the time it takes density waves generated upon impact to travel over the grain surface (``Rayleigh time''). For stiff materials, this often amounts to steps in the range of microseconds.

\subsubsection{Coupled particle-fluid simulations}

Particles will also experience a force from a surrounding fluid phase. While it may be neglected if no significant relative velocities occur, it can be a crucial factor for particle dynamics otherwise.
The dominant contributions are usually caused by gradients of the pressure and by the drag force, i.e.,\ the resistance against relative velocity between fluid and grain, so that
\begin{equation}
    \bvec{F}_i\spt{(pf)} \approx -V_i \nabla p + \beta \big(\bvec{u}\st{f} - \bvec{v}_i \big).
    \label{eq:Fpf}
\end{equation}
The drag coefficient $\beta$ depends on the particle size and the local flow conditions. A multitude of empirical correlations can be found in the literature to take into account the impact of Reynolds number, particle volume fraction $\alpha\st{p}$, size distribution, etc.~\cite{kieckhefen2020}.

The fluid velocity itself is governed by the filtered Navier-Stokes equations~\cite{anderson1967}
\begin{align}
    & \frac{\partial }{\partial t} \alpha\st{f} + \nabla\cdot \alpha\st{f}\bvec{u}\st{f} = 0 \label{eq:fluideom0} \\
    & \frac{\partial }{\partial t} \alpha\st{f}\bvec{u}\st{f} + \nabla\cdot \alpha\st{f}\bvec{u}\st{f}\bvec{u}\st{f} = \nabla \cdot\bvec{\sigma}\st{f} - f\spt{(pf)}\label{eq:fluideom1}
\end{align}
which differ from their single-phase counterpart in two regards. The presence of particles reduces the locally available volume to a fraction $\alpha\st{f} = 1 - \alpha\st{p}$, and the density of force Equation~\eqref{eq:Fpf} exerted by the fluid on the particles is felt by the fluid in opposite direction because of Newton's third law.
The coupled solution of the CFD Equations~\eqref{eq:fluideom0} and \eqref{eq:fluideom1} and the DEM Equation~\eqref{eq:demeom} gives rise to CFD-DEM simulations.

For a proper definition of the field quantities $\alpha\st{p}$ and $f\spt{(pf)}$, Lagrangian particle information needs to be mapped onto Eulerian fields. To this end, a filter function $g_l(r)$, e.g., a Gaussian with width $l$, is employed in terms of
\begin{align}
    & \alpha\st{p}(\bvec{r}) \equiv \sum_i g_l(|\bvec{r}-\bvec{r}_i|)V_i \\
    & f\spt{(pf)}(\bvec{r}) \equiv \frac{\sum_i g_l(|\bvec{r}-\bvec{r}_i|)V_i\bvec{F}_i\spt{(pf)}}{\sum_i g_l(|\bvec{r}-\bvec{r}_i|)V_i }.\label{eq:LatoEu}
\end{align}
An analogous definition as Equation~\eqref{eq:LatoEu} can be invoked to define the \textit{spatial field distribution} of any particle property. As a matter of fact, the target quantities of most particle simulations are not necessarily connected to single-particle properties located exactly at the positions of each grain. Instead, one might be interested in the spatial distribution of, e.g., particle volume fraction, residence time or temperature. Two strategies are available to obtain these fields: (i) One carries out a DEM simulation and postprocesses particle data according to Equation~\eqref{eq:LatoEu}. The trajectory and properties of each grain are only needed as an intermediate step for the DEM simulation.
(ii) One can try to directly formulate particle EOMs in an Eulerian fashion by filtering the Lagrangian ones and solving them disregarding discrete properties. As demonstrated by the two-fluid model~\cite{gidaspow1994}, this can significantly reduce computational costs but can come with a serious degree of uncertainty~\cite{chen2014} because not all particle properties lend themselves to a straight-forward formulation in terms of fields.

Even if the resulting inaccuracies are acceptable, such simulations are still cumbersome because of the restriction to small timesteps (which is also present in an Eulerian formulation) and the lack of a direct relationship between macroscopic behavior and microscopic parameters. An attractive solution to this predicament might be offered by neural operators that can be trained with detailed particle data to predict any underlying field quantities in a highly efficient way.

\subsection{Neural operators learning for scientific and engineering applications}

In recent years, deep learning tools have been extensively integrated into scientific modeling, and have resulted in breakthroughs in, e.g., protein folding~\citep{jumper2021highly,abramson2024accurate}, material discovery~\citep{Batzner:22,Batatia:22,Merchant:23,Zeni:23}, or weather modeling~\citep{pathak2022fourcastnet,bi2023accurate,lam2023learning,bodnar2024aurora,nguyen2023climax}.    
Driven by applications in CFD~\citep{Vinuesa:22, Guo:16, Li:20, Kochkov:21, Gupta:22}, deep neural network based surrogates, most importantly neural operators~\citep{Li:20,Lu:21,Kovachki:21},
have emerged as a computationally efficient alternative~\citep{Zhang:23}. In addition to computational efficiency, neural operators offer the potential to introduce generalization capabilities across phenomena, as well as generalization across characteristics such as boundary conditions or coefficients~\citep{McCabe:23,Herde:24}.

\textit{Neural operators}~\citep{Lu:21, Li:20graph, Li:20,Kovachki:21} are formulated with the aim of learning a mapping between function spaces, enabling outputs that remain consistent across varying input sampling resolutions.
Following the framework of~\citet{Kovachki:21}, we assume $\mathcal{U}, \mathcal{V}$ to be Banach spaces of functions defined on compact domains $\cX \subset \dR^{d_x}$ or $\cY \subset \dR^{d_y}$, respectively, which map into $\dR^{d_u}$ or $\dR^{d_v}$.
A neural operator $\hat{\gG} : \mathcal{U} \rightarrow \mathcal{V}$ approximates the ground truth operator $\gG: \mathcal{U} \rightarrow \mathcal{V}$. 

\begin{figure}[h]
    \centering
    \includegraphics[width=\linewidth,trim={0cm 0cm 0cm 0.5cm},clip]{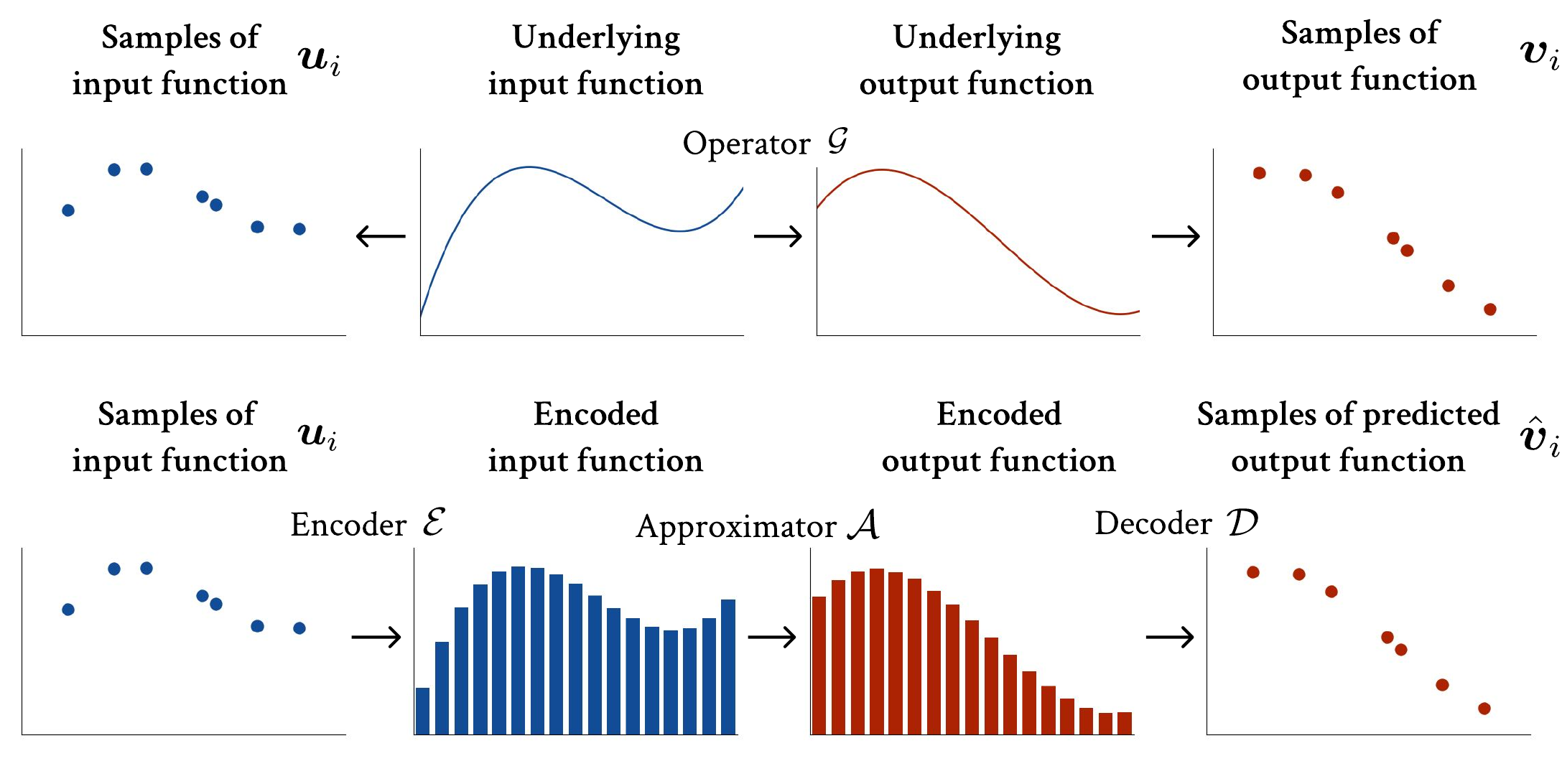}
    \caption{Neural operator learning. Neural operators aim to learn a mapping between function spaces, enabling outputs that remain consistent across varying input sampling resolutions. The neural operator $\hat{\cG}$ approximates the ground truth operator $\cG$ with three maps, composing encoder $\cE$, approximator $\cA$, and decoder $\cD$. The approximation of $\hat{\cG}$ is ideally independent from the number of sampled input points, and approximates the output function for an arbitrary number of points. 
    }
    \label{fig:NO_sketch}
\end{figure}

When training a neural operator $\hat{\cG}$, a widely adopted approach is to construct a dataset of $N$ discrete data pairs $(\Bu_{i,j}, \Bv_{i,j'})$, $i=1,\ldots,N$, which correspond to $\Bu_i$ and $\Bv_i$ evaluated at spatial locations $j=1,\ldots,K$ and $j'=1,\ldots,K'$, respectively. Note that $K$ and $K'$ can, but need not be equal, and can vary for different $i$, which we omit for notational simplicity. 
Figure~\ref{fig:NO_sketch} (first row) shows the operator learning problem, i.e., the mapping of an input function $\Bu_i$ to an output function $\Bv_i$ via an operator $\cG$. The functions are given via $K$ and $K'$ discretized input and output points, respectively. 
On this dataset, $\hat{\cG}$ is trained to map $\Bu_{i,j}$ to $\Bv_{i,j'}$ via supervised learning, as sketched in Figure~\ref{fig:NO_sketch} (bottom row), where $\hat{\gG}$ is composed of three maps~\citep{Seidman:22,alkin2024upt}:
$\hat{\gG} \coloneqq \cD \circ \cA \circ \ \cE
$, comprising the encoder $\cE$, the approximator $\cA$, 
and the decoder $\cD$. First, the encoder $\cE$ transforms the discrete function samples $\Bu_{i,j}$ to a latent representation of the input function. Then, the approximator $\cA$ 
maps the latent representation to a representation of the output function. Lastly, the decoder evaluates the output function at spatial locations $j'$. The neural network $\hat{\gG}$ is then trained via gradient descent, using the gradient of, e.g., a mean squared error loss in the discretized space $\mathcal{L}_i = \frac{1}{K'} \sum_{j'}\lVert \hat{\Bv}_{i,j'} - \Bv_{i,j'}\rVert_2^2$, where $\rVert_2$ is the Euclidean norm.

For temporally evolving systems, we assume the Banach spaces of functions to be equal, i.e., $\mathcal{U}=\mathcal{V}$, where models are trained on next-step prediction $\Bu^t_{i,j} \rightarrow \Bv^{t'}_{i,j'}$. To fully evolve a system after training, the model is then applied repeatedly in autoregressive fashion, i.e.,\ every prediction serves as new input.

\subsubsection{Discretization convergence}
Neural operators are well suited to describe the evolution and interaction of physical quantities over space and time, i.e., continuously changing fields. 
Most notably, since the solutions are continuous fields, the mapping should neither depend on $K$, the number of input locations, nor on $K'$, the number of decoded output locations. The property that neural network outputs remain consistent across different input sampling resolutions is referred to as \textit{discretization convergence}~\citep{Lu:21,Li:20,Kovachki:21}.
Neural operators are proven to be discretization convergent in the limit of mesh refinement~\citep{Kovachki:21}, meaning they converge to a continuum operator in the limit as the discretization is refined. In theory and if properly designed, neural operators can be evaluated at any data discretization.
Nevertheless, there is a minimum threshold for the number of points needed for accurate representation. However, strong evidence indicates that neural networks can capture physical phenomena effectively without requiring the same level of fine-grained discretization as traditional numerical methods~\citep{Kochkov:21}.

Neural operator architectures need to ensure discretization convergence of their components. When encoding the discretized input function, popular choices to preserve discretization convergence are graph neural operators~\citep{li2020gno,li2023gino}, transformers~\cite{vaswani2017transformer,hao23gnot,wu2024transsolver} or combinations thereof~\cite{alkin2024upt}. 
For decoding, recent works~\citep{Wang:24} have shown that $\cD$ can be considered as neural field~\citep{Sitzmann:20,Mildenhall:21,Xie:22}, which allows for point-wise evaluation at the output grid or output mesh~\citep{alkin2024upt,Wang:24,Knigge:24,kofinas2024latent}.

\subsubsection{Deep learning for particulate systems}
While most state-of-the-art neural operator approaches are predominantly designed for geometrically simple domains with regular grids, neural operator formulations for particle- or mesh-based dynamics remain limited. In such cases, graph neural networks (GNNs)~\citep{Scarselli:08, Kipf:17, Brandstetter:22} with graph-based latent space representations are a prevalent approach to build neural surrogates. Often, predicted node accelerations are numerically integrated to simulate the time evolution of multi-particle systems~\citep{Sanchez:20, Pfaff:20, Mayr:23, Toshev:23,toshev2024neural}.
For the modeling of granular dynamics, \citet{li2023prediction} predict contact forces when inputting microstructures of grain packings. Similarly, \citet{cheng2022estimation} estimate contact forces in compressed granular assemblies. \citet{Mayr:23} introduce Boundary-GNNs to model granular flows through hoppers, rotating drums, and mixers. All these models are limited by the number of particles, and operate on 10k -- 20k particles at most, although often much less. 

GNNs inherently possess a strong inductive bias for Lagrangian dynamics, which, however, presents a significant downside since the number of nodes, and thus the computational complexity grows with the number of Lagrangian particles. Thus, computational complexity gets quickly infeasible for an increasing number of particles~\citep{alkin2024upt,Musaelian:23}. However, motivated by recent successes in latent space generative modeling~\citep{rombach2022latentdiffusion,esser2024stablediffusion3}, latent space modeling has emerged as a new modeling paradigm in neural operator learning~\citep{alkin2024upt,hemmasian2024pretraining,wang2024latent,zhou2024text2pde}. In this work, we follow the argumentation of~\citet{alkin2024upt}, i.e., neural operators with large model complexity are powerful enough to capture inherent field characteristics when applied to Lagrangian or multiphysics simulations. For DEM simulations, we argue that those field characteristics need not be explicitly present in the training data, rather, they might emerge from the bulk behavior of the particulate systems.

\section{NeuralDEM}
NeuralDEM presents the first end-to-end solution for replacing computationally intensive numerical DEM routines and coupled CFD-DEM simulations with fast and flexible deep learning surrogates. 
NeuralDEM introduces two conceptually novel modeling paradigms:
\begin{enumerate}
\item \emph{Physics representation}: We model the Lagrangian discretization of DEM as an underlying
continuous field, while simultaneously modeling macroscopic behavior directly
as additional auxiliary fields. NeuralDEM encodes different physics inputs which are representative for DEM dynamics and/or multi-physics scenarios. Examples are particle displacement, particle mixing, solid fraction, or particle transport. 
\item \emph{Multi-branch neural operators}: 
We introduce multi-branch neural operators scalable to real-time modeling of industrial-size scenarios. Multi-branch neural operators build on the flexible and scalable ``Universal Physics Transformer''~\citep{alkin2024upt} framework by enhancing encoder, decoder, and approximator components using multi-branch transformers to allow for modeling of multi-physics systems.
The system quantities fundamental to predicting the evolution of the state in time are modeled in the main-branches, where they are tightly coupled.
Additionally, auxiliary \textit{off-branches} can be added to directly model macroscopic quantities by retrieving information from the main-branch state and further refining the prediction using relevant inputs.
\end{enumerate}

\begin{figure}[!htb]
    \centering
    \includegraphics[width=0.9\linewidth,trim={9cm 0cm 17cm 0cm},clip]{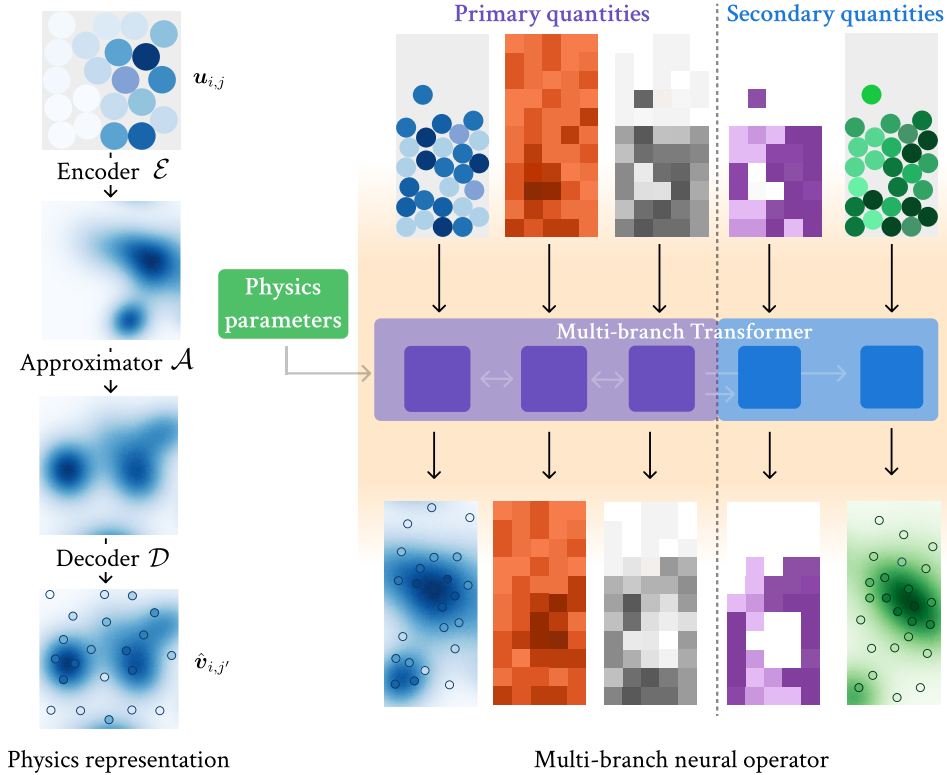}%
    \caption{In our \textit{physics representation} we model the Lagrangian discretization of DEM as an assumed underlying continuous field. The approximator maps the encoded representation to one that can be decoded at any specified spatial location $j'$. The \textit{multi-branch neural operator} is a family of deep learning architectures that processes multi-physics quantities and can distinguish between primary quantities, used to model the core physics in the main-branches, and secondary quantities which are used to predict additional desired quantities in the off-branches, both modeled as fields. The quantities come, e.g., from DEM simulations with coupled particles and fluid, which the architecture handles using specialized encoders and decoders. All modules processing the primary quantities influence each other. In contrast, those that process secondary quantities are independent, and use the tokens from the primary branch as additional information but cannot affect them.}
    \label{fig:approach_overview}
\end{figure}

\subsection{Physics representation}

As common when training deep learning surrogates, we train on orders of magnitude coarser timescales than what a classical solver requires to be stable and accurate. For the numerical experiments in this paper, the timescale relation is at least $1000 \Delta t_\text{DEM} = \Delta t_\text{ML}$. Additionally, for learning the dynamics of particle movement, we use the \textit{particle displacement}, which is defined as the difference between the position $\Br_i \in \mathbb{R}^3$ of particle $i$ at timestep  $t_\text{ML}$ and the position of the same particle at timestep $t_\text{ML} + \Delta t_\text{ML}$. Finally, we use the term \textit{transport} to denote the particle movement integrated over multiple timesteps $\Delta t_\text{ML}$.

NeuralDEM models the Lagrangian discretization of DEM as a continuous field in a compressed latent space, leveraging the insight that the effective degrees of freedom of physical systems is often much smaller than its input dimensionality~\citep{Lichtenegger2018b}. Therefore, we assume that there exists some underlying field that describes the particle displacements in a DEM simulation and learn this underlying field over the whole domain instead of a displacement per particle. However, particle displacements can fluctuate depending on their exact position within the bulk of the material. Such fine-grained details are lost when going to a field-based representation which smoothes out these variations.
This makes field-based models unable to move particles accurately around in space, which would be required to get macroscopic insights into the simulation dynamics.

To circumvent this issue, we introduce additional auxiliary fields that model the macroscopic insights directly instead of calculating them in post-processing from the particle locations. For example, by modeling the accumulated particle movement over a long period of time via a ``transport'' field, we can learn macroscopic properties directly instead of integrating short-term movements which would require precise prediction of the fluctuations thereof. This is visualized in Figure~\ref{fig:approach_overview}.

Even over such a large timestep, the evolution of a flow and its properties at each point is mainly determined by the field values in a nearby, bounded subdomain which grows with the the step size, and hardly influenced by very distant points~\cite{lichtenegger2024b}. This behavior can be resembled by the attention mechanism of transformer networks~\citep{vaswani2017transformer}.

\subsection{Multi-branch neural operators} 

\paragraph{Emerging bulk behavior of classical solvers as motivation.}
Classical solvers can create full simulations via precisely updating microscopic properties such as the particle positions at extremely high time resolution, with optional coupling to, e.g., a fluid phase, which is updated with similar precision and timescale. Similarly, NeuralDEM aims to extract the physical dynamics and simulation state updates from the microscopic properties also used in classical solvers, which we call ``main-phase(s)'' where each main-phase is processed by one main-branch transformer in our model. Using an example of a particle-fluid coupled simulation, one main-branch predicts particle displacements, while a second main-branch predicts fluid velocities and pressures. All main-branches are tightly coupled via frequent information exchange during the model forward pass.

\paragraph{Microscopic inaccuracies of neural operators.}
While the model is trained using microscopic properties, neural operators are not able to predict microscopic properties, such as the particle displacements, accurately enough because neural operators are not as precise as classical solvers and operate on much coarser time resolution. It is therefore not feasible to exclusively rely on an accurate prediction of these microscopic properties in the main-branches. Creating simulations by moving initial particle positions according to the predicted displacements would quickly result in unphysical states (e.g., overlapping particles) and becomes inaccurate.

\paragraph{Macroscopic modeling via auxiliary fields.}
An important observation regarding the systems we model is that the insights that a classical solver can provide into the physical dynamics are rarely on a microscopic level and more often on a macroscopic level, where the macroscopic properties are extracted from the microscopic results of the classical solver. Motivated by this intuition, we introduce additional off-branches, which are trained to model macroscopic processes such as particle mixing or particle transport directly during training. Similar to classical solvers, where the macroscopic process does not influence the microscopic updates, off-branches do not influence any of the main-branches. Instead, each off-branch creates its predictions by repeatedly processing its own data, as well as retrieving information from the microscopic state of the main-branches (without influencing them).

\begin{figure}[h]
    \centering
    \includegraphics[width=0.5\linewidth]{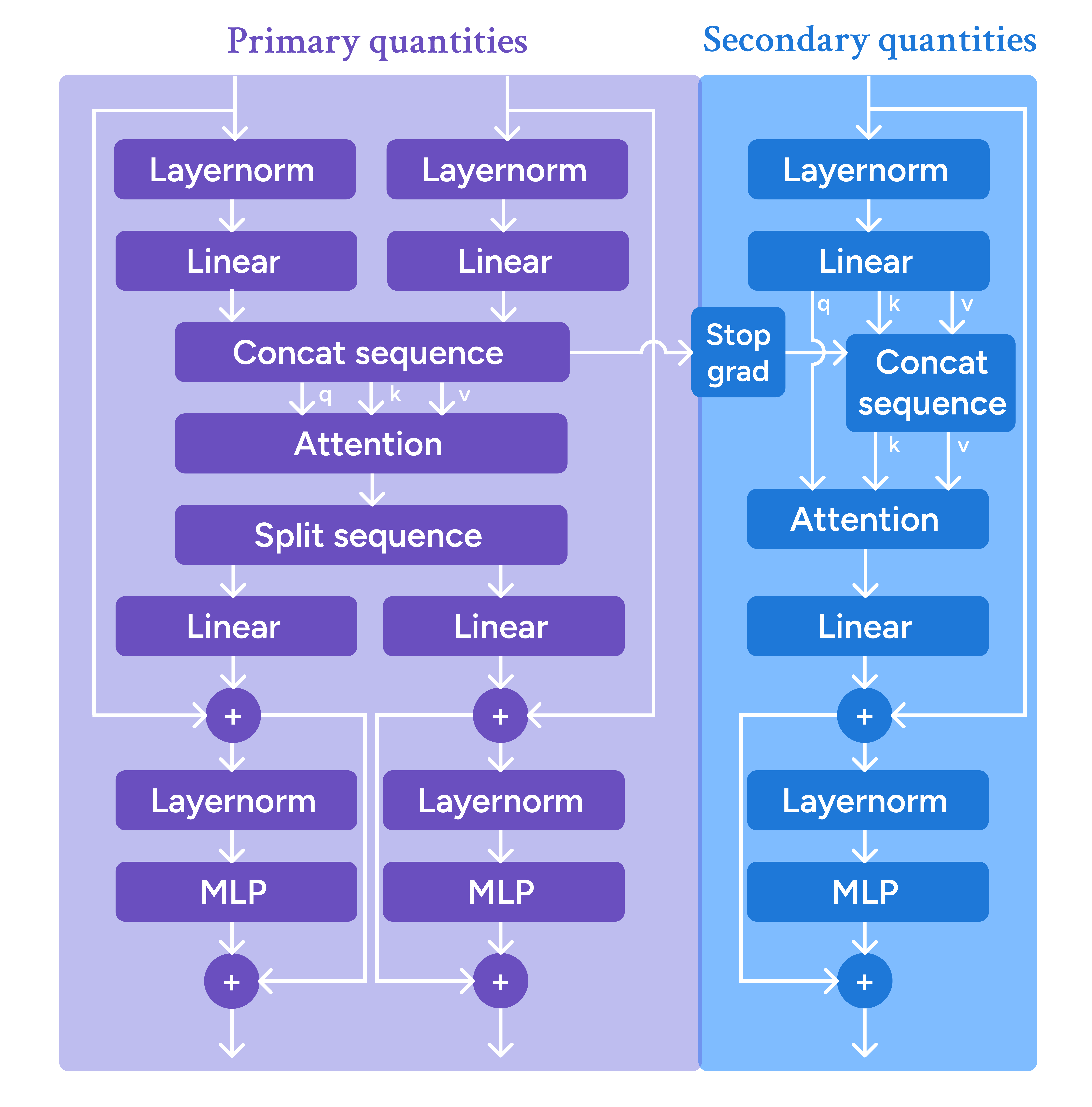}
    \caption{Schematic architecture of a \textit{multi-branch transformer block}. DiT~\cite{li22dit} modulation is applied to each attention and MLP block but is omitted for visual clarity.}
    \label{fig:schematic_multibranchtransformer}
\end{figure}

\paragraph{Multi-branch transformers.}
The central neural network component of NeuralDEM are multi-branch transformers.  Multi-branch transformers, as the name suggests, consist of multiple branches: main-branch(es) and off-branch(es).
Each branch is a stack of transformer~\citep{vaswani2017transformer} blocks where weights are not shared between branches. Each branch operates on a set of so-called tokens, which are obtained by embedding the input into a compressed latent representation. Main-branch(es) concatenate all tokens before each attention operation along the set dimension, allowing interactions between them, followed by splitting tokens again into the different branches, akin to multi-modal diffusion transformer (MMDiT) blocks~\citep{esser2024stablediffusion3}. 
Additionally, multi-branch transformers can include arbitrarily many off-branches, where the self-attention is replaced by a cross-attention which uses only its own off-branch tokens as queries and concatenates its own off-branch tokens with the main-branch tokens to use as keys and values. This roughly corresponds to simultaneous self-attention between the off-branch tokens and cross-attention between off-branch and main-branch tokens. No gradient flows through the cross-attention back to the main-branch tokens. Off-branches are implemented via a modified diffusion transformer block~\citep{li22dit}. A schematic sketch is shown in Figure~\ref{fig:schematic_multibranchtransformer}.

In our numerical experiments, we consider temporally evolving systems of multiple fields. Each input at time $t$ $\Bu^t_i$ consists of $h=1,\ldots,M$ fields, where the $h$th field at timestep $t$ is denoted as $\Bu_i^{h,t}$. Each field is modeled by one branch of the multi-branch transformer.
We create datasets of function pairs that are evaluated at $K$ and $K'$ input and output positions $(\Bu_{i,j}^{h,t}$ $\Bv_{i,j'}^{h,t+\Delta t})$ and train all $M$ branches in parallel to map $\Bu_{i,j}^{h,t}$ to the target $\Bv_{i,j'}^{h,t+\Delta t}$.
Each branch of the multi-branch transformer consists of $M$ encoders $\mathcal{E}^h$, $M$ approximators $\mathcal{A}^h$, and $M$ decoders $\mathcal{D}^h$.

\begin{align*}
&\cE^h: \Bu_{i,j=1,\dots , K}^{h,t} \in \dR^{K\times d} 
\xrightarrow[]{\text{embed}} \dR^{K \times d_{\text{hidden}}}
 \xrightarrow[]{\text{multi-branch transformer}} \Bz^{h,t}_i \in \dR^{n_{\text{latent}} \times d_{\text{hidden}}} \ \\
 &\cA^h: \Bz^{h,t}_i  \in \dR^{n_{\text{latent}} \times d_{\text{hidden}}} \xrightarrow[]{\text{multi-branch transformer}} \Bz^{h,t+\Delta t}_i \in \dR^{n_{\text{latent}} \times d_{\text{hidden}}} \ \\
 &\cD^h: (\Bz^{h,t+\Delta t}_i, \By^{h}_{i,j'=1, \dots ,K'}) \xrightarrow[]{\text{perceiver decoder}} \hat{\Bv}_{i,j'=1,\dots,K'}^{h,t+\Delta t} \in \dR^{K' \times d} \ .
\end{align*}

\subsection{Scalar parameter conditioning} \label{sec:method_conditioning}

Physical simulations often require various scalar parameters such as material properties (e.g., friction or particle size) or geometry variations (e.g., slope angles or outlet width) to define the simulation properties. It is vital to provide these scalars also to the machine learning model to produce accurate results. A common way to do this is by feature modulation~\cite{perez18film} which scales and shifts intermediate feature activations based on a vector representation of the scaler parameters. As NeuralDEM is a transformer architecture, we use DiT-style modulation~\citep{li22dit} which scales, shifts and gates the activations of each attention and MLP block based on a learned vector representation of the scalar parameters.
This form of conditioning allows NeuralDEM to generalize across geometries and across non-trivial particle-particle interactions by condition on respective variables.

A particular intriguing property of this conditioning mechanism is that it allows us to condition on parameters that describe only the macroscopic material behavior. For example, our model can be conditioned on the measured parameters, like the internal friction angle or the flow function coefficient 
from a shear cell device,
instead of requiring the microscopic friction parameters necessary for simulating with a classical DEM solver (e.g., particle sliding friction coefficient). In practice, this allows us to simulate any material by simply using a shear cell to determine its friction angle. In contrast, for classical solvers, one would need to estimate the microscopic friction parameters of the material using a calibration procedure~\cite{coetzee2017} in order to simulate it, which is tedious, error prone and often inaccurate. We showcase this in Section~\ref{sec:macroscopic_conditioning_experiments}.

\subsection{Flexible model architecture for variable simulation use-cases.} \label{sec:flexible_model_architecture}

As physical simulations exhibit a broad range of dynamics, and relevant macroscopic insights can vary drastically depending on the use-case, our multi-branch transformer architecture should be seen as a flexible framework that enables various use-cases instead of a ``set in stone'' architecture. Components can become redundant in certain settings, or special use-cases could require additional components. For example, in simulations with laminar or pseudo-steady dynamics, the whole simulation is fully specified by the initial state, making encoding subsequent states and interaction between branches redundant. 
However, in very unsteady systems all components of the multi-branch transformer architecture are very much necessary to produce accurate time evolution as slightly different initial states can lead to vastly different instantiations of dynamics, which requires a physically accurate state at each timestep and interactions between the states of different branches.

Additionally, the initial encoding of physics phases benefits from specialized designs, depending on the input data. For irregular grid data (e.g., particles), we use the supernode pooling from UPT~\citep{alkin2024upt} which aggregates information around particles via message passing to so-called \textit{supernodes}, which are randomly selected particles. Fluid phases are typically represented via regular grid data, which is computationally more efficient and allows efficient coupling to, e.g., particle simulations. For regular grid data, we use the vision transformer patch embedding~\cite{dosovitsky2021vit} which splits the input into non-overlapping patches and embeds them using a shared linear projection.

Finally, decoding is performed using the same architecture for both particle and grid data, using a perceiver-based neural field decoder~\citep{jaegle2021perceiver}, which is queried at locations $\By_{i, j'=1,\dots,K'}$ in parallel. This type of decoding first embeds query locations to be used as queries for the perceiver cross-attention and uses the latent tokens as keys and values. This results in a point-wise evaluation of the latent space based on the query position, which is what enables effective parallelization. 

We use a standard pre-norm vision transformers architecture~\cite{baevski2019prenorm,dosovitsky2021vit} where each branch of the multi-branch transformer corresponds to a single vision transformer. The total number of blocks is evenly distributed across encoder, approximator and decoder.

\clearpage
\section{Numerical experiments}
\begin{figure}[h]
    \centering
    \begin{subfigure}{0.22\textwidth}
        \centering
        \includegraphics[width=0.7\linewidth]{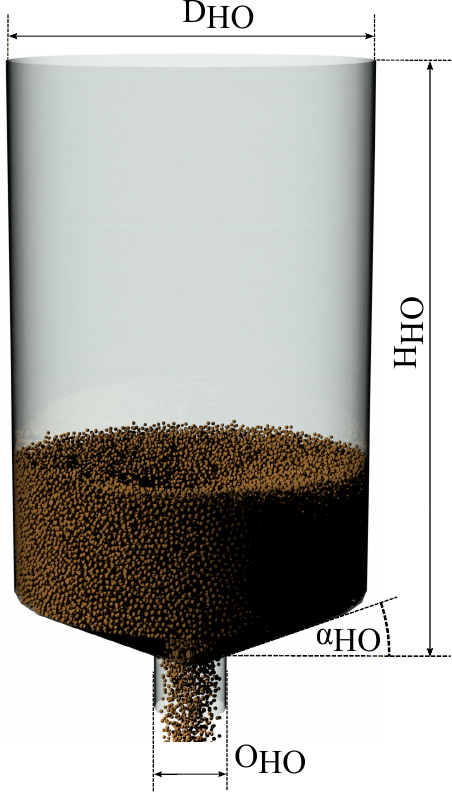}
        \vspace{0.5cm}
        \hspace{4cm}
        \caption{Hopper}
        \label{fig:hopper_geometry}
    \end{subfigure}
    \hspace*{0.025\textwidth}
    \begin{subfigure}{0.2\textwidth}
        \centering
        \includegraphics[width=0.7\linewidth]{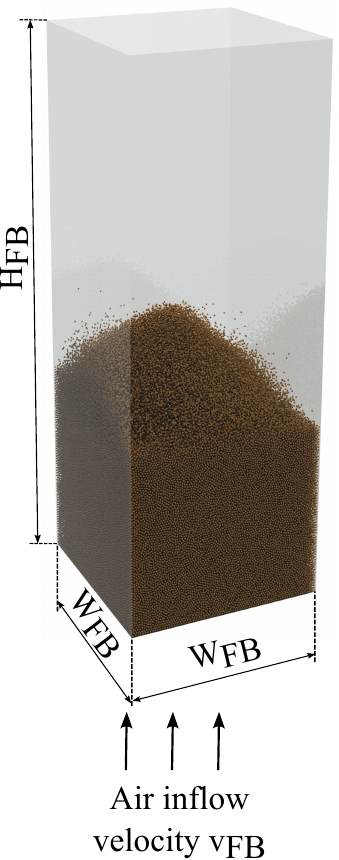}
        \caption{Fluidized bed}
        \label{fig:fb_geometry}
    \end{subfigure}
    \begin{subfigure}{0.45\textwidth}
        \begin{tabular}{lr}\toprule
            Parameter & Value  \\ \midrule
            Hopper diameter $D_{\text{HO}}$ & \SI{0.24}{\meter}  \\
            Hopper height $H_{\text{HO}}$ & \SI{0.40}{\meter}  \\
            Hopper outlet diameter $O_{\text{HO}}$ & \SI{0.05}{\meter}  \\
            Hopper angle $\alpha_\text{HO}$ & 0-60° \\
            Hopper particle diameter $d$ & \SI{0.004}{\meter}  \\ \midrule
            FB width $w_{\text{FB}}$ & \SI{0.096}{\meter}  \\
            FB height $h_{\text{FB}}$ & \SI{0.24}{\meter}  \\
            FB particle diameter $d_{\text{FB}}$ & \SI{0.0012}{\meter}  \\\bottomrule
            \\
        \end{tabular}
        \caption{Dimensions of the cases}
        \label{table:dimension}
    \end{subfigure}
    \caption{Schematic of the two numerical experimental cases and the associated dimensions.}
    \label{fig:setups}
\end{figure}

We test the NeuralDEM framework on two industrially relevant use cases: hoppers and fluidized bed reactors, both visualized in Figure \ref{fig:setups} and described in  Table~\ref{tab:dataset_overview}. We evaluate NeuralDEM on different metrics: (i) \textbf{Effectiveness} of field-based modeling w.r.t.\ macroscopic quantities. We extract and compare emerging macroscopic physics phenomena. (ii) \textbf{Scalability} towards industry relevant simulation sizes. We train on simulations with up to half a million particles and anticipate good scaling behavior to much higher numbers. (iii) Physically accurate \textbf{time extrapolation}. We show that our models can faithfully model dynamics for long-time horizons -- in a fraction of the time that a classical solver would take. (iv) \textbf{Generalization} to unseen regions in the design parameter space.

\begin{table}[h]
    \centering
    \resizebox{\textwidth}{!}{
    \begin{tabular}{lcccl}\toprule
        Name & Simulation method & \#Particles & \#CFD cells  & Variability \\
        \midrule
        Hopper & DEM & 250k & -  &  Hopper angle, material friction\\
        Fluidized bed & CFD-DEM & 500k & 160k  & Fluid inlet velocity \\\bottomrule
    \end{tabular}
    }
    \vspace{1em}
    \caption{
    Overview of dataset properties used in the experiments.}
    \label{tab:dataset_overview}
\end{table}

\subsection{Simulation setup and problem scale}
 \textbf{Hoppers} are industrially used for short as well as long term storage of particulate material, showcasing slow and pseudo-steady macroscopic behavior. DEM is the preferred method since the air around the particles can usually be neglected due to the slow velocities in the system. In our experiments, the hopper geometry, as shown in Figure \ref{fig:hopper_geometry}, is filled with 250k particles, which gradually exit the domain over the simulation duration when the hopper empties. Timestepping of DEM solvers strongly depends on particle size as well as particle properties. A timestep of $\Delta t\st{DEM} = \SI{10}{\micro\second}$ is required for the tested numerical experiments with LIGGGHTS~\cite{kloss2012}.

\textbf{Fluidized bed reactors} are characterized by fast and transient phenomena and are widely used in industry for a variety of processes. Fluidized bed reactors showcase strong interactions of the particles with the surrounding fluid, necessitating an accurate modeling of particles, the gas phase, as well as particle-gas interactions. Thus, modeling approaches need to combine DEM parts with simulations of the surrounding fluid. For data generation, we use a coupled CFD-DEM approach~\cite{goniva2012} 
which is built upon LIGGGHTS~\cite{kloss2012} and OpenFOAM~\cite{weller:98}.
    The geometry of the setup is sketched in Figure \ref{fig:fb_geometry} and the dimensions for both cases can be found in Table \ref{table:dimension}. The reactor is filled with 500k particles and the fluid, i.e., air, that is uniformly pushed into the reactor from the bottom is modeled on a grid of 160k hexahedral cells.

\clearpage

\subsection{Hopper}\label{sec:hopper}

\begin{figure}[h!]
\centering
\begin{subfigure}[b]{0.8\linewidth}
\begin{tabular}{ c m{2.1cm} m{2.1cm} m{2.1cm} m{2.1cm}}
\rotatebox[origin=c]{90}{~\parbox[c]{3cm}{\footnotesize\centering Ground truth \\ DEM simulation}~}
&\includegraphics[width=\linewidth ,trim={3cm 3cm 3cm 3cm},clip]{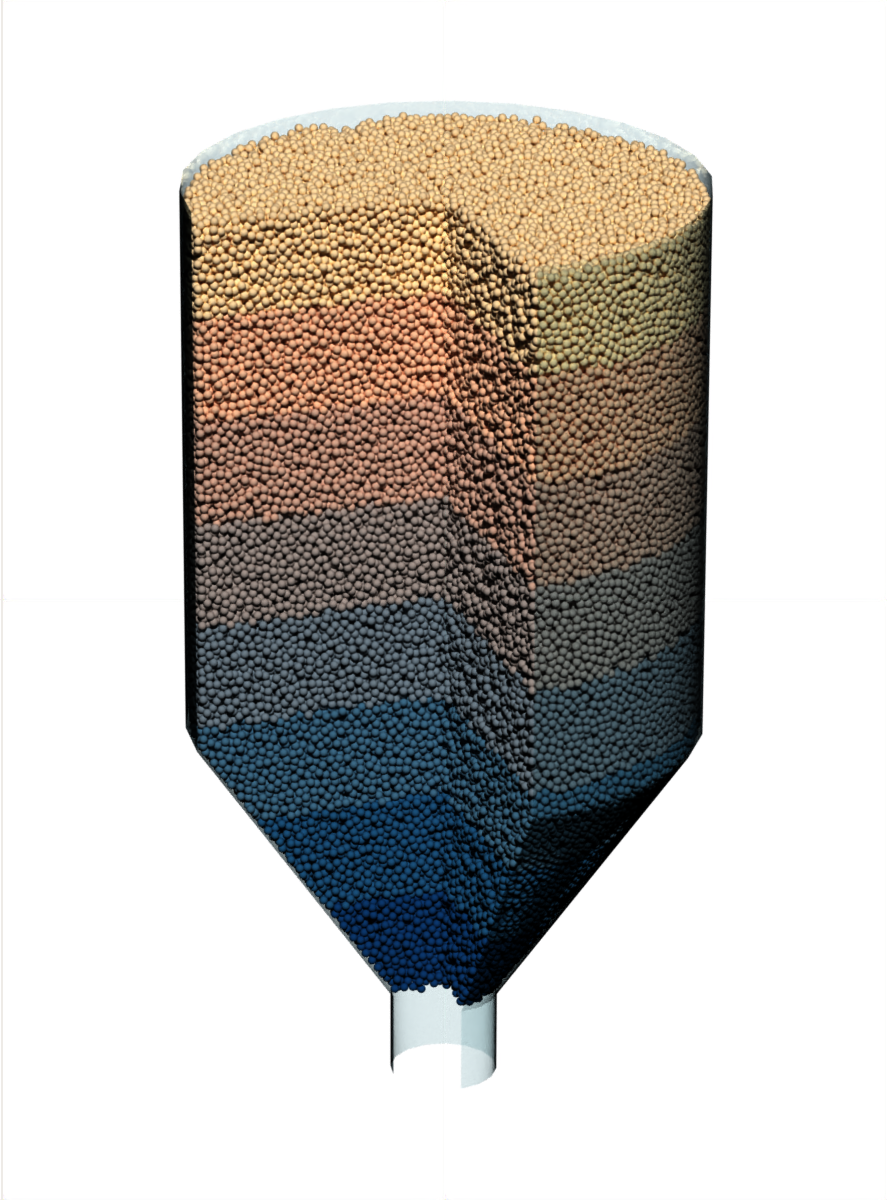} & \includegraphics[width=\linewidth,trim={3cm 3cm 3cm 3cm},clip]{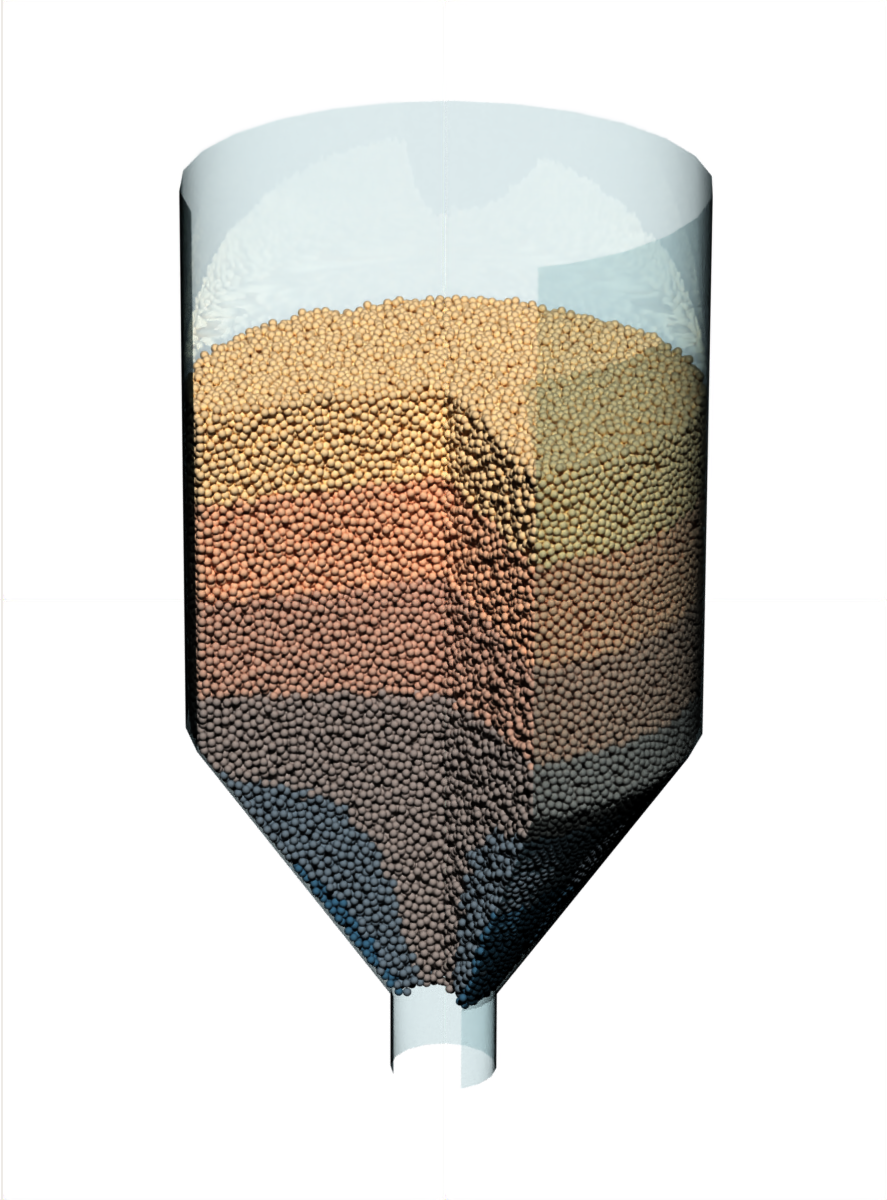} & \includegraphics[width=\linewidth,trim={3cm 3cm 3cm 3cm},clip]{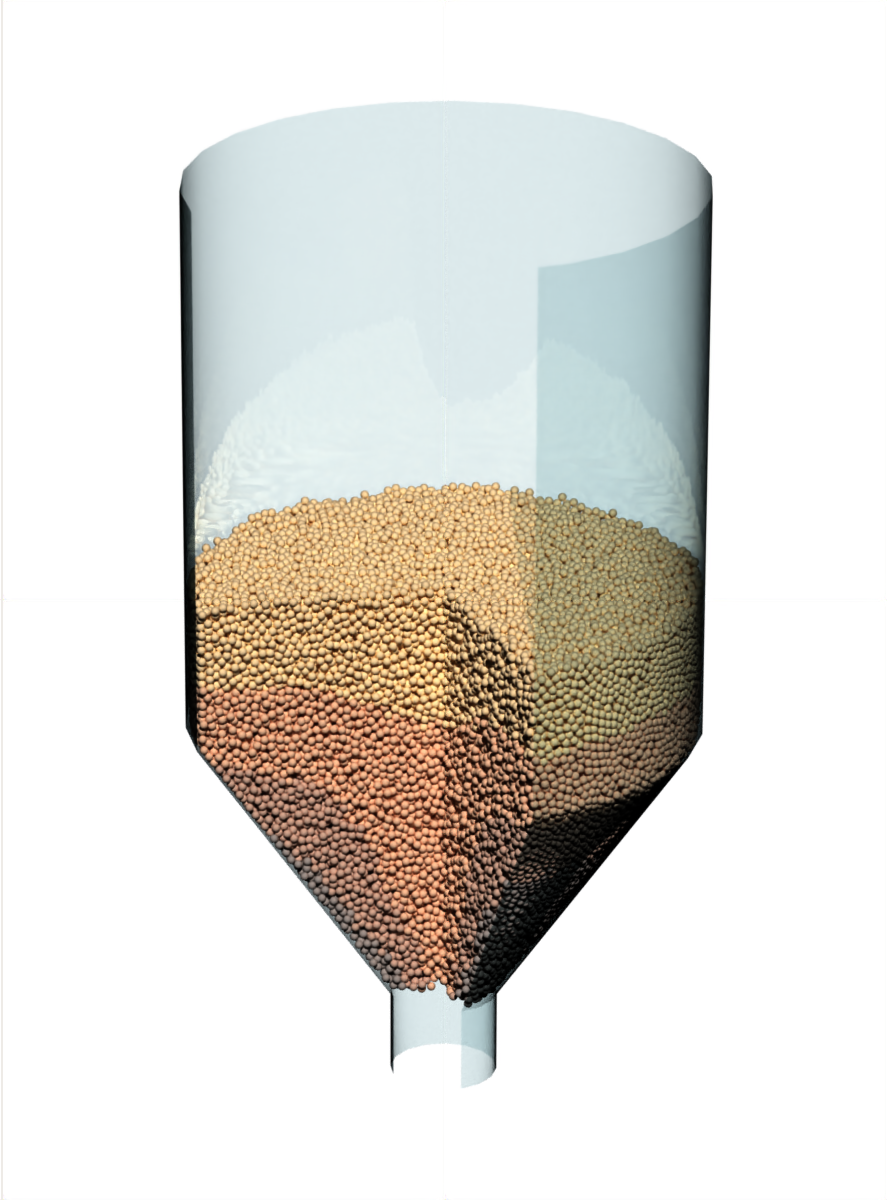} & \includegraphics[width=\linewidth,trim={3cm 3cm 3cm 3cm},clip]{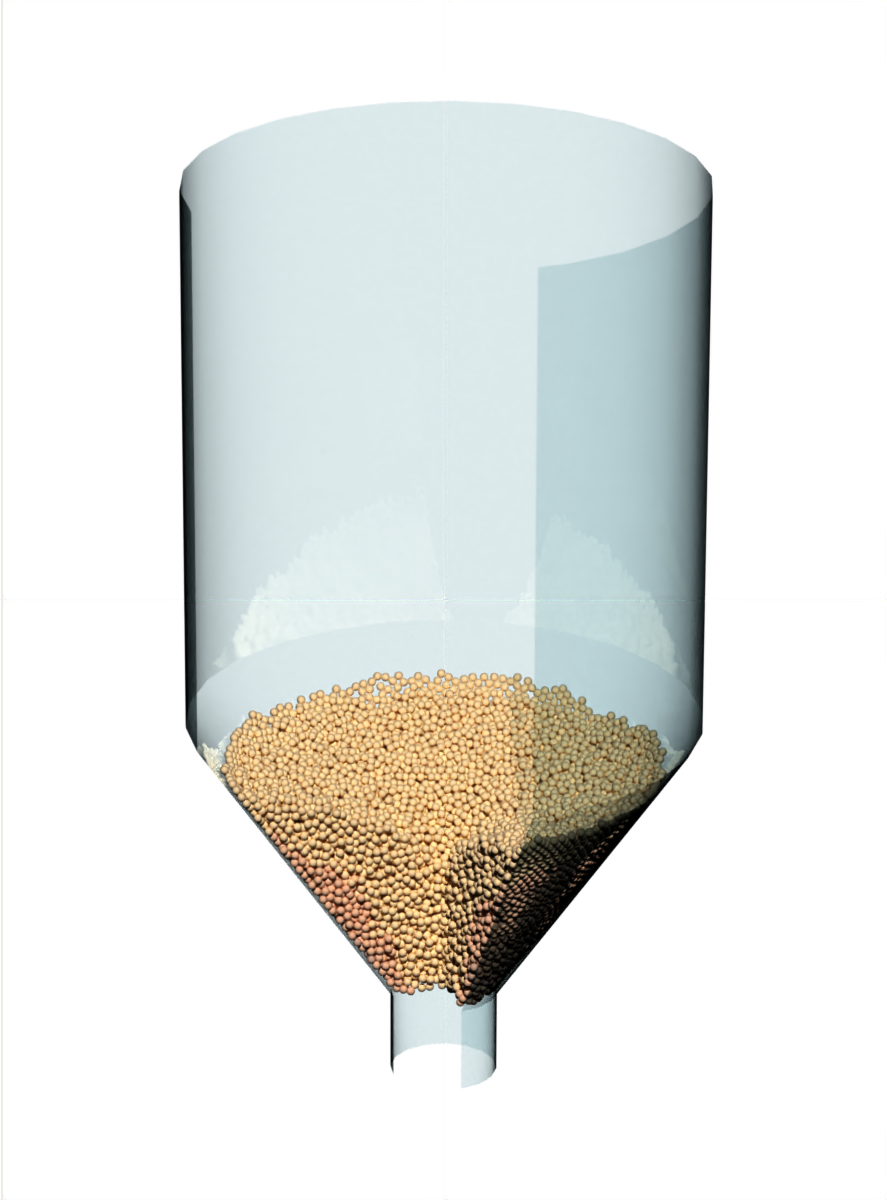} \\

\rotatebox[origin=c]{90}{~\parbox[c]{3cm}{\footnotesize\centering NeuralDEM \\ generated trajectory}~}
&\includegraphics[width=\linewidth ,trim={3cm 3cm 3cm 3cm},clip]{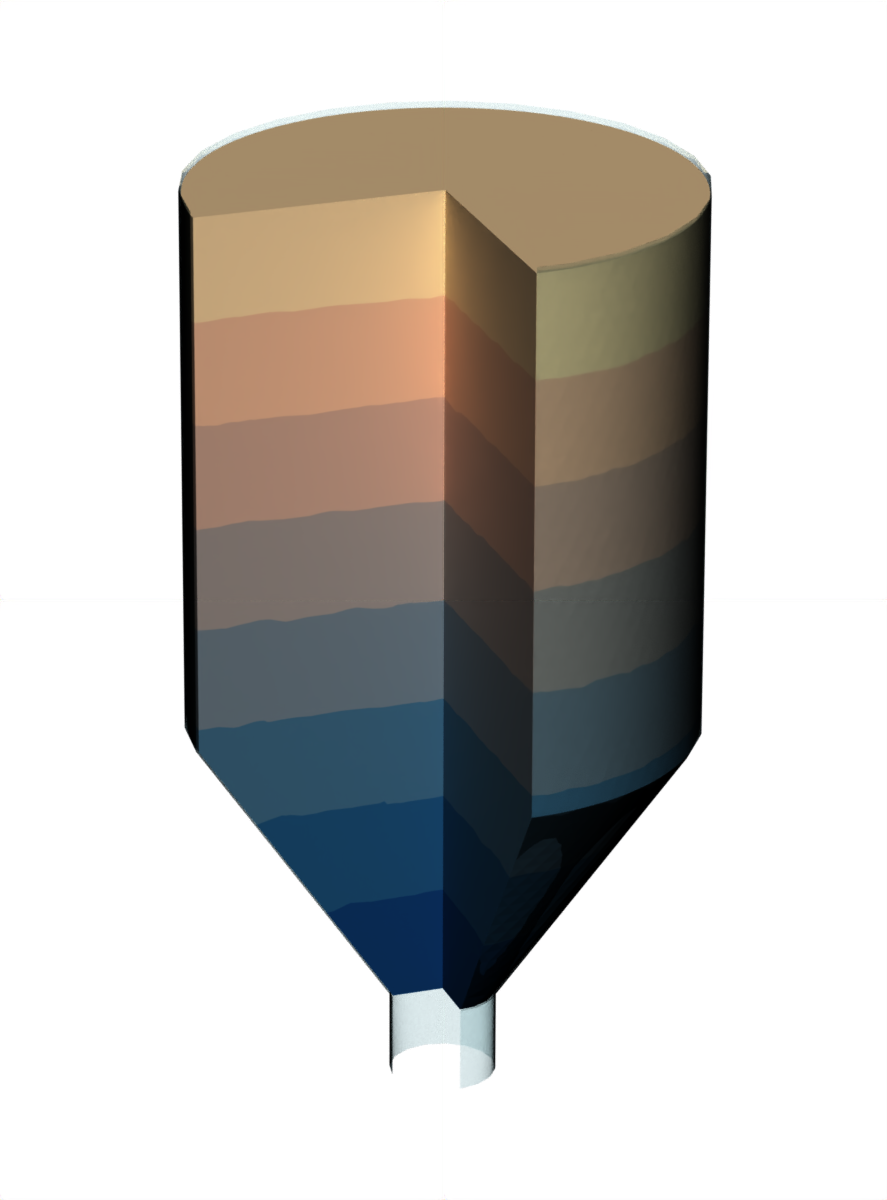} & \includegraphics[width=\linewidth,trim={3cm 3cm 3cm 3cm},clip]{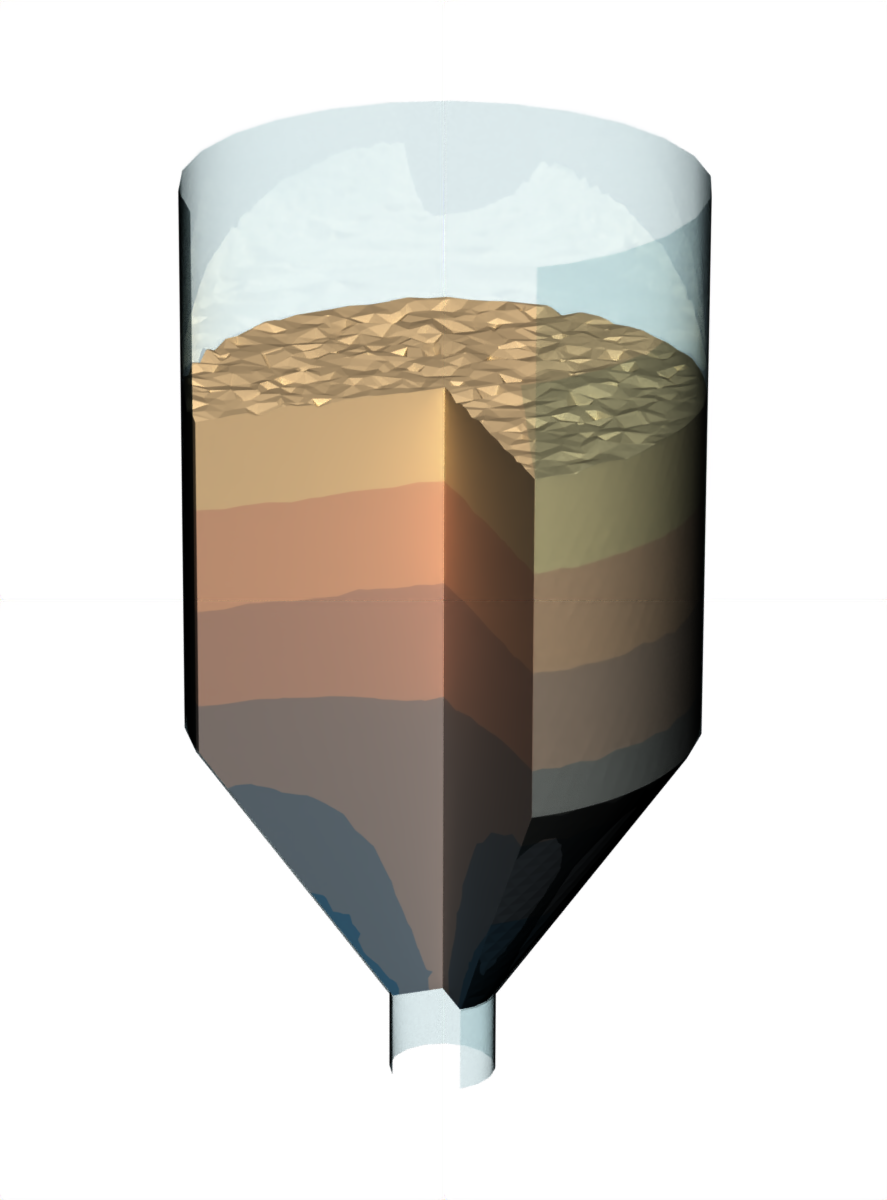} & \includegraphics[width=\linewidth,trim={3cm 3cm 3cm 3cm},clip]{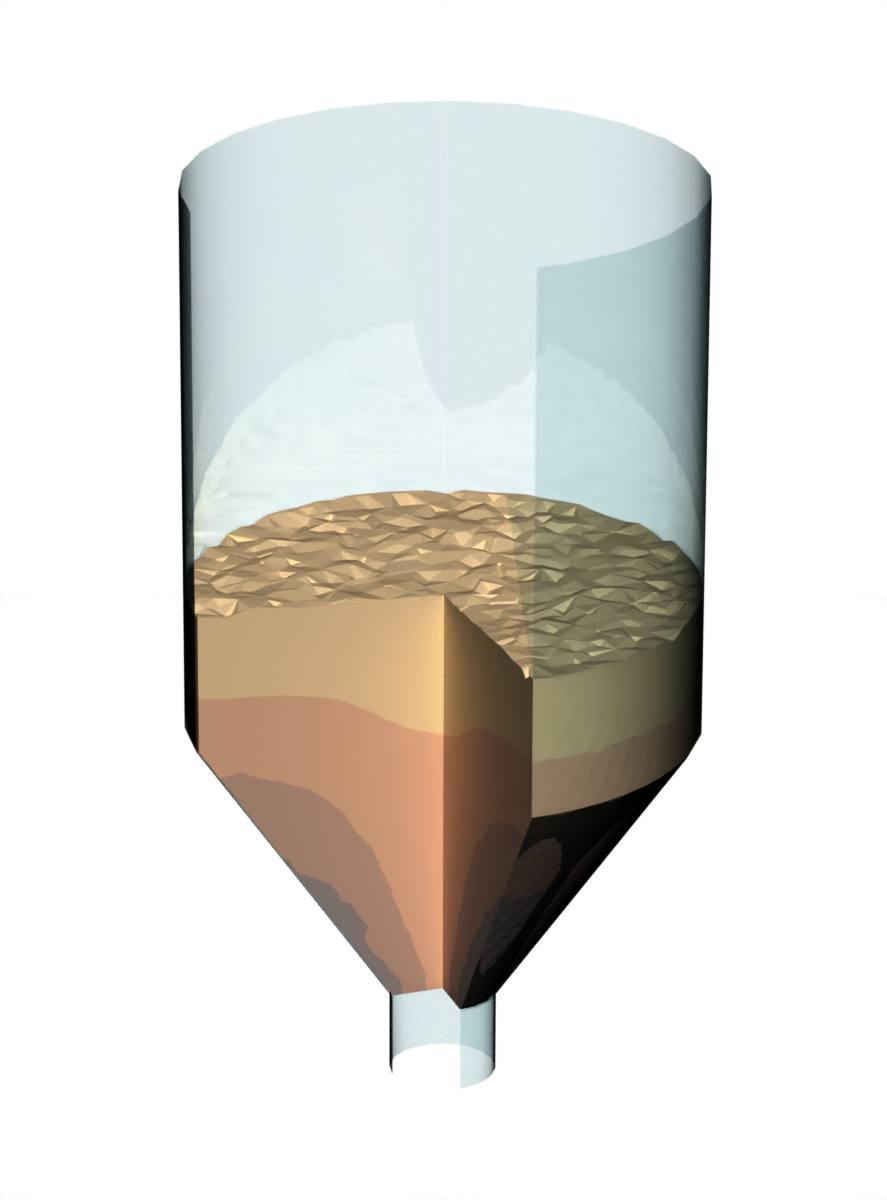} & \includegraphics[width=\linewidth,trim={3cm 3cm 3cm 3cm},clip]{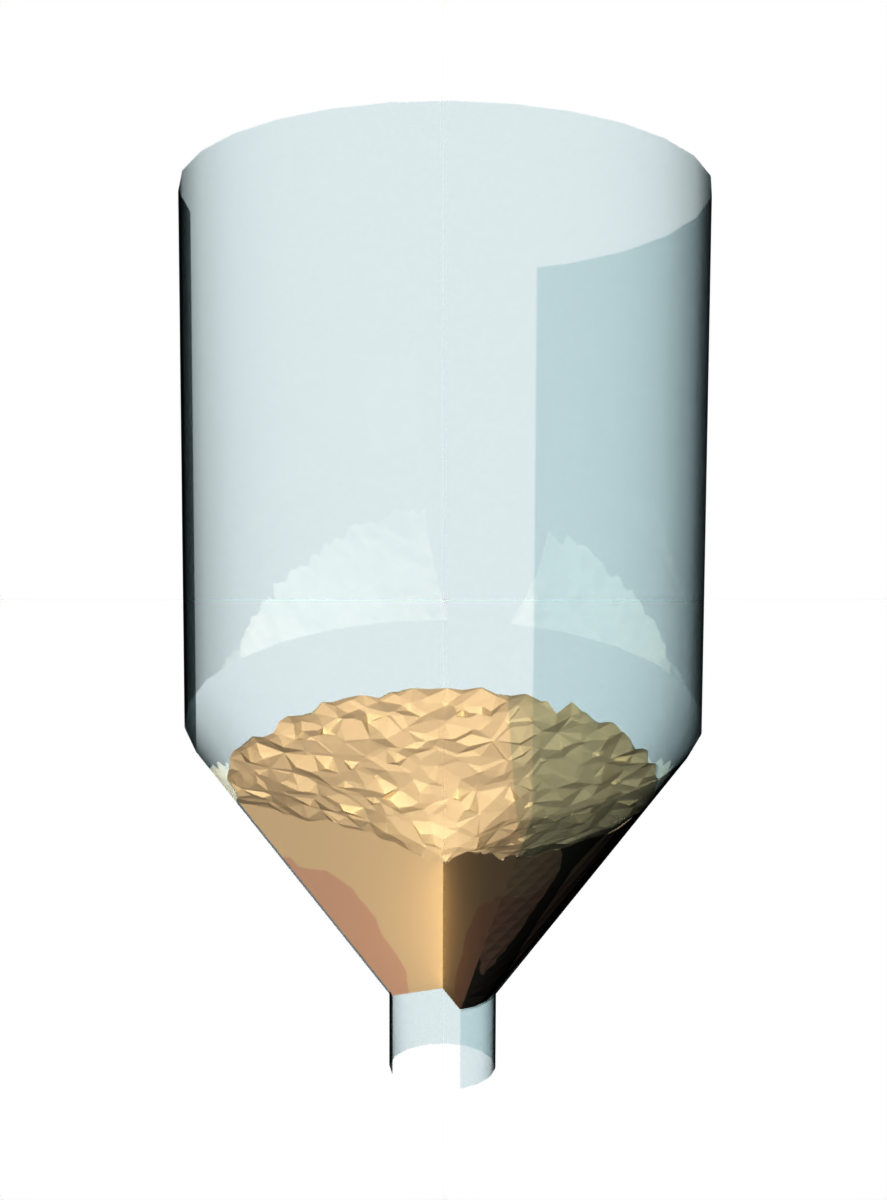} \\

 \end{tabular}

 \caption{Mass flow regime}
\label{fig:draining_hopper_transport_visualization:mass}
\end{subfigure}
\par\bigskip

\begin{subfigure}[b]{0.8\linewidth}
\begin{tabular}{ c m{2.1cm} m{2.1cm} m{2.1cm} m{2.1cm} }
\rotatebox[origin=c]{90}{~\parbox[c]{3cm}{\footnotesize\centering Ground truth DEM simulation}~}
&\includegraphics[width=\linewidth ,trim={3cm 3cm 3cm 3cm},clip]{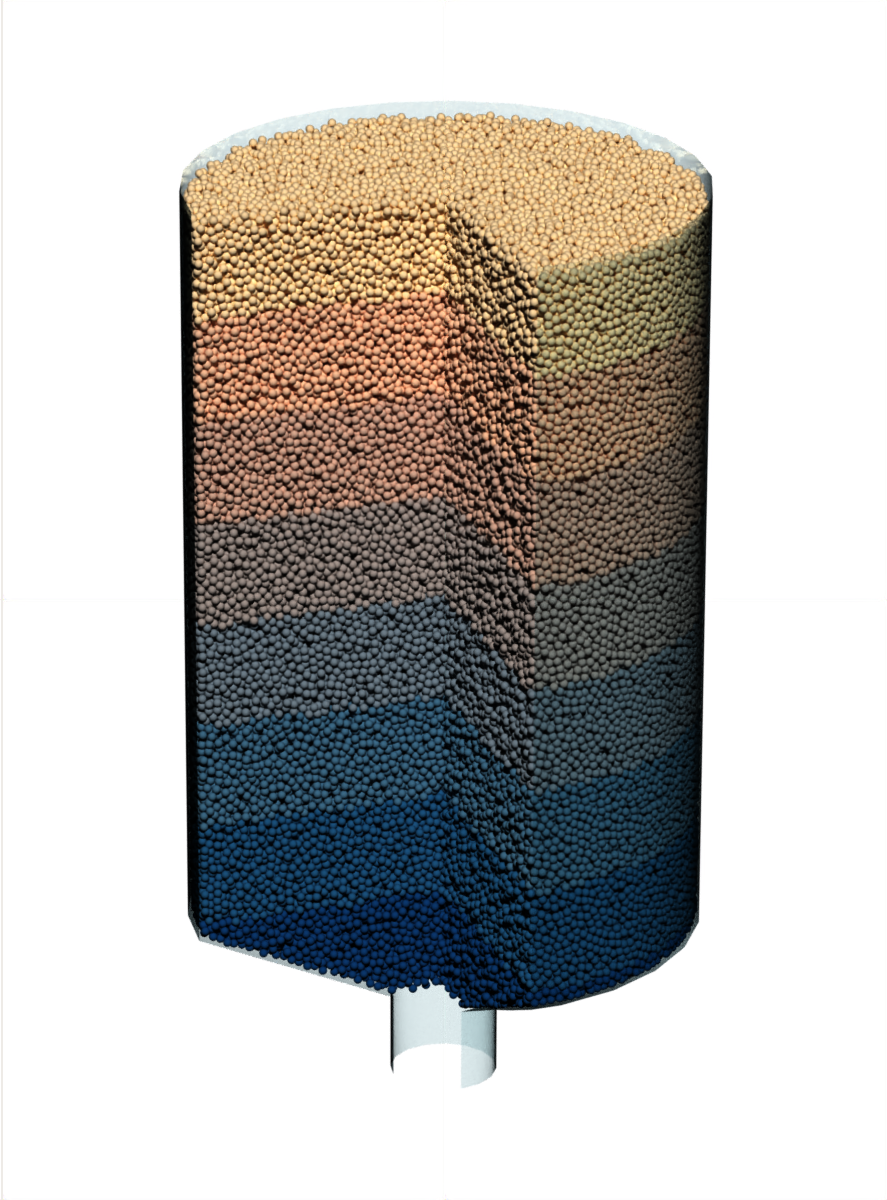} & \includegraphics[width=\linewidth,trim={3cm 3cm 3cm 3cm},clip]{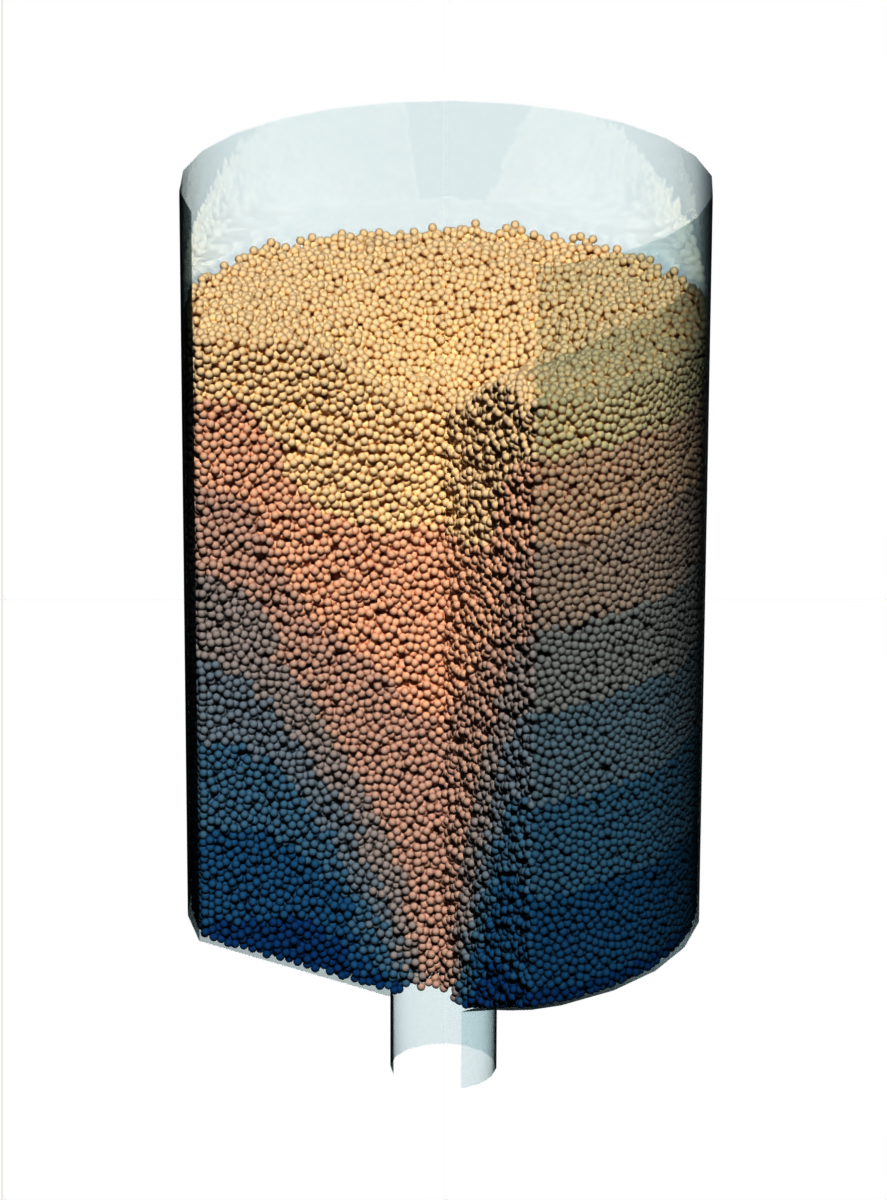} & \includegraphics[width=\linewidth,trim={3cm 3cm 3cm 3cm},clip]{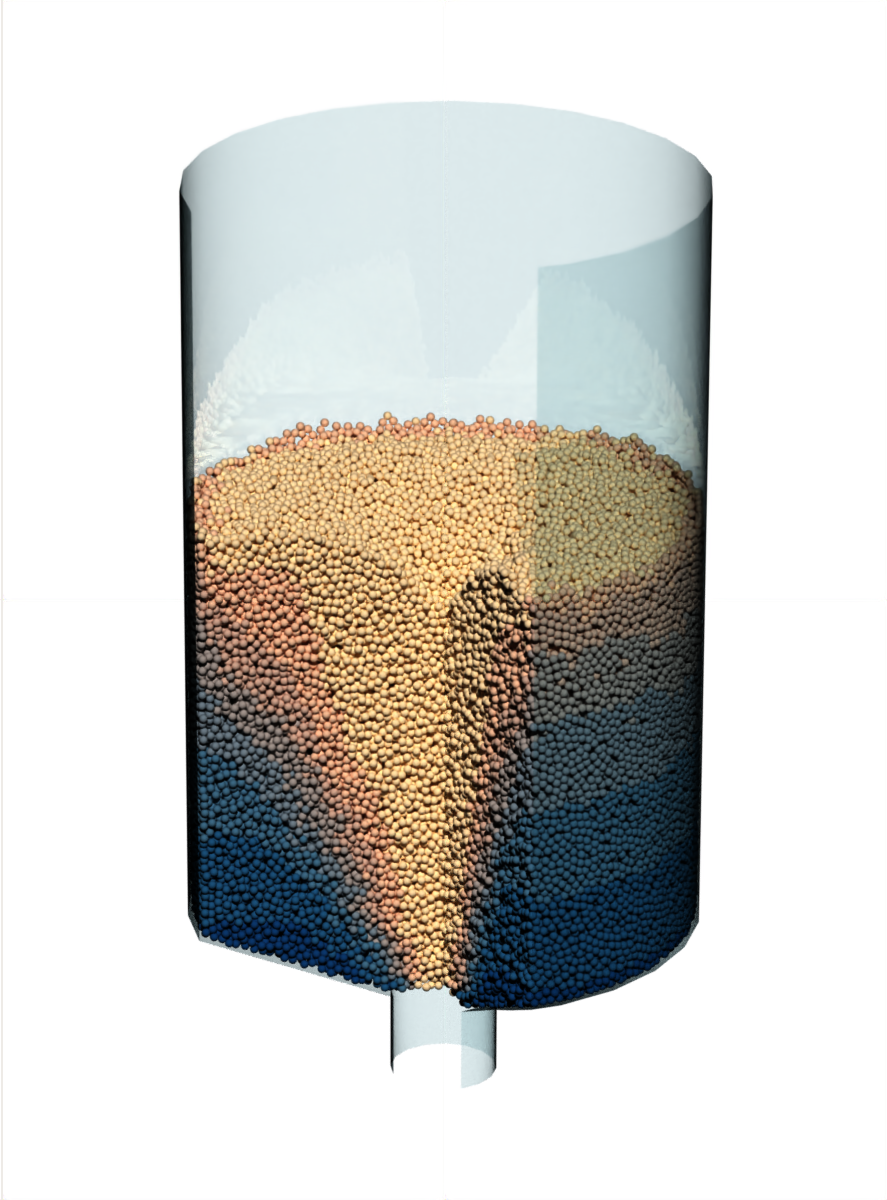} & \includegraphics[width=\linewidth,trim={3cm 3cm 3cm 3cm},clip]{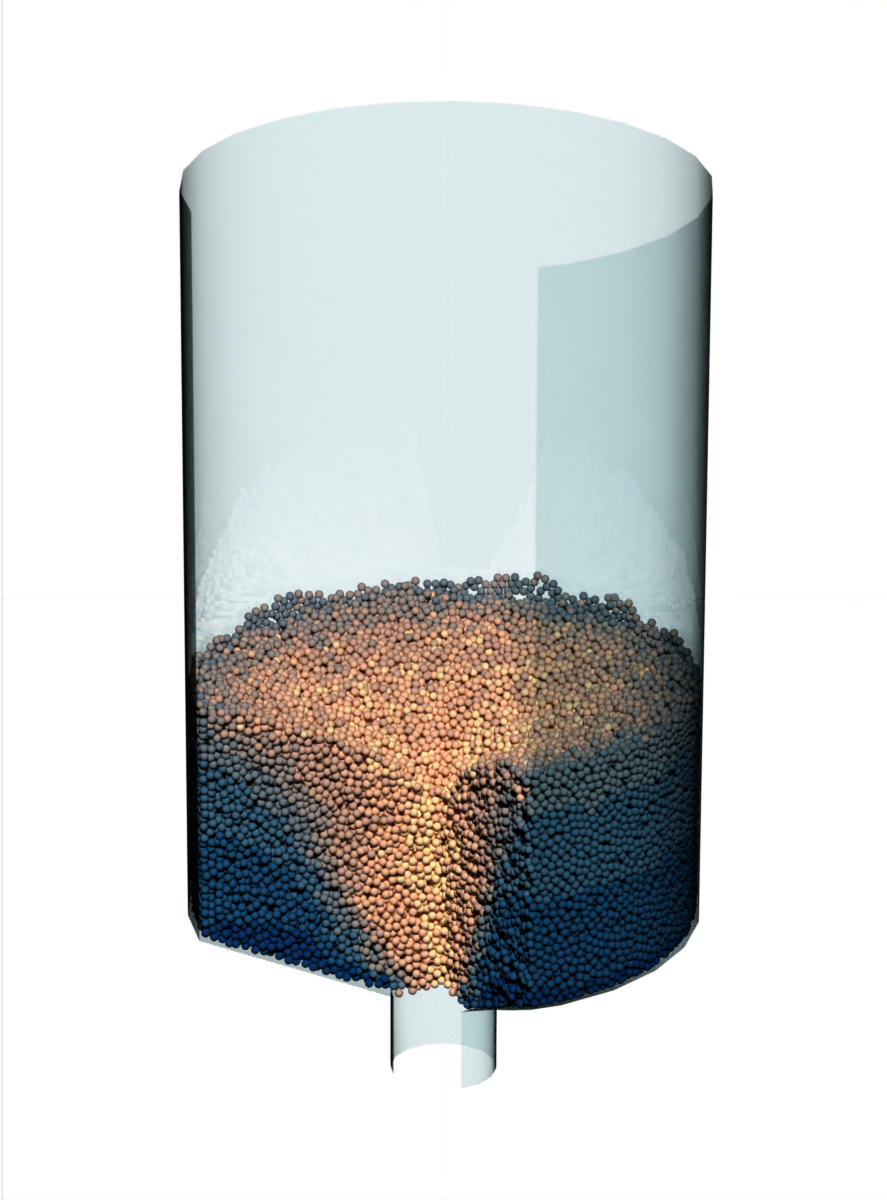} \\

\rotatebox[origin=c]{90}{~\parbox[c]{3cm}{\footnotesize \centering NeuralDEM \\ generated trajectory}~}
&\includegraphics[width=\linewidth ,trim={3cm 3cm 3cm 3cm},clip]{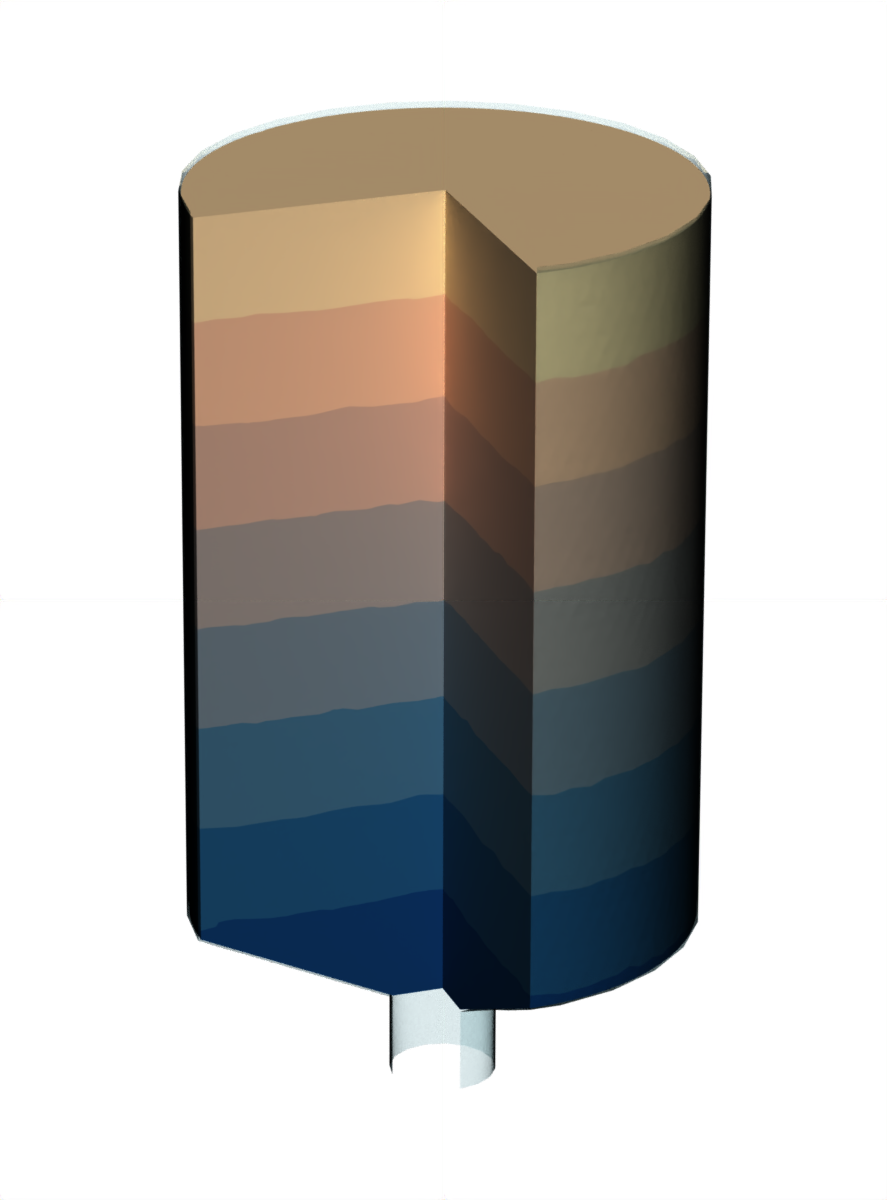} & \includegraphics[width=\linewidth,trim={3cm 3cm 3cm 3cm},clip]{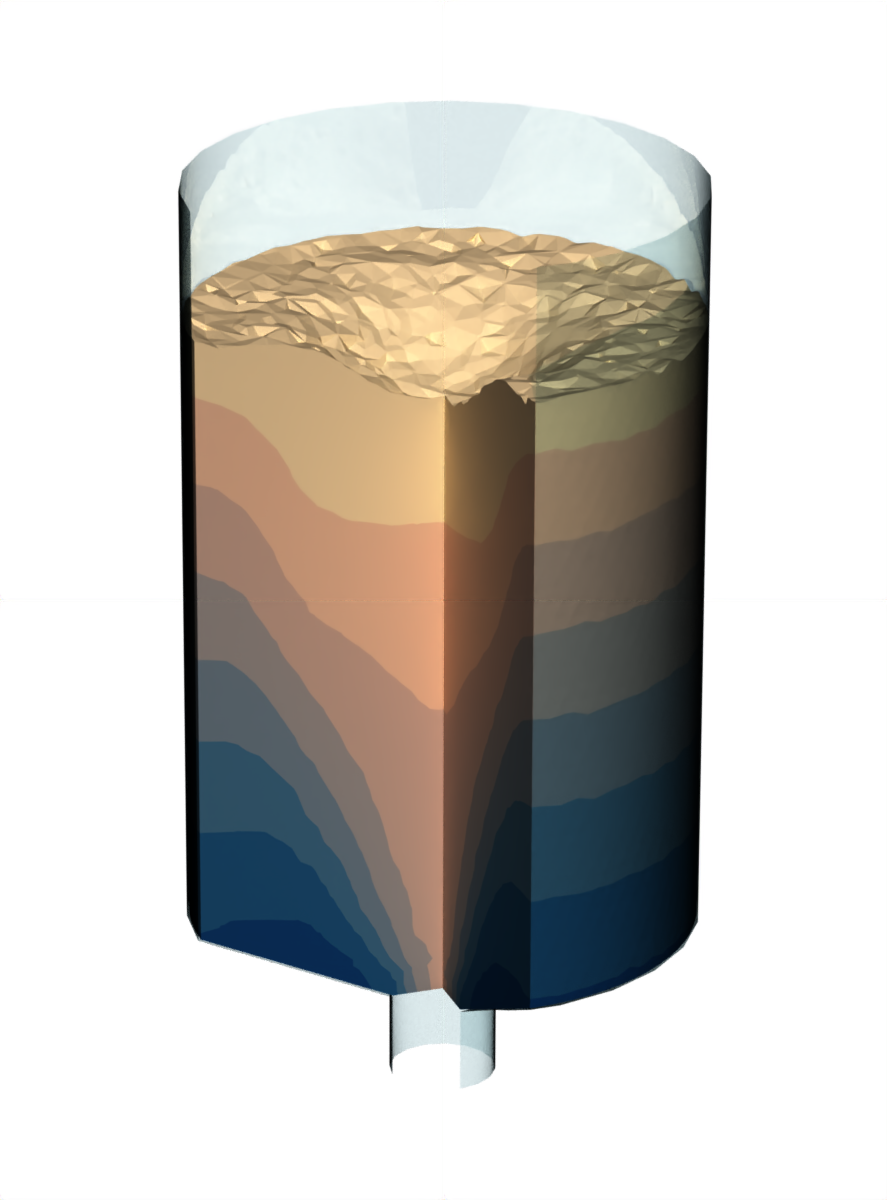} & \includegraphics[width=\linewidth,trim={3cm 3cm 3cm 3cm},clip]{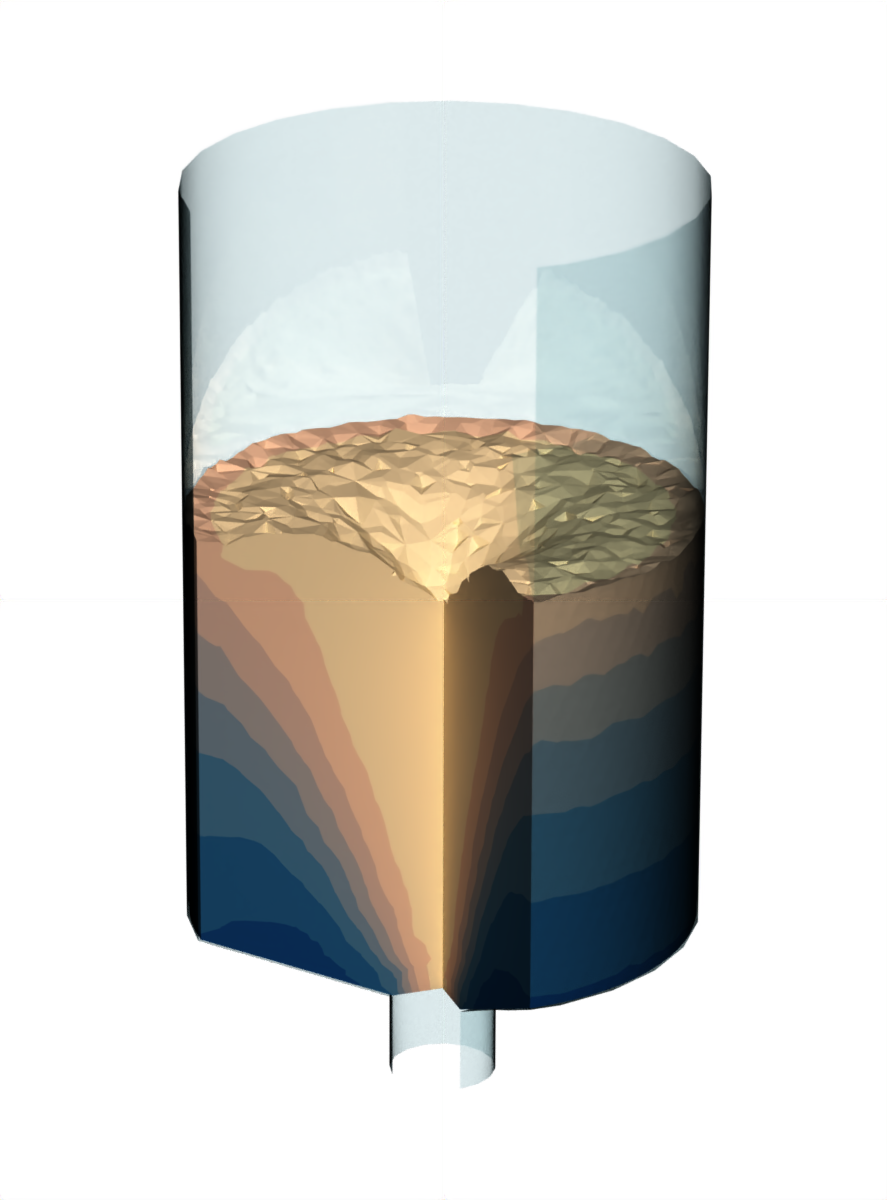} & \includegraphics[width=\linewidth,trim={3cm 3cm 3cm 3cm},clip]{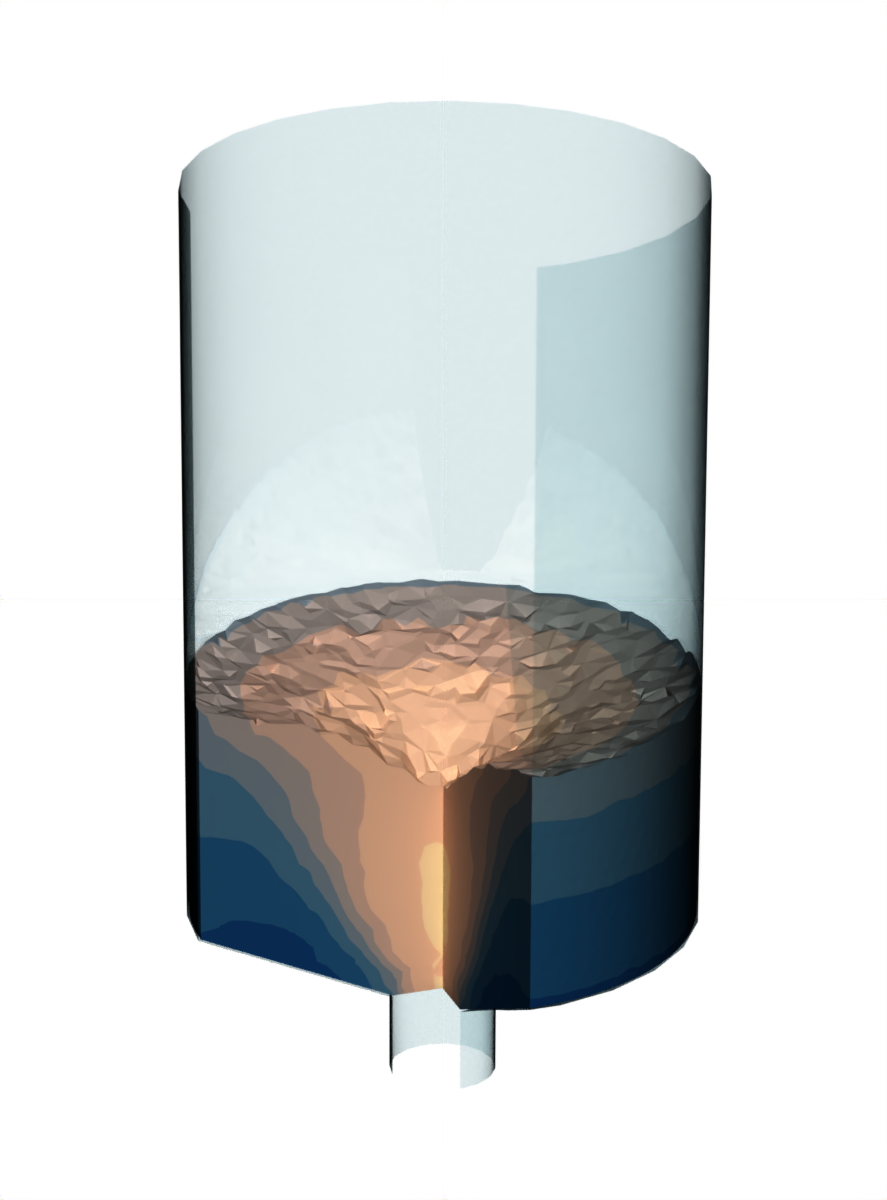} \\

 \end{tabular}
  \caption{Funnel flow regime}
\label{fig:draining_hopper_transport_visualization:funnel}
 \end{subfigure}
\par\bigskip
 \begin{subfigure}[b]{0.4\linewidth}
\includegraphics[width=\linewidth]{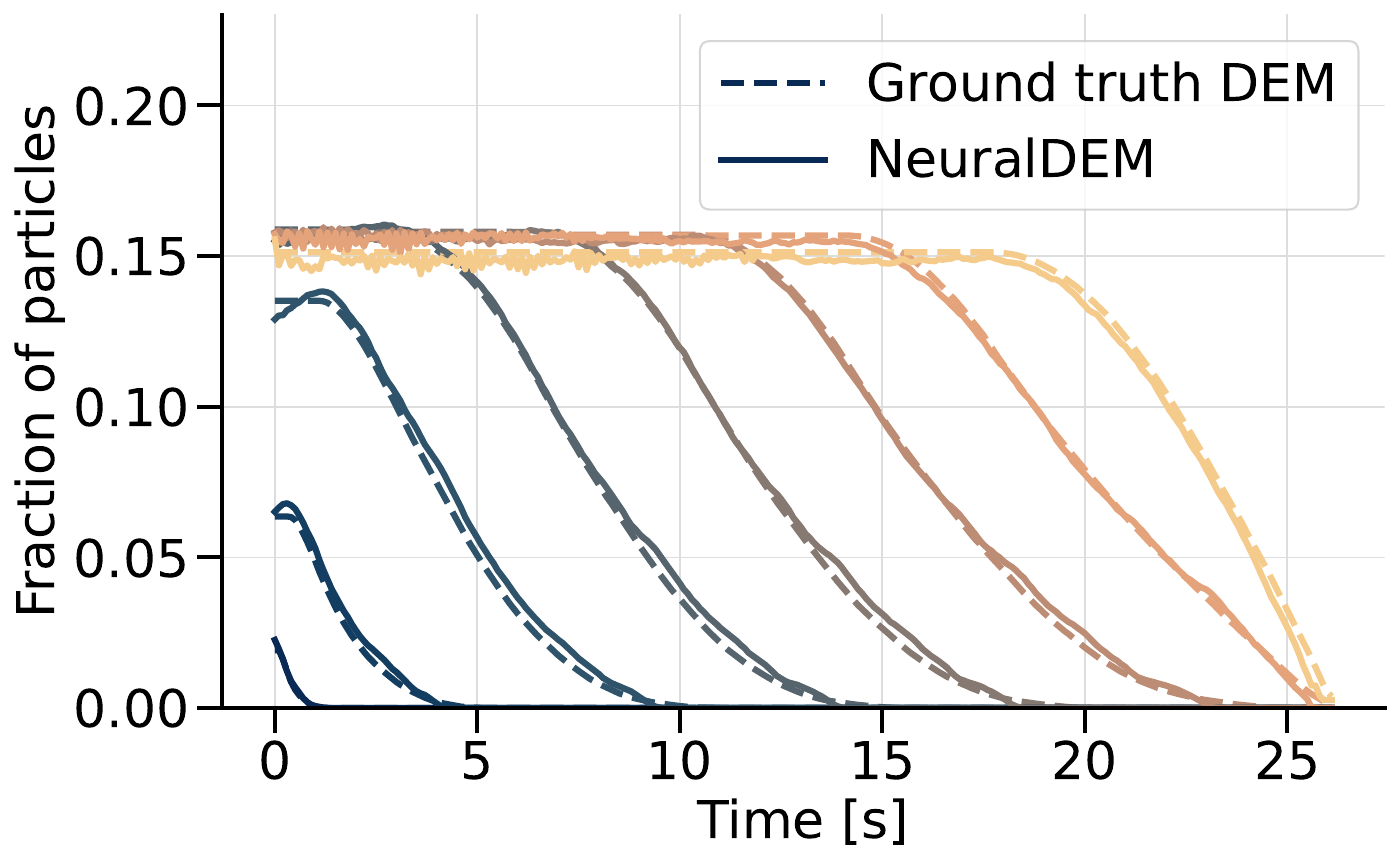}
\caption{Mass flow regime}
\label{fig:draining_hopper_transport_visualization:massdiag}
\end{subfigure}
\begin{subfigure}[b]{0.4\linewidth}
\includegraphics[width=\linewidth]{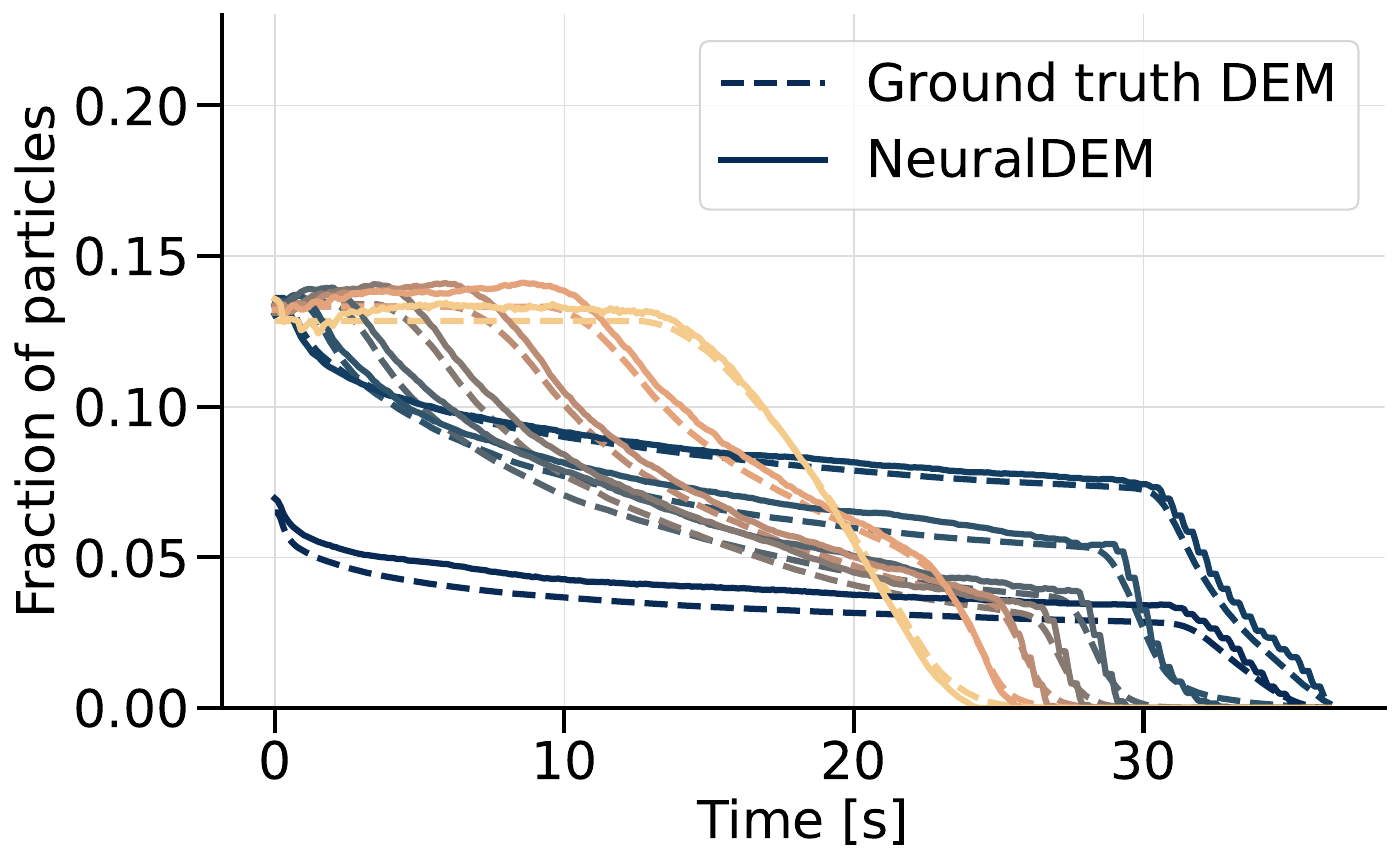}
\caption{Funnel flow regime}
\label{fig:draining_hopper_transport_visualization:funneldiag}

\end{subfigure}

\caption{Visualization of a NeuralDEM generated trajectory vs the DEM simulation. Different colors indicate different initial particle layers. Emerging funnel and mass flow regimes are clearly visible. In mass flow (a), mass moves uniformly down towards the outlet. In funnel flow (b), particles primarily move down a funnel above the outlet.  The fractions of colored material over time are shown for (c) mass flow regime and (d) funnel flow. In the funnel flow regime particle layer inversion happens, i.e., particles from higher layers overtake particles from the lower layers through the funnel. This emerging macroscopic phenomena is perfectly modeled by NeuralDEM.}
\label{fig:draining_hopper_transport_visualization}
\end{figure}

We consider hopper simulations with the hopper geometry depicted in Figure~\ref{fig:hopper_geometry}. With its outlet closed, the hopper is initially filled with particles, to roughly 250k grains on average (particle counts can vary based on the outlet slope $\alpha_\text{HO}$). Then the outlet at the bottom of the hopper is opened and grains start to flow out. 
By default, we do not refill any new particles into the hopper but consider an operation mode where the material is continuously refilled in Section~\ref{sec:hopper_refilling}.
Different simulations in the dataset vary hopper geometry and particle friction as specified in Table~\ref{tab:hopper_dataset_generation_parameter_sampling}. We create a dataset of 1000 simulations with a train/validation/test split of 800/100/100.
This variability in simulation parameters results in different flow regimes (``funnel flow'' or ``mass flow'') where the dynamics are slow and pseudo-steady. 
In funnel flow, particles primarily move down a funnel above the outlet, whereas in mass flow, material moves uniformly down towards the outlet, see Figure~\ref{fig:draining_hopper_transport_visualization}.
Each simulation is run to cover $40$ physical seconds at most (the simulation is stopped if no particles remain in the hopper). Snapshots are stored in \SI{0.1}{s} intervals (resulting in 400 ML timesteps $\Delta t\st{ML}$) which is the data that NeuralDEM models are trained on. The DEM solver requires 10k timesteps per 0.1 physical seconds and a single simulation takes roughly 3 hours on 16 CPUs. We additionally evaluate the parameters of each simulation in a shear cell to get its internal friction angle $\theta$ and flow function coefficient ffc to use it as conditioning instead of the microscopic friction parameters (see Section~\ref{sec:macroscopic_conditioning_experiments}).

\begin{table}[h]
    \centering
    \begin{tabular}{llcc}\toprule
        Name & Description & Range & Sampling \\
        \midrule
        $\alpha\st{HO}$ & Angle of the slope towards the outlet & [0°, 60°] & LHS \\
        $\mu\st{s}$ & Particle sliding friction & [0.05, 1.00] & LHS \\
        $\mu\st{r}$ & Particle rolling friction & [0.00, 0.50] & LHS \\
        $\theta$ & Angle of internal friction & evaluated & - \\
        $\text{ffc}$ & Flow function coefficient & evaluated & - \\\bottomrule
    \end{tabular}
    \vspace{1em}
    \caption{Parameters of the hopper dataset.}
    \label{tab:hopper_dataset_generation_parameter_sampling}
\end{table}

The following macroscopic properties of hopper simulations are of interest to practitioners. 
\begin{itemize}
    \item Peak outflow rate: How many particles exit the hopper within a certain timeframe at most?
    \item Drainage time: How long does it take to empty the hopper?
    \item Residual material: How much material got stuck inside the hopper after full drainage? 
    \item Flow regime: Does the material exhibit mass flow or funnel flow?
    \item Visualization: How does the simulation look like?
    \item Residence time: How long are certain particles inside the simulation?
\end{itemize}

All these macroscopic  properties emerge from the microscopic particle-particle interactions modeled by DEM.
To re-generate these emerging phenomena with NeuralDEM, we train our models to predict three auxiliary fields at each timestep of the simulation. 
First, the \textbf{occupancy} field classifies whether an arbitrary position within a rectangular volume around the hopper is occupied, teaching the model which regions are filled with particles. The occupancy field allows us to calculate an outflow rate by evaluating the occupancy field with the particle positions of the initial packing and subtracting the number of occupied positions from $t_\text{ML}$ and $t_\text{ML} + \Delta t_\text{ML}$. Similarly, we can extract the drainage time by repeatedly checking if positions above the outlet are still occupied and the residual material by evaluating how many of the initial particle positions are occupied after the hopper is drained. Second, the \textbf{transport} field is trained by predicting the initial position of each particle, which corresponds to the cumulative displacement of each particle across all previous timesteps, enabling macroscopic modeling of particle movement over long time horizons. This allows us to model insights into the flow regime and visualize the simulation by evaluating the transport field of occupied positions. Finally, the \textbf{residence} time of each particle models the time it takes each particle to exit the hopper and can be used to easily identify stale regions.

\subsubsection{Multi-branch neural operator architecture for hopper experiments}

The considered hopper simulations exhibit slow and pseudo-steady dynamics, resulting in a state that is well-defined from the scalar input parameters (timestep, particle friction, hopper angle) alone. Therefore, it is neither necessary to encode the previous state nor to have interactions between branches, as the scalar input parameters already provide full information about the state to all branches. Therefore, we opt for a decoder-only architecture consisting of a single transformer block per branch that starts from 32 static learnable latent tokens and prepares them for decoding via a perceiver cross-attention block that takes positional embeddings as queries and uses the latent tokens as keys and values. Both the transformer and the cross-attention block use DiT~\cite{li22dit} modulation to incorporate the scalar parameters (timestep, $\alpha_\text{HO}$, $\mu_s$ and $\mu_r$). The whole model consists of 50M parameters. 

We train models in the hopper setting for 10k updates using a batchsize of 256, a peak learning rate of $10^{-4}$ which is warmed up for 1k updates followed by a decaying cosine schedule afterwards. The loss is summed for all branches and LION~\cite{chen2023lion} is used as optimizer. 

\subsubsection{Flow regime and visualization via occupancy and transport field}

To evaluate the macroscopic modeling of the flow regime inside the hopper, we use the transport field and evaluate it at the occupied positions as defined via the occupancy field. We bin the $z$ coordinate of the transport prediction into 8 bins which results in ``stripes'' of particles at the initial timestep. These stripes then evolve in time as particles flow out. By evaluating the volume of each stripe at every timestep, we get detailed information on the different emerging flow regimes in the hopper.

In the mass flow regime, as shown in Figure~\ref{fig:draining_hopper_transport_visualization:mass}, material flows to the outlet quite uniformly, resulting in a steady flow and first in - first out operation. NeuralDEM can model this behavior accurately, as visualized in Figure~\ref{fig:draining_hopper_transport_visualization:massdiag}, where layer after layer leaves the hopper. 
Contrary, in the funnel flow regime, as shown in Figure~\ref{fig:draining_hopper_transport_visualization:funnel}, the material primarily moves down a funnel above the outlet. This results in a layer inversion, as visualized in Figure~\ref{fig:draining_hopper_transport_visualization:funneldiag}, meaning particles from higher layers overtake particles from the lower layers through the funnel and the layers higher up in the hopper will empty first. These emerging macroscopic phenomena are perfectly modeled by NeuralDEM as well.

Notably, NeuralDEM exclusively models fields, which allows us to evaluate transport and occupancy at arbitrary positions with arbitrary resolution. For example, in the evaluations of this section, we use a tetrahedral grid with 80k cells. Our model can seamlessly make predictions thereof despite seeing only particle positions during training.

\subsubsection{Outflow rate, drainage time, and residual material via occupancy field}

The occupancy field defines the occupied volume of the remaining mass in the draining hopper. Given an initial packing, we can evaluate the occupancy field at the initial positions of each particle and track the number of occupied positions over the whole simulation to then create predictions of outflow rate, drainage time, and residual material. 

\paragraph{Occupancy field.} To define whether or not a position is occupied, we introduce a hyperparameter that defines a radius around each particle position. All positions within this radius are considered occupied whereas positions that are not within the radius of any particle are considered unoccupied.
We choose the radius to be larger than the particle radius to avoid classifying empty spaces between densely packed spherical particles as unoccupied. 

\begin{figure}[h]
    \centering
    \begin{subfigure}{0.33\textwidth}
        \centering
        \includegraphics[width=\linewidth]{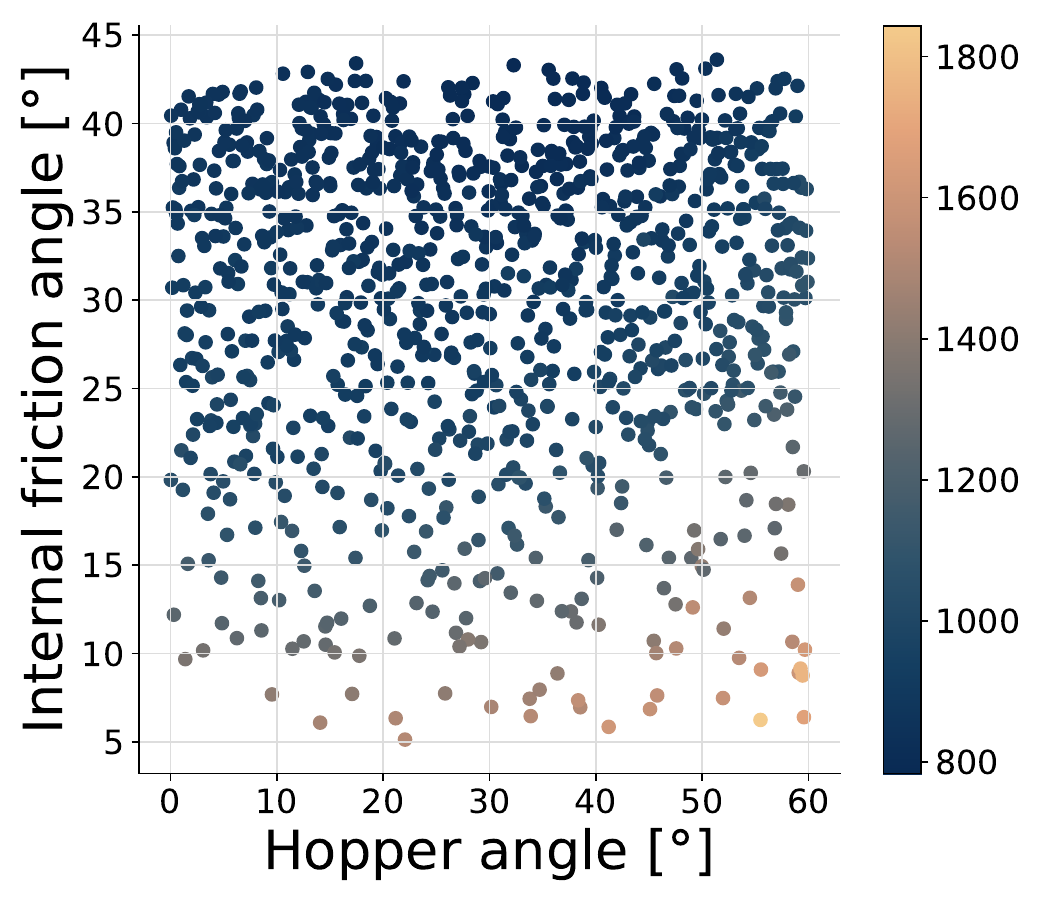}
        \caption{ Outflow rate }
        \label{fig:hopper_draining_angles2d_over_outflow}
    \end{subfigure}%
    \hfill
    \begin{subfigure}{0.33\textwidth}
        \centering
        \includegraphics[width=\linewidth]{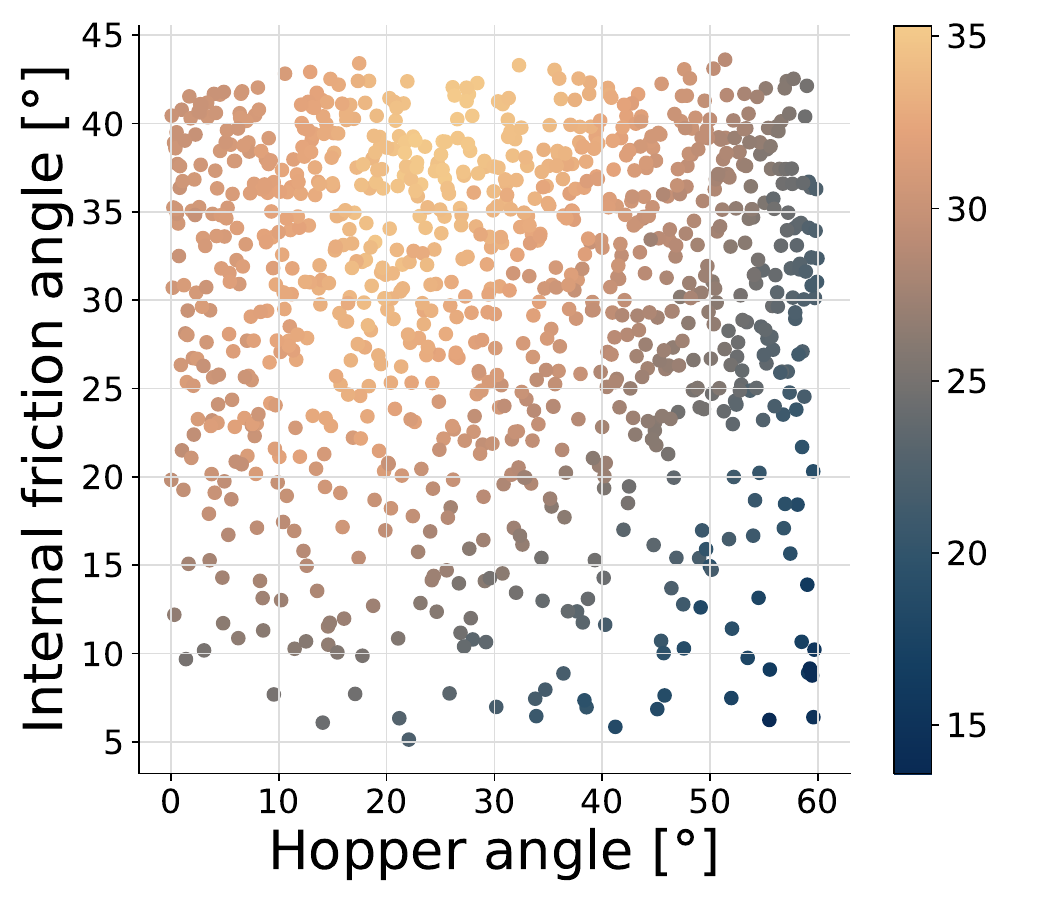}
        \caption{ Drainage time [s] }
        \label{fig:hopper_draining_angles2d_over_simdur}
    \end{subfigure}%
    \hfill
    \begin{subfigure}{0.33\textwidth}
        \centering
        \includegraphics[width=\linewidth]{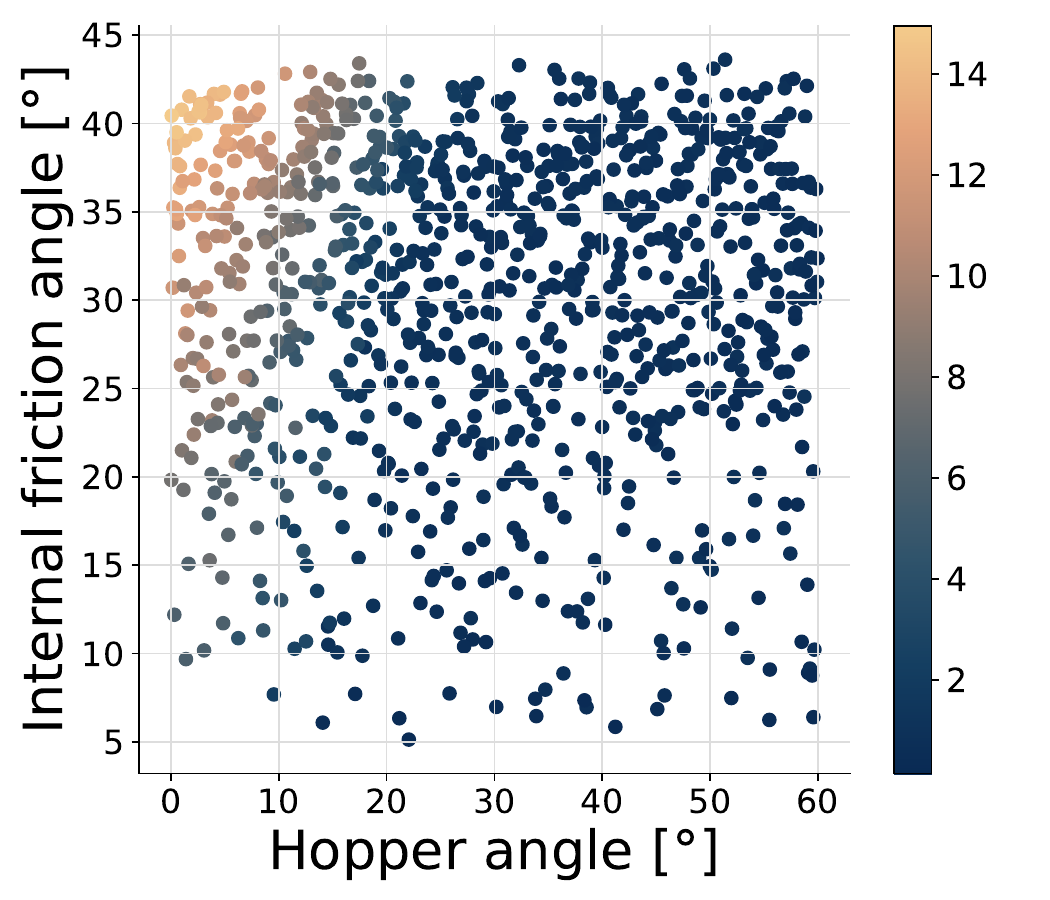}
        \caption{ Residual volume [\%] }
        \label{fig:hopper_draining_angles2d_over_residual}
    \end{subfigure}
    \caption{ Distribution of macroscopic measurements over the whole DEM generated hopper dataset. Combinations of different geometry and friction parameters lead to different simulation behaviors. }
    
\end{figure}

\begin{figure}[h]
    \centering
    \begin{subfigure}{0.3\textwidth}
        \centering
        \includegraphics[width=\linewidth]{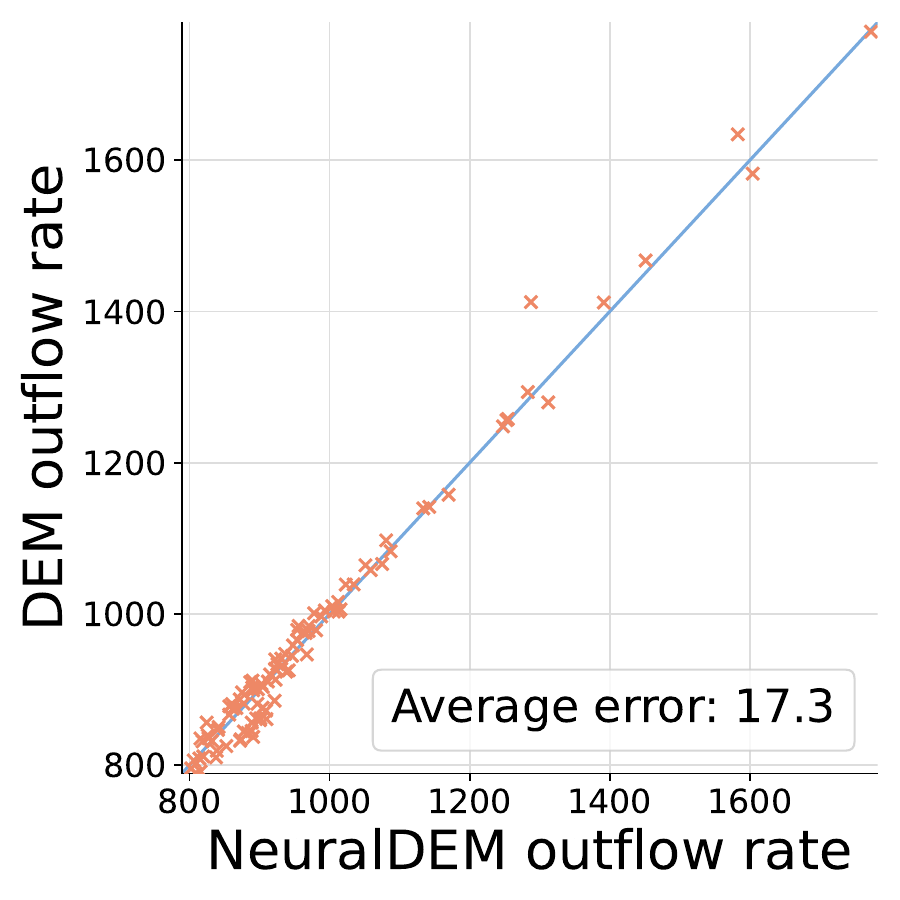}
        \caption{ Outflow rate }
        \label{fig:hopper_draining_prediction_outflow}
    \end{subfigure}%
    \hfill
    \begin{subfigure}{0.3\textwidth}
        \centering
        \includegraphics[width=\linewidth]{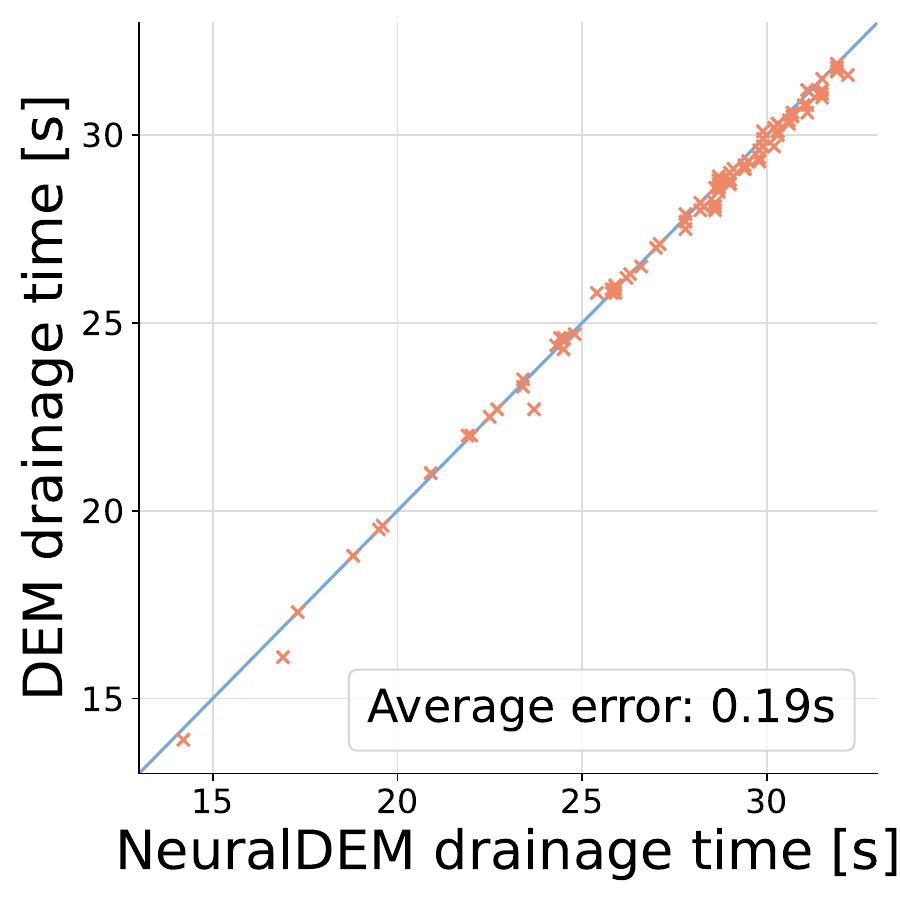}
        \caption{ Drainage time }
        \label{fig:hopper_draining_prediction_drainage}
    \end{subfigure}%
    \hfill
    \begin{subfigure}{0.3\textwidth}
        \centering
        \includegraphics[width=\linewidth]{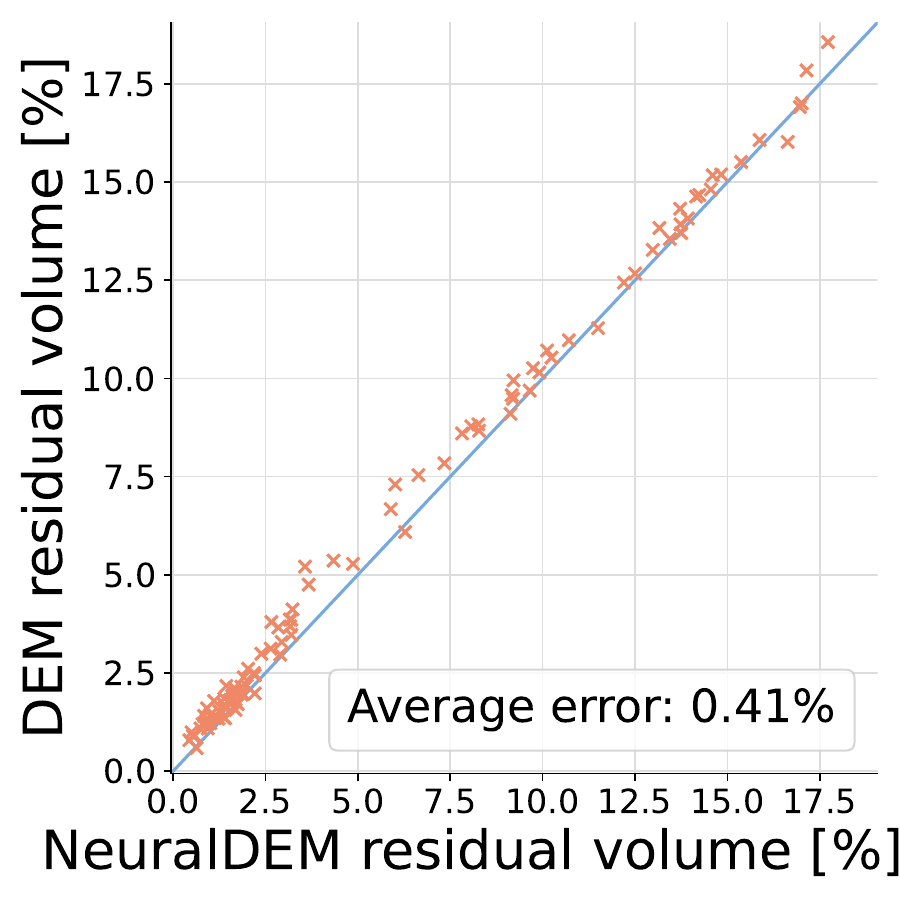}
        \caption{ Residual volume }
        \label{fig:hopper_draining_prediction_residual}
    \end{subfigure}
    \caption{ Macroscopic simulation insights from the predicted occupancy field. The NeuralDEM model can accurately capture macroscopic measurements from the learned dynamics. }
    \label{fig:hopper_draining_predictions_occupancy}
\end{figure}

\paragraph{Outflow rate.} The outflow rate can simply be calculated by subtracting the number of occupied positions at time $t$ from that one timestep later at $t+\Delta t\st{ML}$. To avoid fluctuations due to the burn-in and ending phase of the simulation, we calculate the outflow rate as the average outflow starting from timestep $50$ (\SI{5}{\second}) over a $100$ timestep (\SI{10}{\second}) duration and normalize it by the initial particle count. The ground truth over the whole dataset is visualized in Figure~\ref{fig:hopper_draining_angles2d_over_outflow} and respective NeuralDEM predictions are shown in Figure~\ref{fig:hopper_draining_prediction_outflow}.

\paragraph{Drainage time.} We consider the hopper to be ``drained'' by specifying a threshold of particles that are located above the outlet (``are falling down'') for both classical DEM and NeuralDEM generated trajectories. This definition is necessary as the material gets stuck on the outlet slope if it is very cohesive or if the slope is flat. To evaluate the drainage time of the NeuralDEM model, we query the occupancy field at the particle positions of the initial packing above the outlet until the number of occupied positions is less than the specified threshold. 
The drainage time obtained when running numerical DEM simulations is visualized in Figure~\ref{fig:hopper_draining_angles2d_over_simdur}. Further, the NeuralDEM predicted drainage time is visualized in Figure~\ref{fig:hopper_draining_prediction_drainage}. The NeuralDEM predicted drainage time shows high agreement with the drainage time of the DEM simulation. 
We set the threshold of particles above the outlet to 64 whereas other thresholds like 32 or 128 result in almost the exact same behavior.

\paragraph{Residual volume.} Once the hopper is drained, the number of particles that remain in the hopper (due to a flat outlet slope or high friction) defines the residual particle count. As the volume of the hopper fluctuates due to different outlet slopes shrinking/expanding the volume, we normalize the residual particle count by the initial particle count to predict the residual material as a percentage of the total particle count. Figure~\ref{fig:hopper_draining_angles2d_over_residual} visualizes the ground truth (DEM) values and Figure~\ref{fig:hopper_draining_prediction_residual} the NeuralDEM predictions.

\subsubsection{Residence time}

\begin{figure}[!htp]
    \centering
    \begin{subfigure}{0.5\textwidth}
        \centering
        \begin{tabular}{m{0.32\textwidth} m{0.32\textwidth} m{0.2\textwidth}}

        \parbox[c]{0.32\textwidth}{\centering \footnotesize Ground truth DEM simulation}
         &
        \parbox[c]{0.32\textwidth}{\centering \footnotesize NeuralDEM prediction}  \\
        \includegraphics[width=\linewidth,trim={16cm 0cm 9cm 0cm},clip]{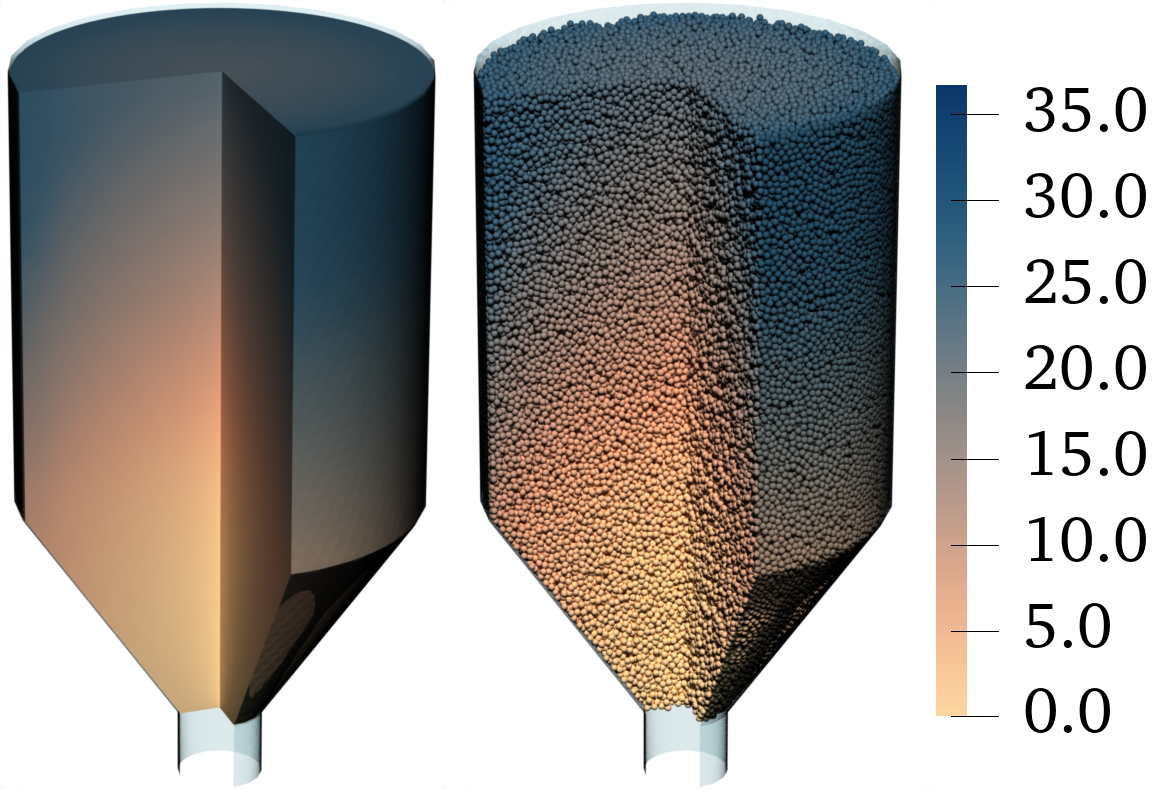} & 
        \includegraphics[width=\linewidth,trim={0cm 0cm 25cm 0cm},clip]{images/Draining_Hopper_Flow/res/res122.png} &

        \includegraphics[width=\linewidth,trim={1cm 0cm 1cm 0cm}]{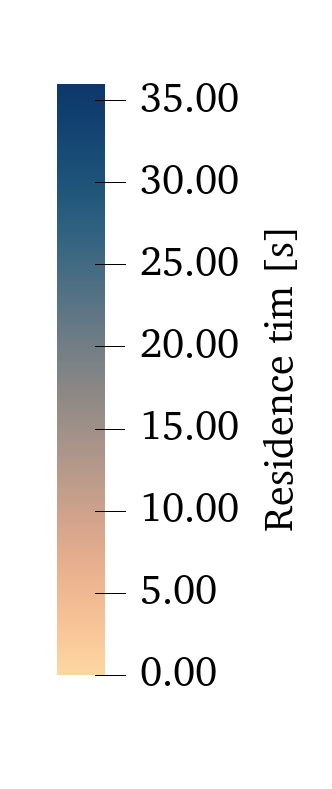}
        \end{tabular}
        \caption{Mass flow regime}
    \end{subfigure}%
    \begin{subfigure}{0.5\textwidth}
       \centering
        \begin{tabular}{m{0.32\textwidth} m{0.32\textwidth} m{0.2\textwidth}}

        \parbox[c]{0.32\textwidth}{\centering \footnotesize Ground truth DEM simulation}
         &
        \parbox[c]{0.32\textwidth}{\centering \footnotesize NeuralDEM prediction}  \\
        
        \includegraphics[width=\linewidth,trim={16cm 0cm 9cm 0cm},clip]{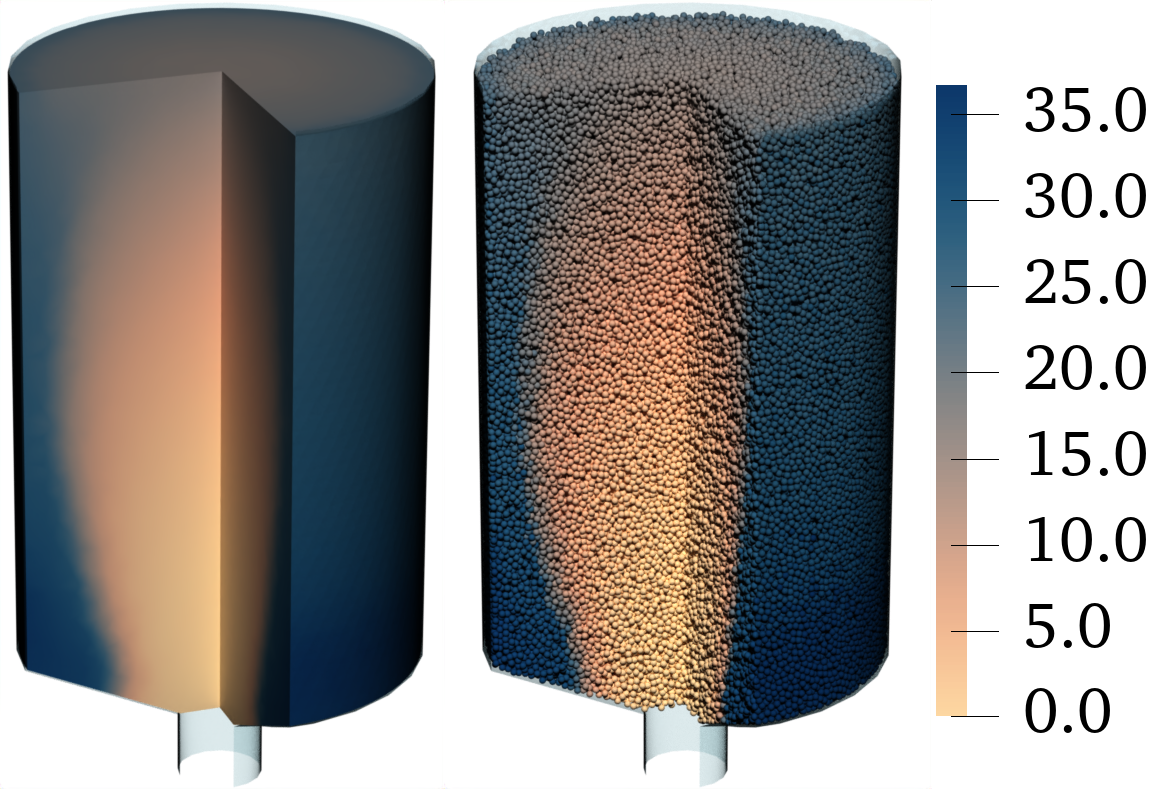} & 
        \includegraphics[width=\linewidth,trim={0cm 0cm 25cm 0cm},clip]{images/Draining_Hopper_Flow/res/res300.png} &

        \includegraphics[width=\linewidth,trim={1cm 0cm 1cm 0cm}]{images/Draining_Hopper_Flow/res/res35.png}
        \\
        \end{tabular}

        \caption{Funnel flow regime}
    \end{subfigure}%
    
    \caption{
    Visualization of a NeuralDEM generated material residence time vs the DEM simulation for (a) mass flow regime and (b) funnel flow regime. The darker a region is the longer the material in that spot stays inside the hopper. While the hopper in the mass flow regime shows a smooth transition from bottom to top, the one with funnel flow shows clear stagnation zone in the bottom with high residence time. Both cases are accurately predicted by Neural DEM. }
    \label{fig:hopper_draining_prediction_residence}
\end{figure}

Another interesting quantity that emerges at the macroscopic level is residence time, i.e., how long it takes for each particle to exit the hopper. 
Therefore, we predict the number of timesteps that each particle resides within the hopper.
Visualizing the residence time at the initial timestep can identify stale regions and can also be used to characterizes the flow regime. We show the initial frame of the residence time prediction for different flow regimes and stale regions in Figure~\ref{fig:hopper_draining_prediction_residence}.

\subsubsection{Generalization capabilities}

\begin{figure}[!htb]
    \centering
    \begin{subfigure}{0.33\textwidth}
        \centering
        \includegraphics[width=\linewidth]{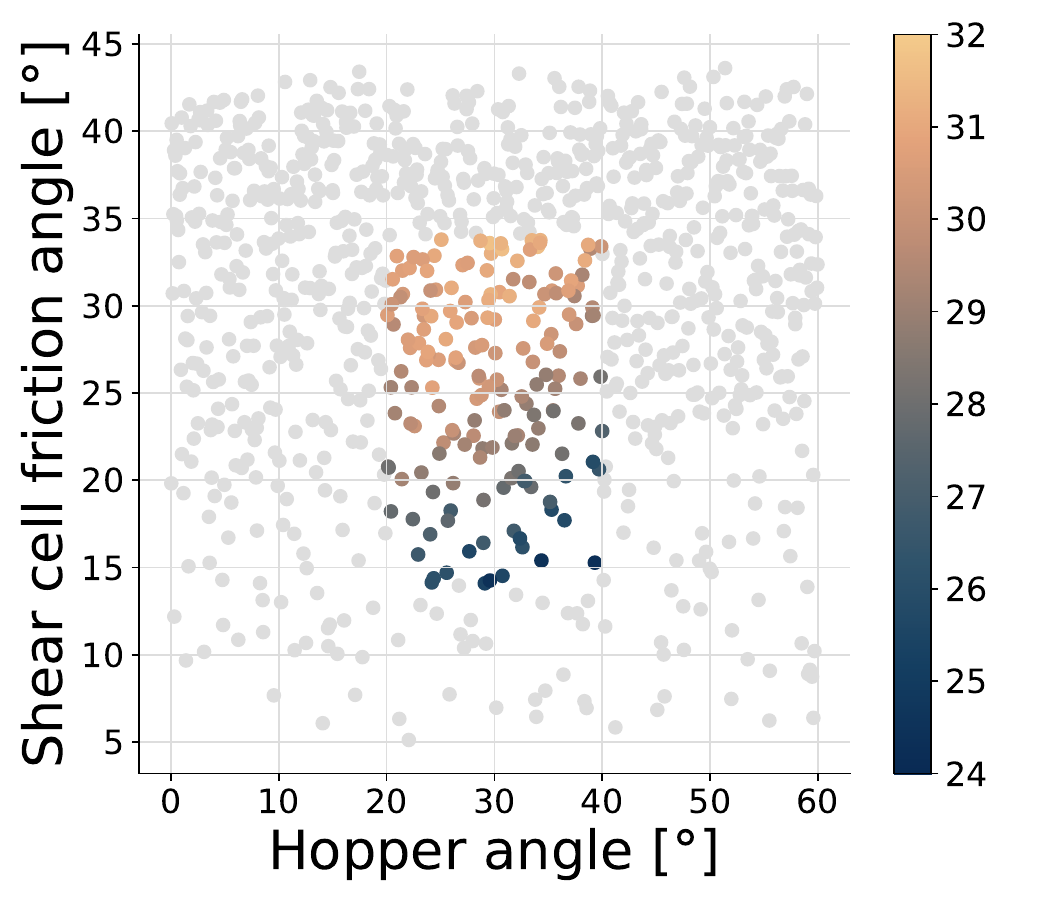}
        \caption{ DEM drainage time [s] }
    \end{subfigure}%
    \hfill
    \begin{subfigure}{0.33\textwidth}
        \centering
        \includegraphics[width=\linewidth]{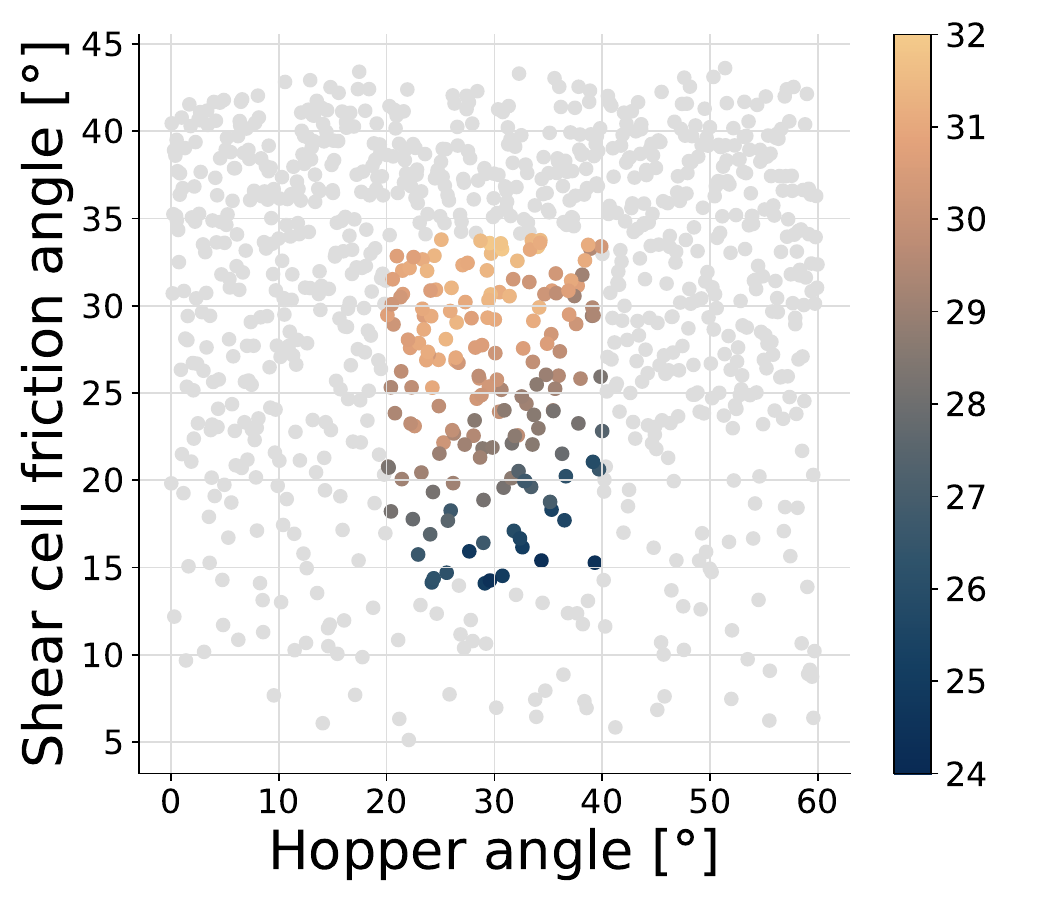}
        \caption{ NeuralDEM drainage time [s] }
    \end{subfigure}%
    \hfill
    \begin{subfigure}{0.33\textwidth}
        \centering
        \includegraphics[width=\linewidth]{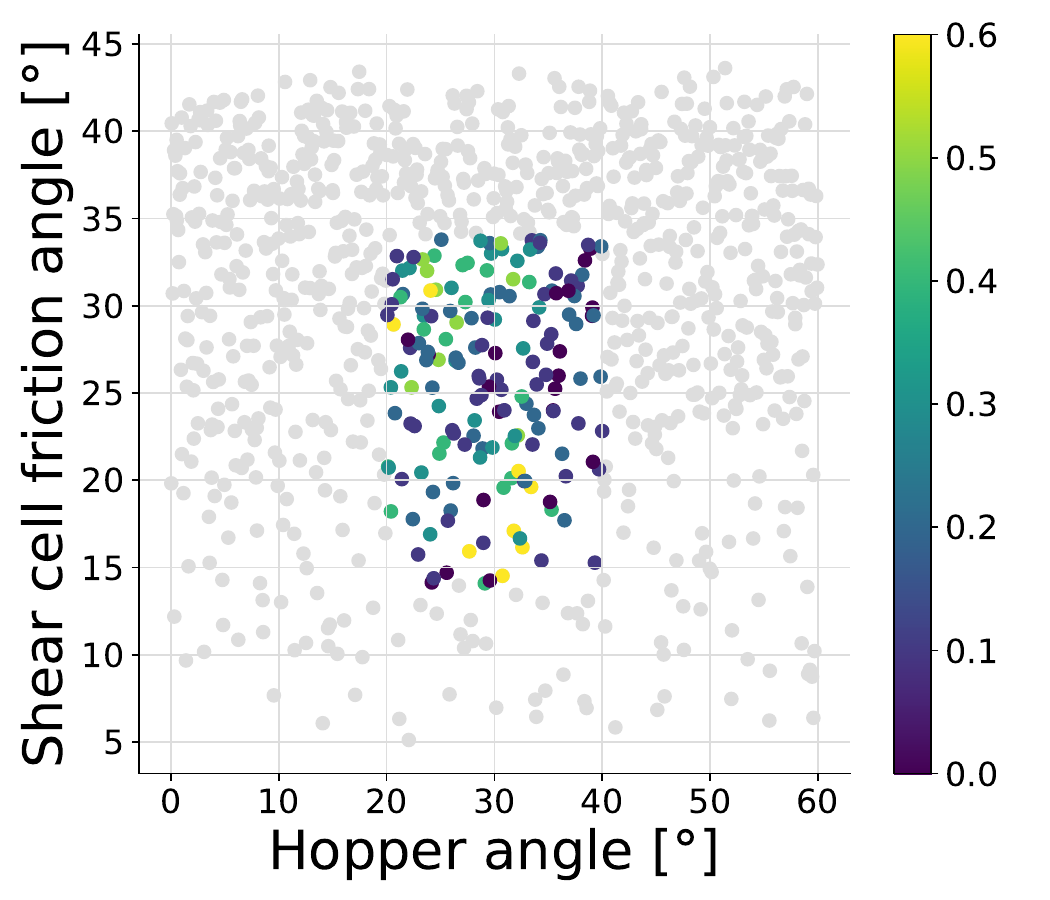}
        \caption{ Absolute error [s] }
    \end{subfigure}
    \caption{Generalization to DEM simulation settings that are outside the range seen during training. We exclude all settings in a range of 20 degrees hopper angles and internal friction angles from the training set and evaluate drainage time on the excluded settings. The NeuralDEM model makes reasonable predictions despite lacking those parameter combinations during training. For reference, the average absolute error is $\SI{0.19}{\second}$ when randomly splitting the data into train/test splits (Figure~\ref{fig:hopper_draining_prediction_drainage}). Gray dots indicate training data and colored dots denote simulations used for evaluation only. }
    \label{fig:hopper_draining_angles2d_over_simdur_generalization}
\end{figure}

To investigate the generalization capabilities of NeuralDEM, we split the dataset by excluding a parameter range of 20 degrees from the training set for testing. In this setting, the NeuralDEM model is evaluated on parameter combinations that are far away from the ones seen during training. Nevertheless, the NeuralDEM model makes reasonable predictions as shown in Figure~\ref{fig:hopper_draining_angles2d_over_simdur_generalization}.

\subsubsection{Macroscopic parameter conditioning} \label{sec:macroscopic_conditioning_experiments}

A common use case in the industry is that given some material, one wants to simulate the material in, e.g., a hopper to get insights into its behavior or flow dynamics. However, DEM solvers require a precise specification of the microscopic material parameters, which are often unknown.
In order to run a DEM simulation of a material with unknown microscopic parameters, those parameters first have to be inferred via, e.g., a calibration procedure~\cite{coetzee2017} before the DEM simulation of the material can be run. In the case of our hopper simulation, one would need to infer the values of particle sliding and rolling friction.

Instead, the flexible conditioning methodology (as described in Section~\ref{sec:method_conditioning}) allows NeuralDEM to condition on macroscopic parameters instead of microscopic ones. To this end, we evaluate the internal friction angle and flow function coefficient in a separate shear cell simulation to characterize the material macroscopically. We then use the macroscopic internal friction angle and flow function coefficient ($\theta$ and ffc) as conditioning instead of the microscopic sliding and rolling friction parameters ($\mu\st{s}$ and $\mu\st{r}$). This allows our model to simulate any given material by simply characterizing it in a shear cell and using the resulting macroscopic friction parameters as input to the model, without needing particle sliding/rolling friction. 

\begin{figure}[!htb]
\captionsetup[subfigure]{justification=centering}
\centering\begin{subfigure}[b]{0.27\linewidth}
\includegraphics[width=0.8\linewidth]{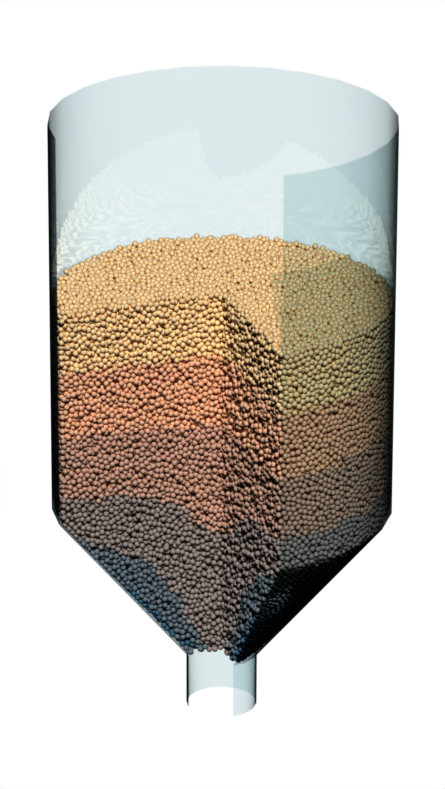}
\caption{Ground truth \\ DEM simulation \\ ~}
\end{subfigure}
\begin{subfigure}[b]{0.27\linewidth}
\includegraphics[width=0.8\linewidth]{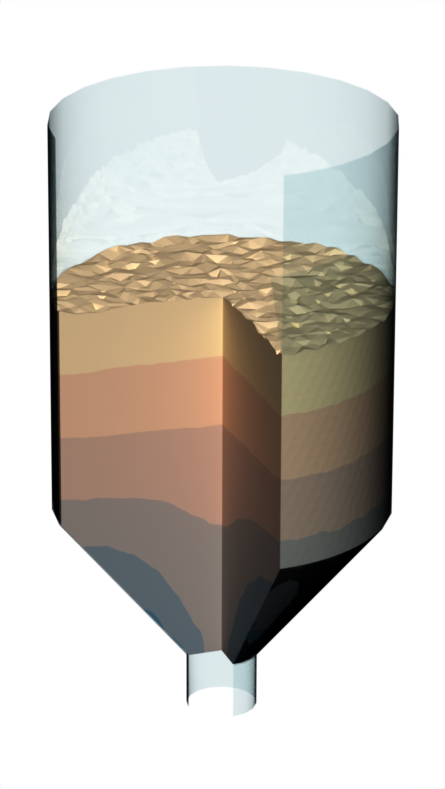}
\caption{NeuralDEM using \\microscopic parameter \\conditioning}
\end{subfigure}
\begin{subfigure}[b]{0.27\linewidth}
\includegraphics[width=0.8\linewidth]{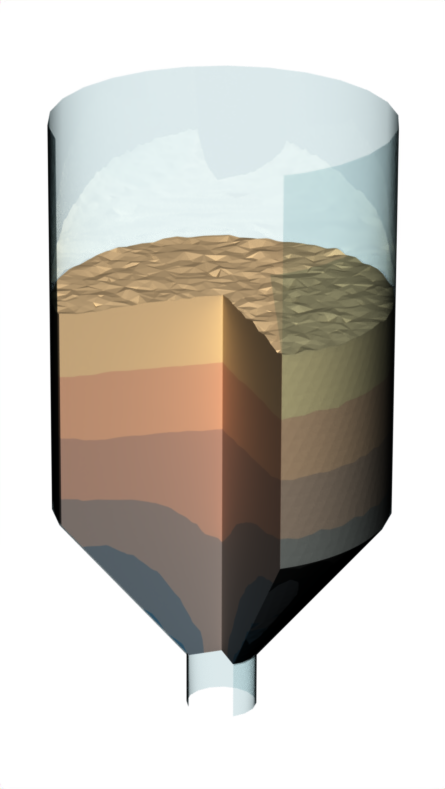}
\caption{NeuralDEM using \\ macroscopic parameter \\
conditioning}
\end{subfigure}

\caption{NeuralDEM transport field prediction using microscopic simulation parameters vs macroscopic material parameters. Conditioning on microscopic sliding and rolling friction parameters ($\mu\st{s}$ and $\mu\st{r}$) reflects the underlying DEM simulation exactly and results in accurate agreement of the transport field.
 However, conditioning only on the measured macroscopic friction angle ($\theta$) and flow function coefficient (ffc) still leads to acceptable accuracy with small deviations. This allows NeuralDEM to work without a calibration routine to get the microscopic parameters on any given material where $\mu\st{s}$ and $\mu\st{r}$ are unknown.}
    
    \label{fig:hopper_draining_conditioning}
\end{figure}

Table~\ref{tab:draining_hopper_macroscopic_conditioning} shows a quantitative comparison with different conditionings. Note that a drop in performance is expected as the microscopic material properties 1:1 reflect the underlying DEM simulation. 
Nevertheless, this drop is within an acceptable range such that model predictions are still useful, as shown in Figure~\ref{fig:hopper_draining_conditioning}.

\begin{table}[!htb]
    \centering
    \resizebox{0.75\textwidth}{!}{
    \begin{tabular}{ccccccccc}\toprule
        \multicolumn{4}{c}{Scalars} & \multicolumn{5}{c}{NeuralDEM predictive performance} \\
        \cmidrule(rl){0-3} \cmidrule(rl){5-9}
        \multirow{2}{*}{$\mu\st{s}$} & \multirow{2}{*}{$\mu\st{r}$} & \multirow{2}{*}{$\theta$} & \multirow{2}{*}{ffc} & Drainage & Residual & Outflow & Transport & Residence  \\
        & &  &  & error [s] & error [\%] & error & MSE [1e-2] & MSE [1e-3] \\
        \midrule
        \cmark & \cmark & \xmark & \xmark & 0.19 & 0.41 & 17.3 & 0.72 & 1.59 \\
        \xmark & \xmark & \cmark & \xmark & 0.41 & 0.64 & 44.6 & 2.30 & 4.46 \\
        \xmark & \xmark & \xmark & \cmark & 0.78 & 0.80 & 38.6 & 4.48 & 7.78 \\
        \xmark & \xmark & \cmark & \cmark & 0.40 & 0.66 & 28.4 & 1.96 & 3.82 \\
        \midrule
        \cmark & \cmark & \cmark & \cmark & 0.23 & 0.62 & 17.2 & 0.74 & 1.46 \\\bottomrule
    \end{tabular}}
    \vspace{1em}
    \caption{
    NeuralDEM performance metrics using different scalar conditions in the model. Conditioning on microscopic sliding and rolling friction parameters ($\mu\st{s}$ and $\mu\st{r}$) fully specifies the underlying DEM simulation and leads to the best performances. Conditioning only on the measured macroscopic friction angle ($\theta$) and flow function coefficient (ffc) from a shear cell device can obtain reasonable results while not requiring calibration procedures, i.e., $\mu\st{s}$/$\mu\st{r}$ are unknown. The measured scalars from the shear cell device provide no additional information and therefore do not improve model performance.
    }
    \label{tab:draining_hopper_macroscopic_conditioning}
\end{table}

\subsubsection{Refilling operation mode} \label{sec:hopper_refilling}

For neural operator modeling, it is difficult to represent particles that do not exist at the initial timeframe. This is due to the typically stochastic nature of the refilling process, and one would need to accurately model this process during inference in order to avoid unrealistic states, e.g., overlapping particles. However, our field-based modeling paradigm allows NeuralDEM to model the macroscopic behavior of refilled particles without requiring their accurate positions.

\begin{figure}[h]
\captionsetup[subfigure]{justification=centering}
\centering\begin{subfigure}[b]{0.32\linewidth}
\includegraphics[width=\linewidth]{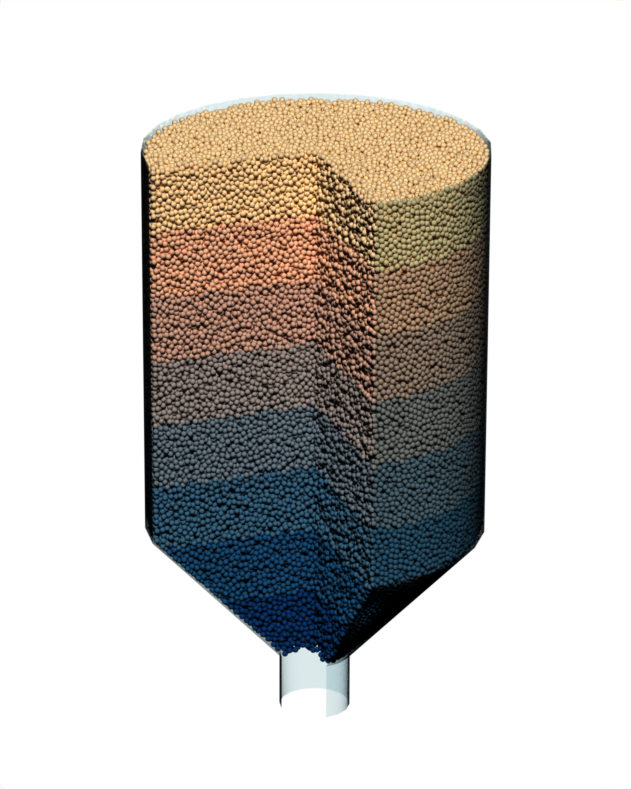}
\caption{Ground truth \\ DEM simulation $t_{ref}$}
\end{subfigure}
\begin{subfigure}[b]{0.32\linewidth}
\includegraphics[width=\linewidth]{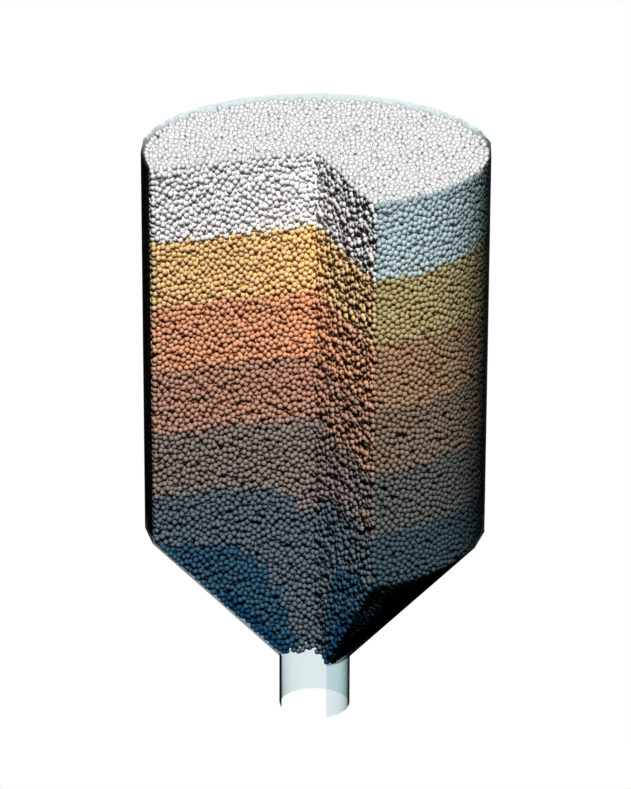}
\caption{Ground truth \\ DEM simulation $t_{ref}+\SI{5}{\second}$}
\end{subfigure}
\begin{subfigure}[b]{0.32\linewidth}
\includegraphics[width=\linewidth]{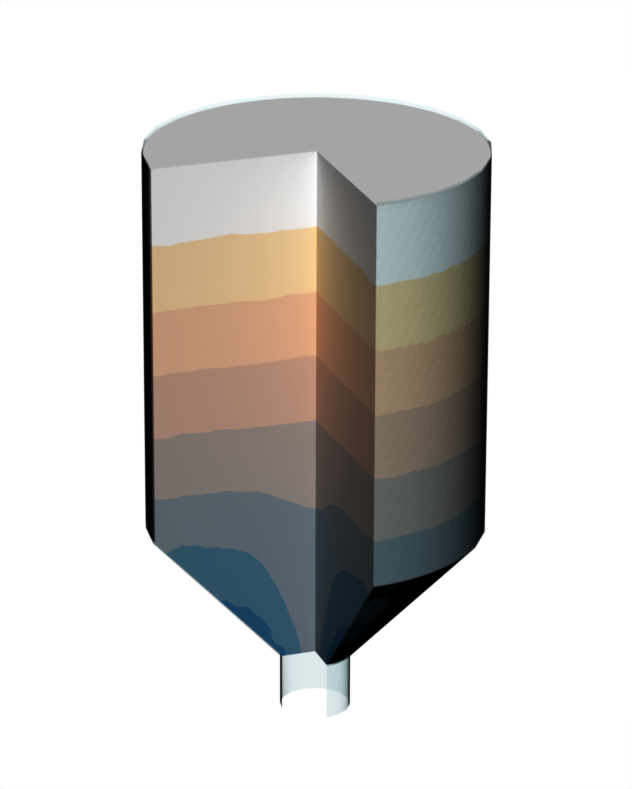}
\caption{NeuralDEM \\ prediction $t_{ref}+\SI{5}{\second}$}
\end{subfigure}
\caption{NeuralDEM transport field prediction for the hopper case in refilling operation mode vs ground truth DEM simulation. The transport field is initialized at a reference time (a). The ground truth of the \SI{5}{\second} evolved field is shown in (b) and the NeuralDEM prediction in (c). Newly Refilled material since is shown in gray. NeuralDEM can accurately model the transport of material in a hopper in refilling operation mode.}
    
\label{fig:hopper_refilling_transport_and_residence}
\end{figure}

To showcase this setting, we train a NeuralDEM model to predict the evolved transport and residence time when continuously refilling new particles into the hopper. As the refilling operation mode can go on indefinitely, the model needs some kind of reference frame to predict ``how did particles move w.r.t.\ the reference frame'' and ``how long particles have been in the hopper w.r.t.\ the reference frame''. To this end, we sample a random past timestep during training and predict transport and residence time w.r.t.\ this past timestep. Figure~\ref{fig:hopper_refilling_transport_and_residence} shows the evolved transport over 50 ML timesteps ($5\si{s}$) in comparison to a ground truth DEM simulation.

The residence time in the hopper can go to infinity when the NeuralDEM trajectory is rolled out for longer. This happens when material is stuck inside the hopper, which the NeuralDEM model cannot predict as it only sees simulations of a fixed maximum duration during training. Conditioning of the residence off-branch to a reference timesteps allows us to periodically reset the reference time during inference and aggregate the residence time predictions before each reset in post-processing.

\subsubsection{Runtime: NeuralDEM enables real-time simulations}\label{sec:runtime_hopper}

Simulating a granular flow of 250k particles through a hopper, with a trajectory spanning $40 \si{\s}$ or 4M numerical timesteps, requires 3 hours on 16 cores of high-performance CPUs when using traditional DEM. In contrast, on a single state-of-the-art GPU, the fastest NeuralDEM inference model faithfully reproduces the physics rollout in just $\SI{1.4}{\second}$. Notably, NeuralDEM requires significantly fewer timesteps compared to the numerical simulation. Further acceleration is achieved through NeuralDEM's field-based output representation, which reduces the number of required output points while still capturing the macroscopic bulk behavior accurately. When processing all 250k output points on a single GPU, NeuralDEM inference takes $\SI{8}{\second}$. Running NeuralDEM on the same 16 CPUs used for the numerical simulation -- notably leveraging the benefits of reduced outputs due to field-based representations -- results in a trajectory rollout of $\SI{41}{\second}$. Those inference times are shorter (GPU) than or in the same order of magnitude (16 CPUs) as the trajectory duration of $\SI{40}{\second}$, highlighting the feasibility of real-time simulations. While traditional DEM could theoretically be further parallelized, for a certain thread number hardware communication bottlenecks present a limiting factor, and thus further underlines the potential of NeuralDEM.

\subsection{Fluidized bed reactors}

As second industrial particulate flow machinery, we consider the highly dynamic system of fluidized bed reactors, where particles and air interact, necessitating a multi-physics modeling, see Figure~\ref{fig:fb_geometry}. These processes are described through a coupled CFD-DEM simulation, where DEM is used to handle particles, while the surrounding fluid is simulated by CFD.
Numerically, the reactor is filled with 500k particles, and the air that is uniformly pushed into the reactor from the bottom is modeled on a grid of 160k hexahedral cells.
The following numerical experiments are carried out on a fixed-geometry reactor with varying inlet velocities, and we evaluate short-term behavior and long-term statistics. In total, $76$ difference inlet velocities sampled uniformly from \SI{0.337}{\meter\per\second} to \SI{0.842}{\meter\per\second} were used. Additionally, $6$ different initial random particle packings per inlet velocity were sampled, resulting in $456$ CFD-DEM trajectories. The physical duration of one DEM simulation amounts to $\SI{5}{s}$, comprising 2M DEM timesteps and 20k CFD timesteps.

To train the NeuralDEM model, each trajectory is sub-sampled to $300$ timesteps, starting from $\SI{2}{s}$ of the original simulation and sampling every $\Delta t_\text{ML}=\SI{0.01}{\second}$. The classical solver requires a much finer time resolution where $\Delta t_\text{ML}=4000 \Delta t_\text{DEM} = 40 \Delta t_\text{CFD}$. We select $60$ inlet velocities at random for the training set and the remaining $16$ are left for validation, thus obtaining a training set of $360$ trajectories and a validation set of $96$. For testing the long rollout performance of NeuralDEM, we further generate $4$ sequences with a physical duration of $\SI{28}{s}$ and different inlet velocities, where new random initial packings are applied. Those trajectories result in sequences of $2800$ timesteps.

\subsubsection{Multi-branch neural operator architecture for fluidized bed reactor experiments}
Modeling a fluidized bed reactor requires interaction between fluid and particles. To this end, we utilize both grid data and particle data as input to the model. 
As described in Section \ref{sec:flexible_model_architecture}, we use specialized designs for different physics phases. Namely, ViT~\cite{dosovitsky2021vit} patch embedding for grid data and UPT supernode pooling~\cite{alkin2024upt} for particle data.
The number of supernodes and patches can be changed after training, enabling stable training and a flexible choices for generating trajectory rollouts. 
Rollouts are carried out in autoregressive fashion, where each timestep is used as input for the next timestep. 
During training, the model receives data from a random timestep $t$, is conditioned on the inlet velocity, and is supervised with the quantities from timestep $t+\Delta t_{\text{ML}}$. The inlet velocity conditioning is performed analogously to the hopper setting, with a DiT~\cite{li22dit} modulation. The particle mixing is modeled as an off-branch quantity. 

The final model has roughly 850M trainable parameters in total where we use a standard DiT~\cite{li22dit} conditioning mechanism which adds a lot of parameters that do not contribute a significant amount of FLOPS to the model. We use 12 multi-branch transformer blocks in total with hidden dimension 768, where each of the 3 branches would correspond to a ViT-Base~\citep{dosovitsky2021vit} (86M parameters) if no DiT modulation was applied.
The model processes sequences with length of 2048 tokens for the particle displacement, 2500 for the fluid velocity and particle solidfraction, and 2500 for the particle mixing, both with $4\times 4\times 4$ patches. The model is trained 126k updates using a batch size of 128, a peak learning rate of $4\times 10^{-5}$ which is warmed up for 13k updates followed by a decaying cosine schedule to $10^{-7}$ using LION~\cite{chen2023lion} as optimizer with 0.5 as weight decay. The loss is summed over all branches.

\begin{figure}[htb!]
\centering
\begin{subfigure}[b]{0.8\linewidth}
\resizebox{\linewidth}{!}{
\begin{tabular}{c c c c c c c c}
\multicolumn{3}{c}{\huge NeuralDEM prediction}&&\multicolumn{3}{c}{\huge Ground truth CFD-DEM simulation}&\\
 \includegraphics[width=.3\linewidth ,trim={0cm 0cm 0cm 0cm},clip]{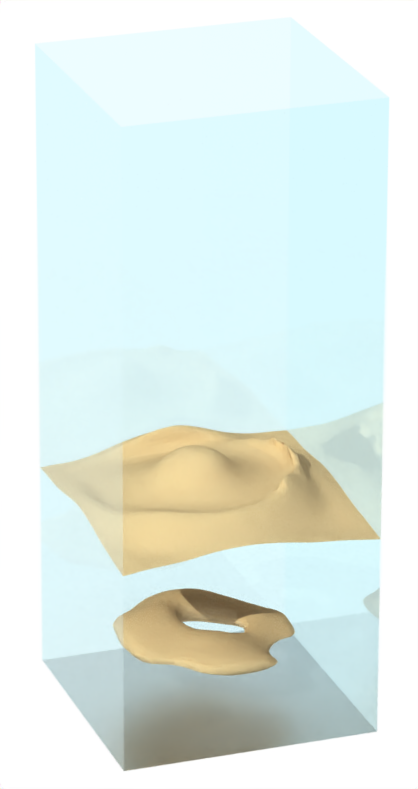} &
\includegraphics[width=.3\linewidth ,trim={0cm 0cm 0cm 0cm},clip]{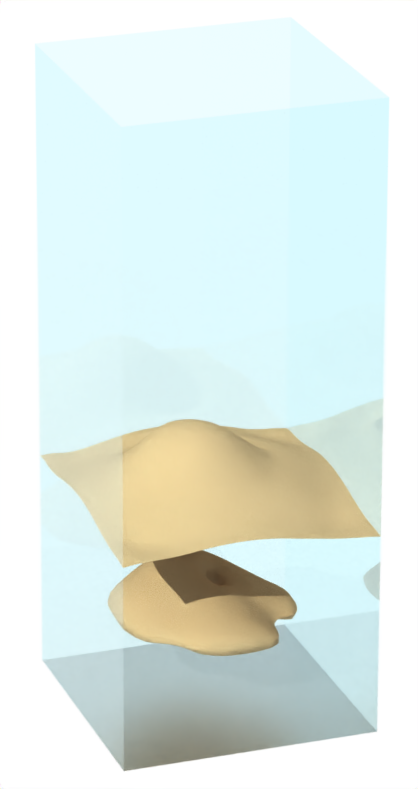} &
\includegraphics[width=.3\linewidth ,trim={0cm 0cm 0cm 0cm},clip]{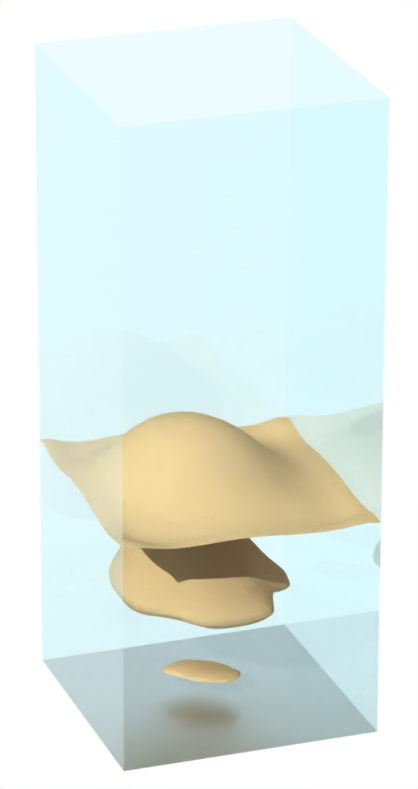}
 & \hspace*{0.01\linewidth}&
 \includegraphics[width=.3\linewidth ,trim={0cm 0cm 0cm 0cm},clip]{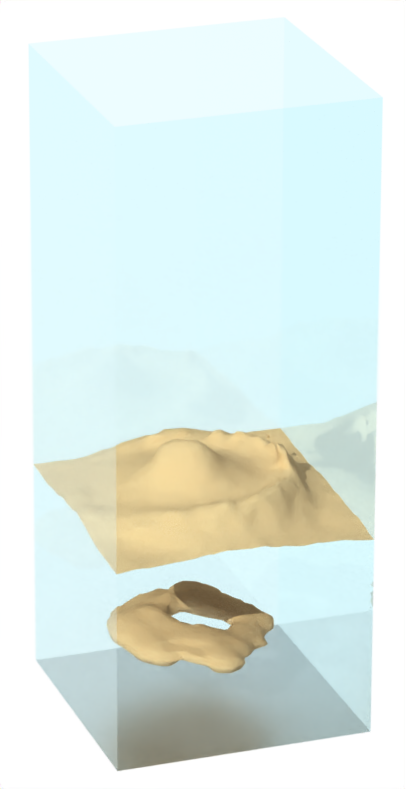} & 
\includegraphics[width=.3\linewidth ,trim={0cm 0cm 0cm 0cm},clip]{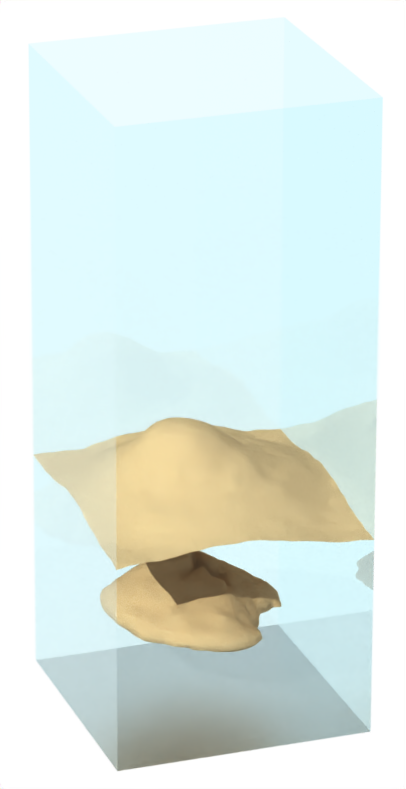} &
\includegraphics[width=.3\linewidth ,trim={0cm 0cm 0cm 0cm},clip]{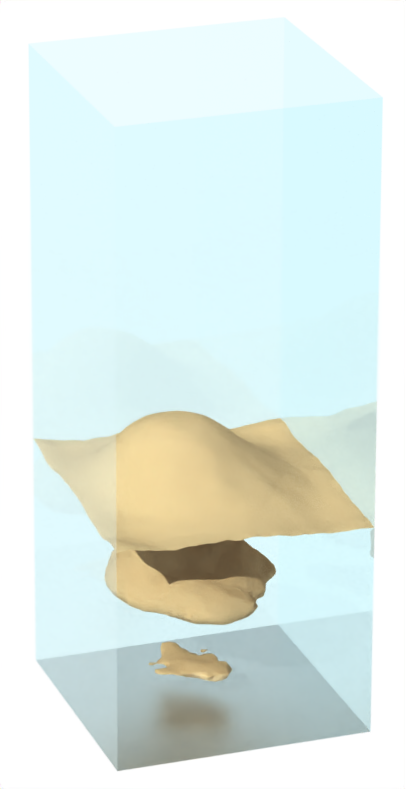} \\
  \includegraphics[width=.3\linewidth ,trim={0cm 0cm 0cm 0cm},clip]{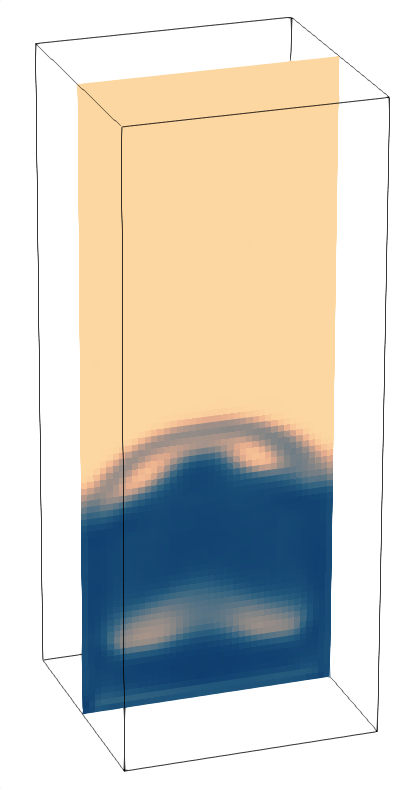} &
\includegraphics[width=.3\linewidth ,trim={0cm 0cm 0cm 0cm},clip]{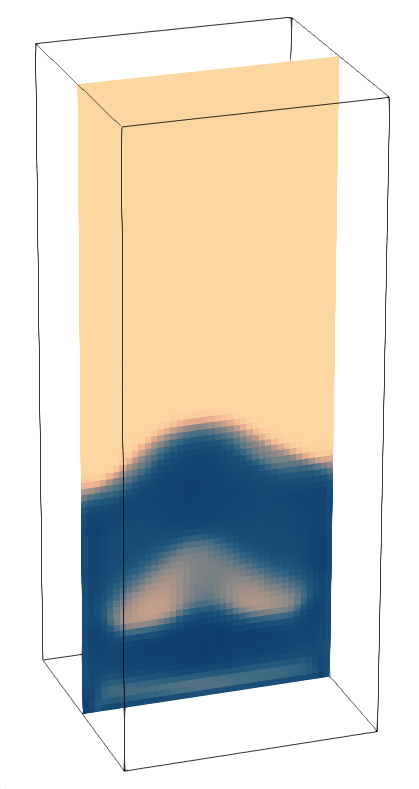} &
\includegraphics[width=.3\linewidth ,trim={0cm 0cm 0cm 0cm},clip]{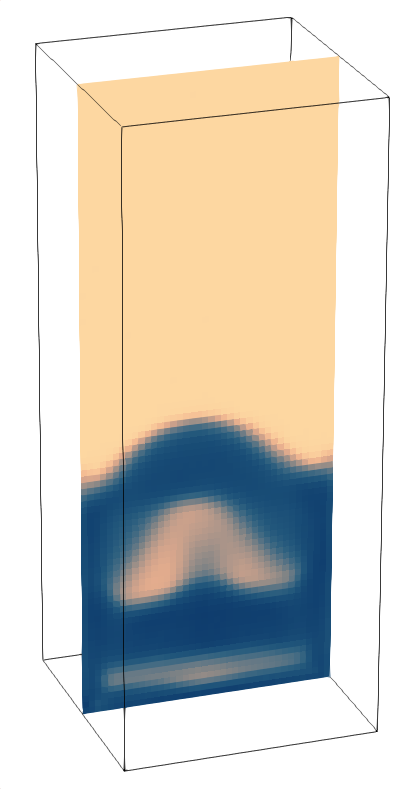}
 & \hspace*{0.01\linewidth}&
 \includegraphics[width=.3\linewidth ,trim={0cm 0cm 0cm 0cm},clip]{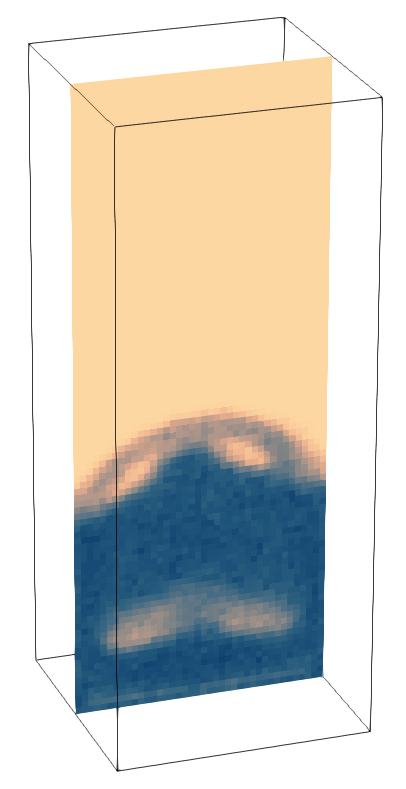} & 
\includegraphics[width=.3\linewidth ,trim={0cm 0cm 0cm 0cm},clip]{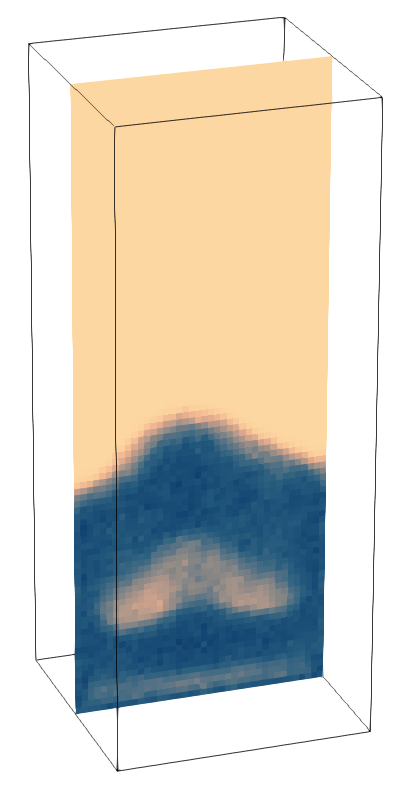} &
\includegraphics[width=.3\linewidth ,trim={0cm 0cm 0cm 0cm},clip]{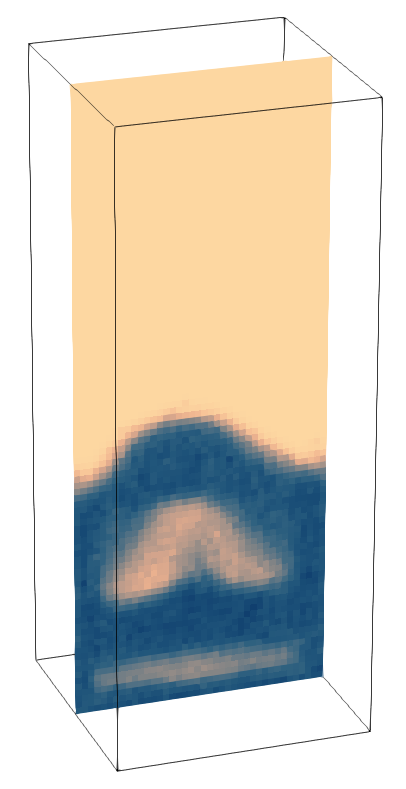}&
\includegraphics[width=.25\linewidth ,trim={0cm 0cm 0cm 0cm},clip]{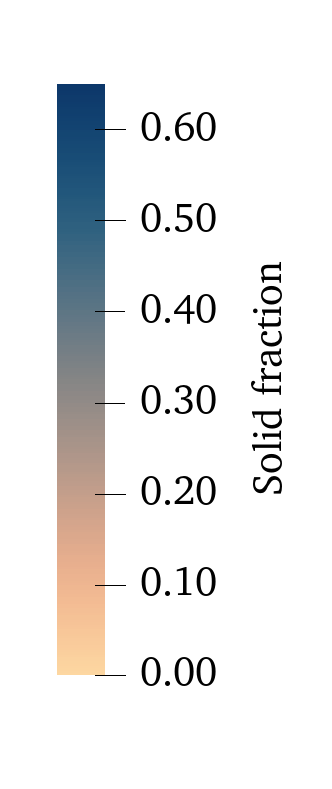}
 \\
  \includegraphics[width=.3\linewidth ,trim={0cm 0cm 0cm 0cm},clip]{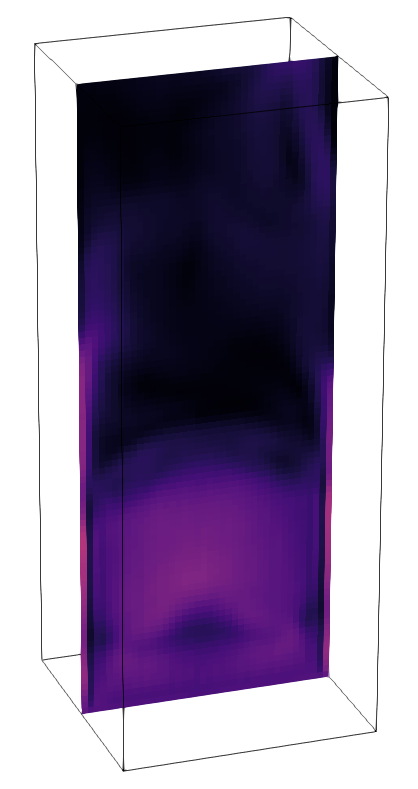} &
\includegraphics[width=.3\linewidth ,trim={0cm 0cm 0cm 0cm},clip]{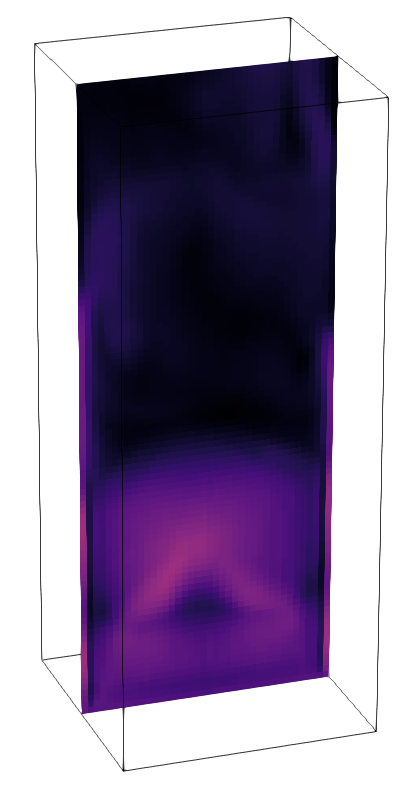} &
\includegraphics[width=.3\linewidth ,trim={0cm 0cm 0cm 0cm},clip]{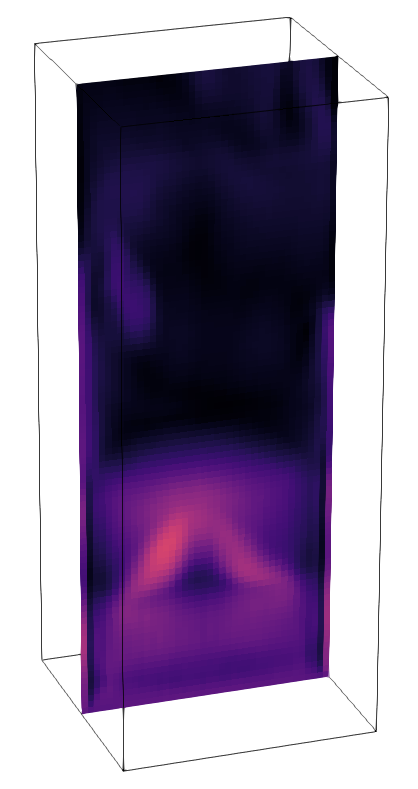}
 & \hspace*{0.01\linewidth}&
 \includegraphics[width=.3\linewidth ,trim={0cm 0cm 0cm 0cm},clip]{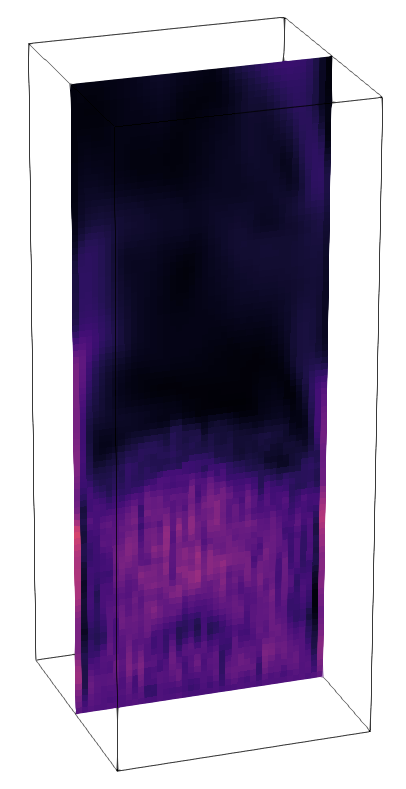} & 
\includegraphics[width=.3\linewidth ,trim={0cm 0cm 0cm 0cm},clip]{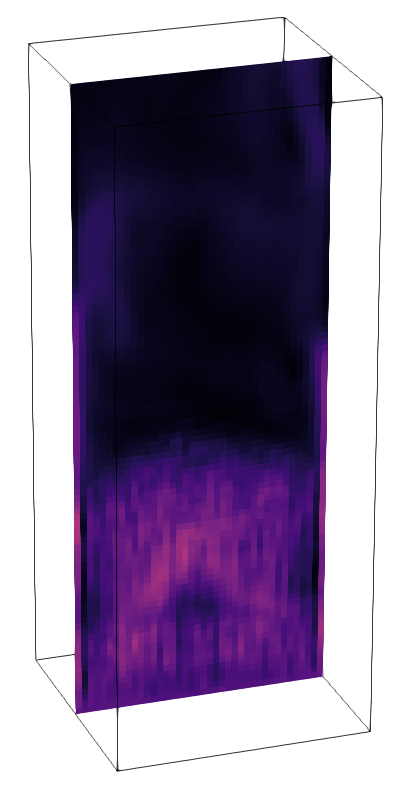} &
\includegraphics[width=.3\linewidth ,trim={0cm 0cm 0cm 0cm},clip]{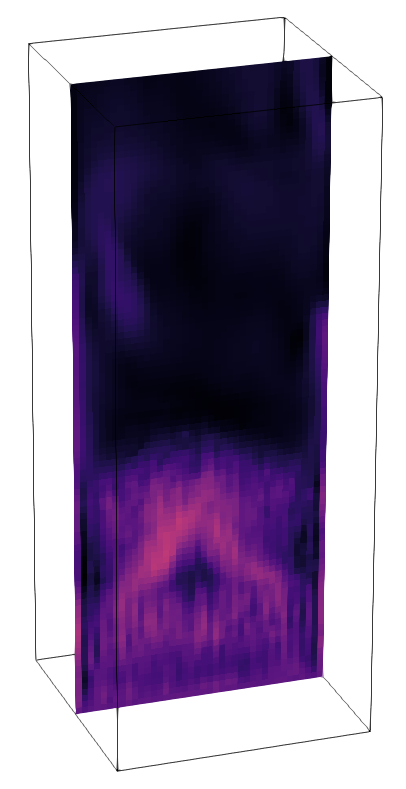}&
\includegraphics[width=.25\linewidth ,trim={0cm 0cm 0cm 0cm},clip]{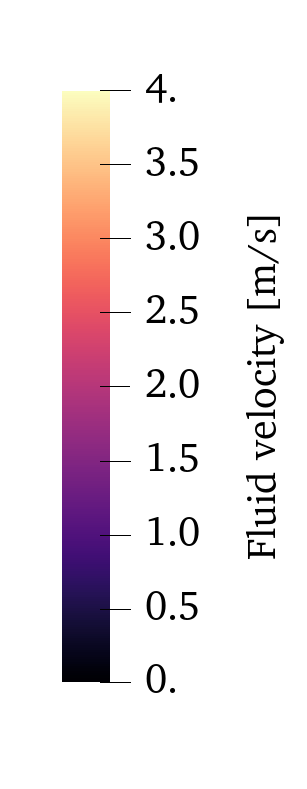}
 \\

 \end{tabular}
} 

\caption{Low inlet velocity}
\end{subfigure}
\par\bigskip

\begin{subfigure}[!htb]{0.8\linewidth}
\resizebox{\linewidth}{!}{
\begin{tabular}{c c c c c c c c}
\multicolumn{3}{c}{\huge NeuralDEM prediction}&&\multicolumn{3}{c}{\huge Ground truth CFD-DEM simulation}\\
 \includegraphics[width=.3\linewidth ,trim={0cm 0cm 0cm 0cm},clip]{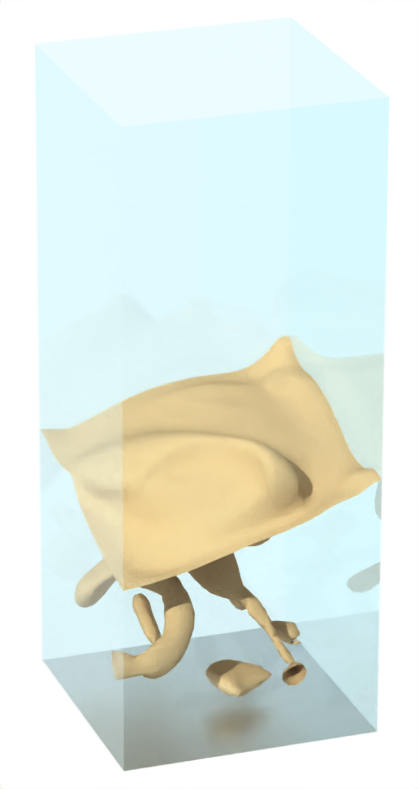} &
\includegraphics[width=.3\linewidth ,trim={0cm 0cm 0cm 0cm},clip]{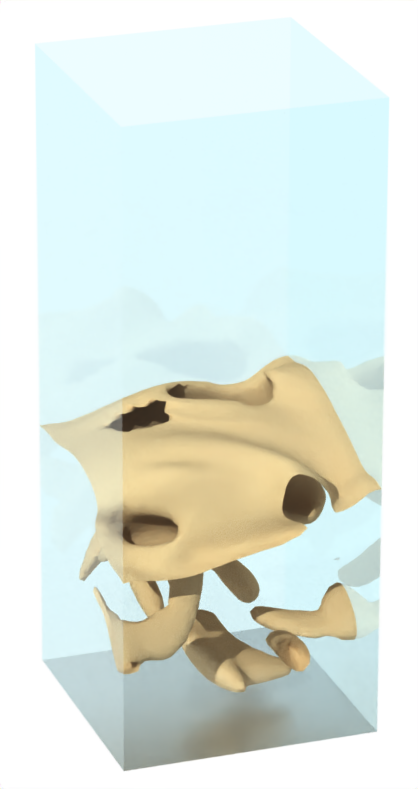} &
\includegraphics[width=.3\linewidth ,trim={0cm 0cm 0cm 0cm},clip]{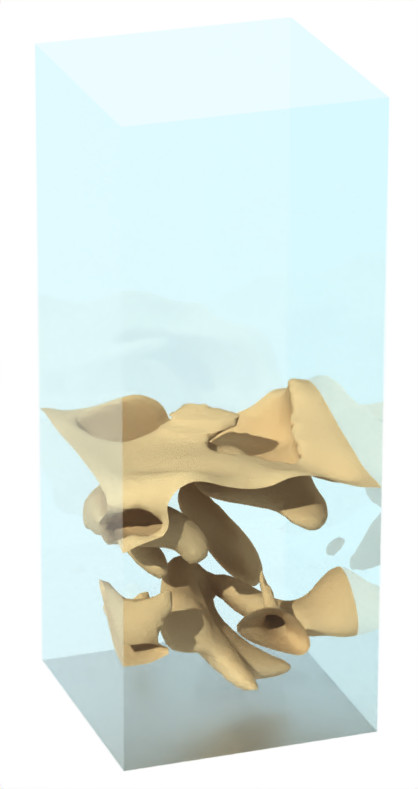}
 & \hspace*{0.1\linewidth}&
 \includegraphics[width=.3\linewidth ,trim={0cm 0cm 0cm 0cm},clip]{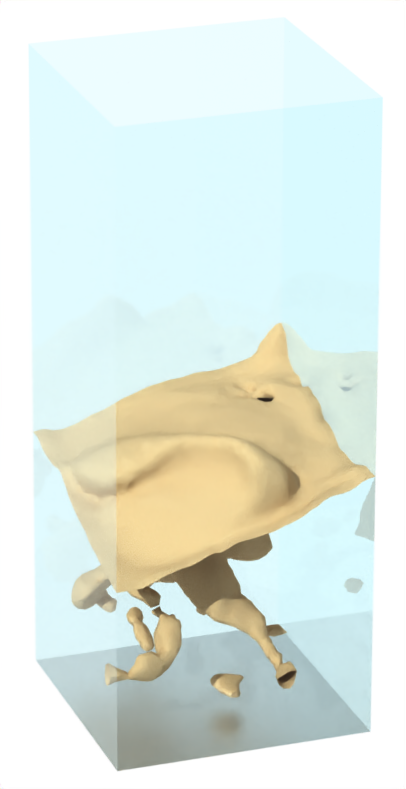} & 
\includegraphics[width=.3\linewidth ,trim={0cm 0cm 0cm 0cm},clip]{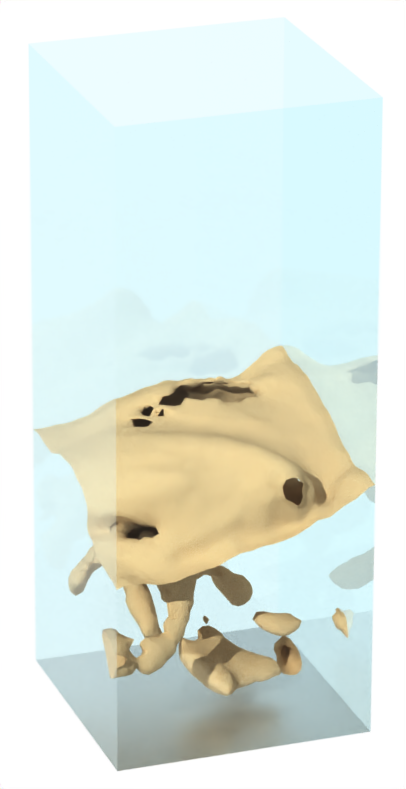} &
\includegraphics[width=.3\linewidth ,trim={0cm 0cm 0cm 0cm},clip]{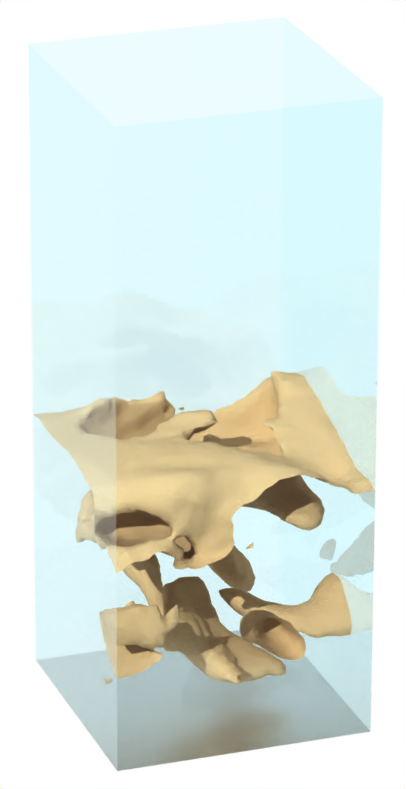}
 \\
  \includegraphics[width=.3\linewidth ,trim={0cm 0cm 0cm 0cm},clip]{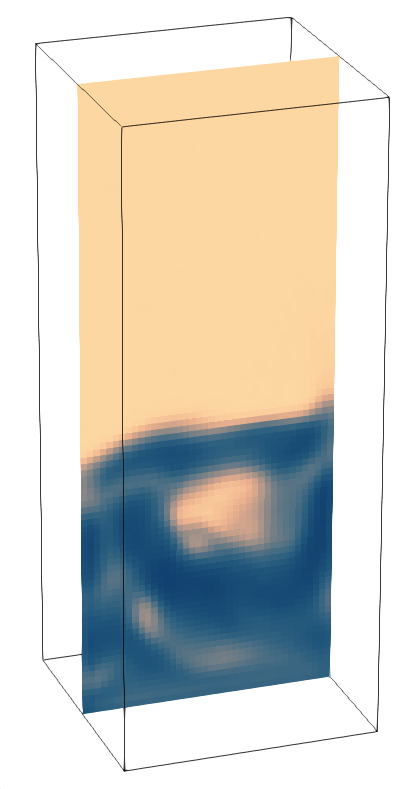} &
\includegraphics[width=.3\linewidth ,trim={0cm 0cm 0cm 0cm},clip]{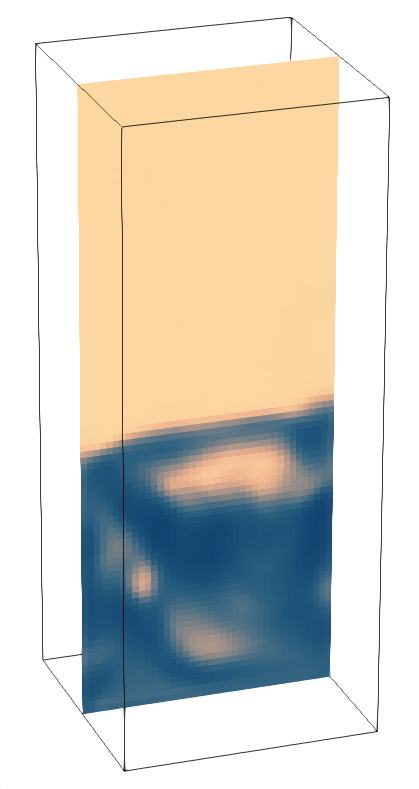} &
\includegraphics[width=.3\linewidth ,trim={0cm 0cm 0cm 0cm},clip]{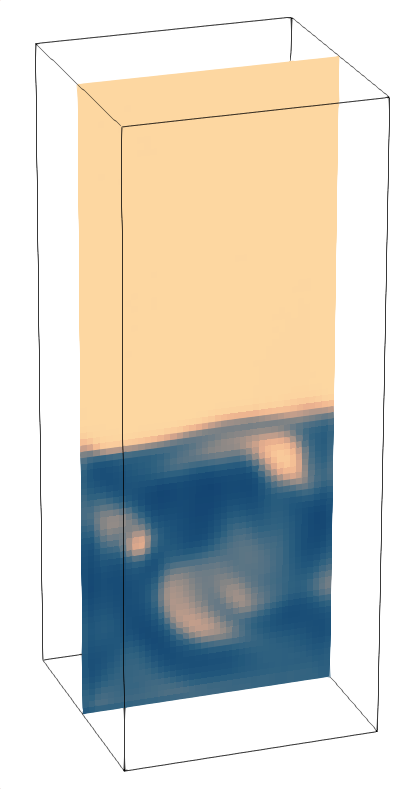}
 & \hspace*{0.1\linewidth}&
 \includegraphics[width=.3\linewidth ,trim={0cm 0cm 0cm 0cm},clip]{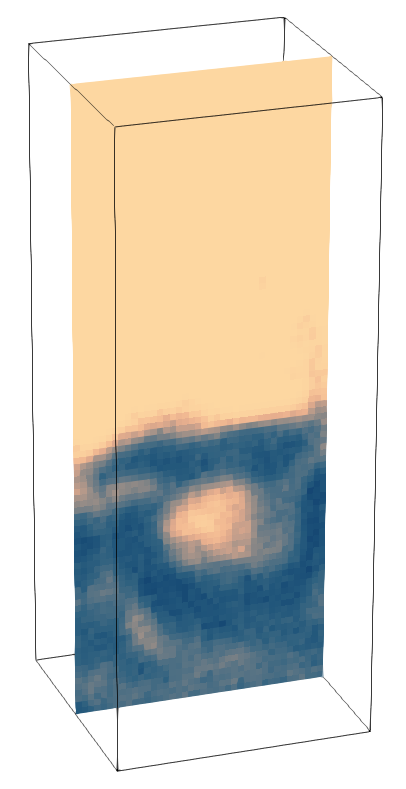} & 
\includegraphics[width=.3\linewidth ,trim={0cm 0cm 0cm 0cm},clip]{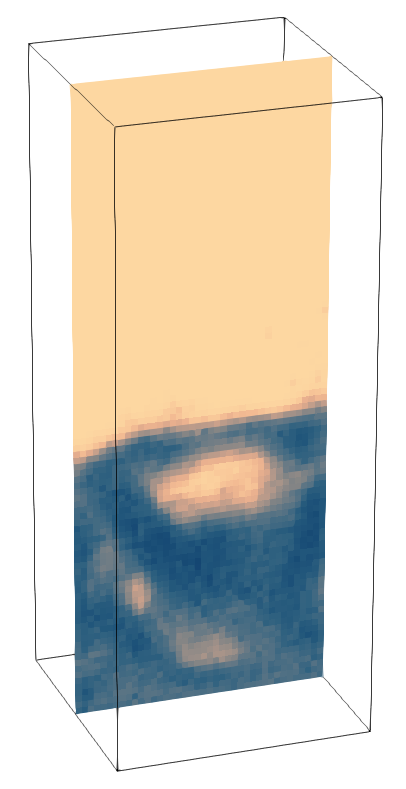} &
\includegraphics[width=.3\linewidth ,trim={0cm 0cm 0cm 0cm},clip]{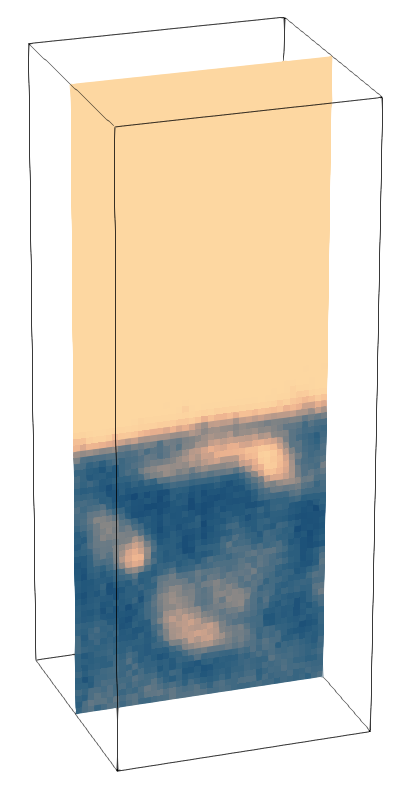}&
\includegraphics[width=.25\linewidth ,trim={0cm 0cm 0cm 0cm},clip]{images/FB/visual/col_void.png}
 \\
  \includegraphics[width=.3\linewidth ,trim={0cm 0cm 0cm 0cm},clip]{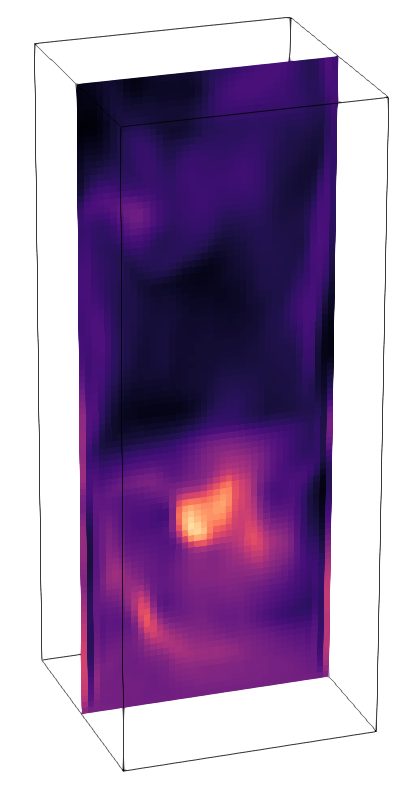} &
\includegraphics[width=.3\linewidth ,trim={0cm 0cm 0cm 0cm},clip]{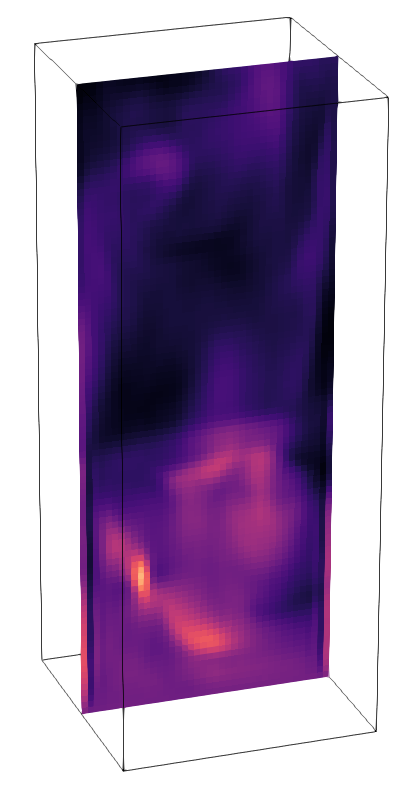} &
\includegraphics[width=.3\linewidth ,trim={0cm 0cm 0cm 0cm},clip]{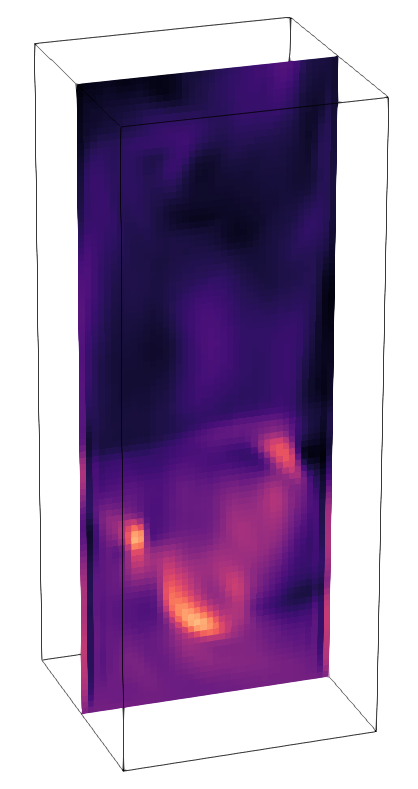}
 & \hspace*{0.1\linewidth}&
 \includegraphics[width=.3\linewidth ,trim={0cm 0cm 0cm 0cm},clip]{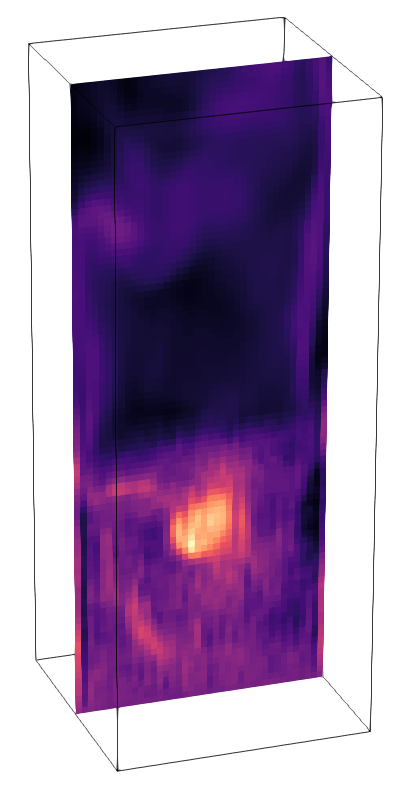} & 
\includegraphics[width=.3\linewidth ,trim={0cm 0cm 0cm 0cm},clip]{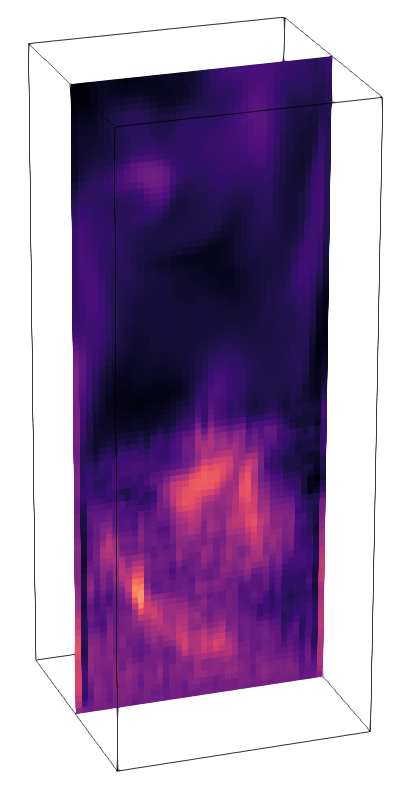} &
\includegraphics[width=.3\linewidth ,trim={0cm 0cm 0cm 0cm},clip]{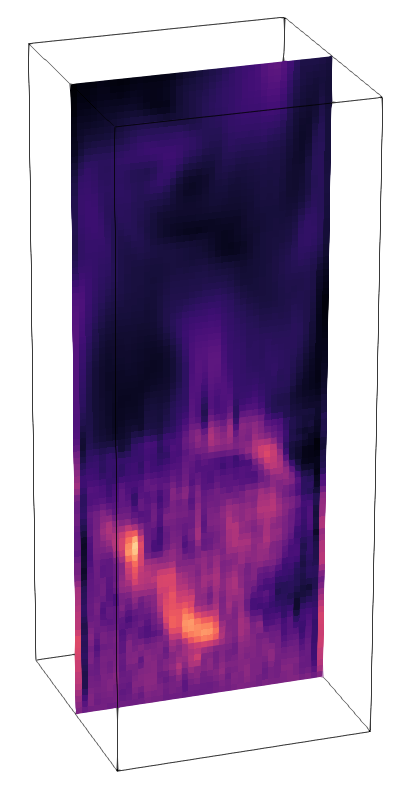}&
\includegraphics[width=.25\linewidth ,trim={0cm 0cm 0cm 0cm},clip]{images/FB/visual/col_vel.png}
 \\

 \end{tabular}
} 

\caption{High inlet velocity}
\end{subfigure}

    \caption{Visualization of three snapshots taken at $\SI{0.06}{\second}$, $\SI{0.09}{\second}$ and $\SI{0.12}{\second}$
    for two different inlet velocities (a) $\SI{0.45}{\meter\per\second}$ and (b) $\SI{0.7}{\meter\per\second}$.
    The first row shows iso-surfaces of solid fraction at $0.35$ uncovering the emerging bubble structure of fluidized bed reactors. The second and the third show the central slice along the $y$-axis for the solid fraction and the magnitude of fluid velocity, respectively. 
    Here, NeuralDEM uses the same initial conditions as the CFD-DEM simulation and faithfully reproduces the evolution of the system in the first few steps of the simulation. Please refer to \url{https://emmi-ai.github.io/NeuralDEM/} for videos of the rollouts.}
    \label{fig:fb_visualization}
\end{figure}
\subsubsection{Modeling the dynamics of the system}

\begin{figure}[htb!]
\centering
\begin{subfigure}[h]{0.46\linewidth}
\begin{tabular}{@{}m{0.24\linewidth}@{} @{}m{0.24\linewidth} m{0.24\linewidth}@{} @{}m{0.24\linewidth}@{}}
\scriptsize \parbox{\linewidth}{\centering CFD-DEM}& \scriptsize \parbox{\linewidth}{\centering NeuralDEM} & \scriptsize \parbox[c]{\linewidth}{\centering CFD-DEM} & \scriptsize \parbox[c]{\linewidth}{\centering NeuralDEM}\\
\includegraphics[width=0.95\linewidth ,trim={0cm 0cm 0cm 0cm},clip]{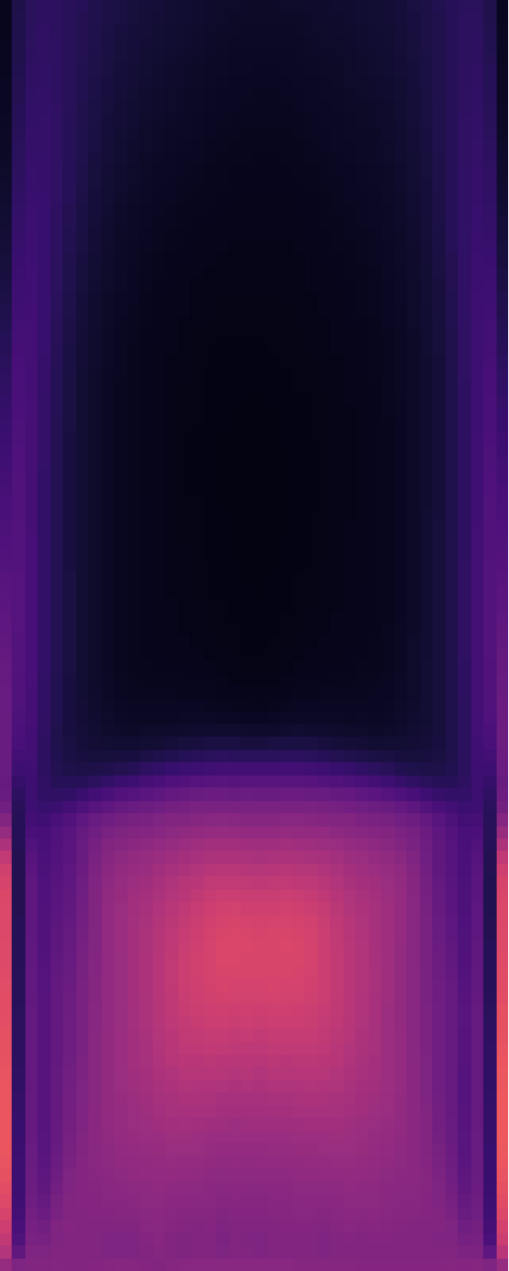}&
\includegraphics[width=0.95\linewidth ,trim={0cm 0cm 0cm 0cm},clip]{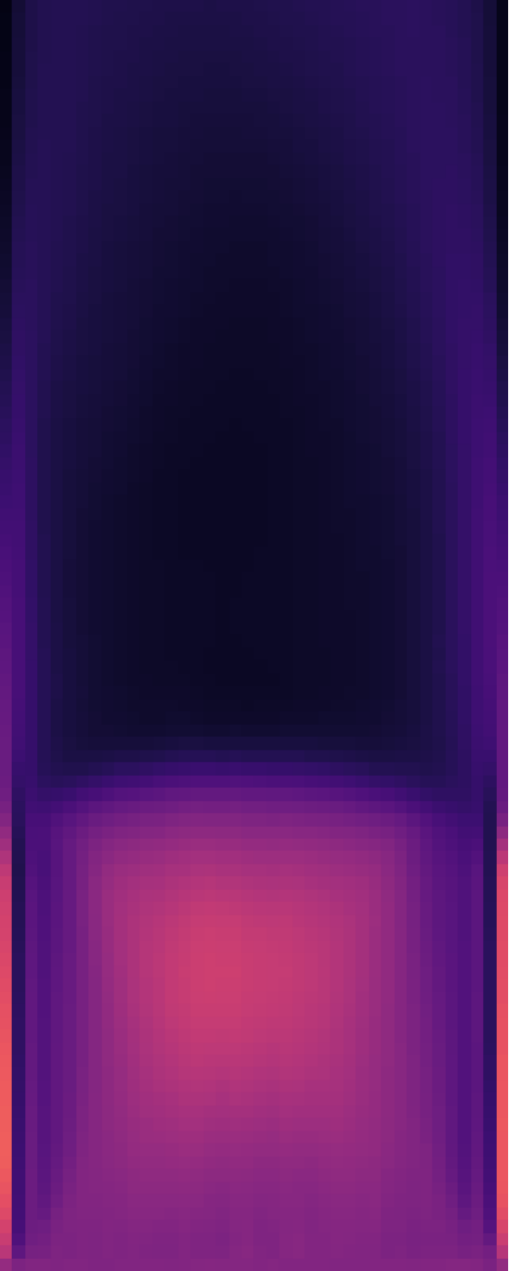}&
\includegraphics[width=0.95\linewidth ,trim={0cm 0cm 0cm 0cm},clip]{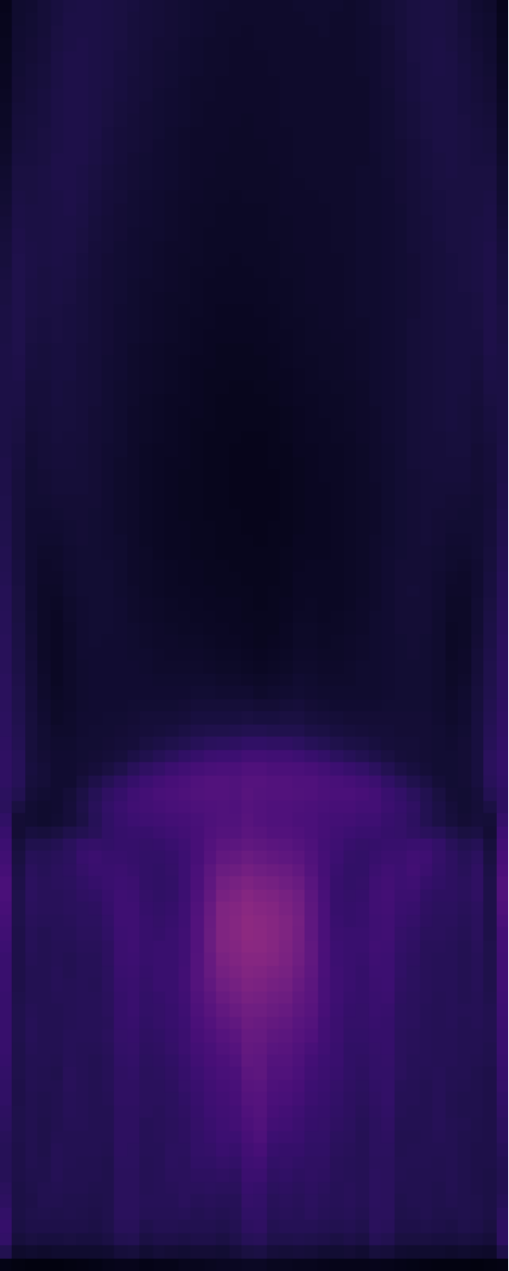}& 
\includegraphics[width=0.95\linewidth ,trim={0cm 0cm 0cm 0cm},clip]{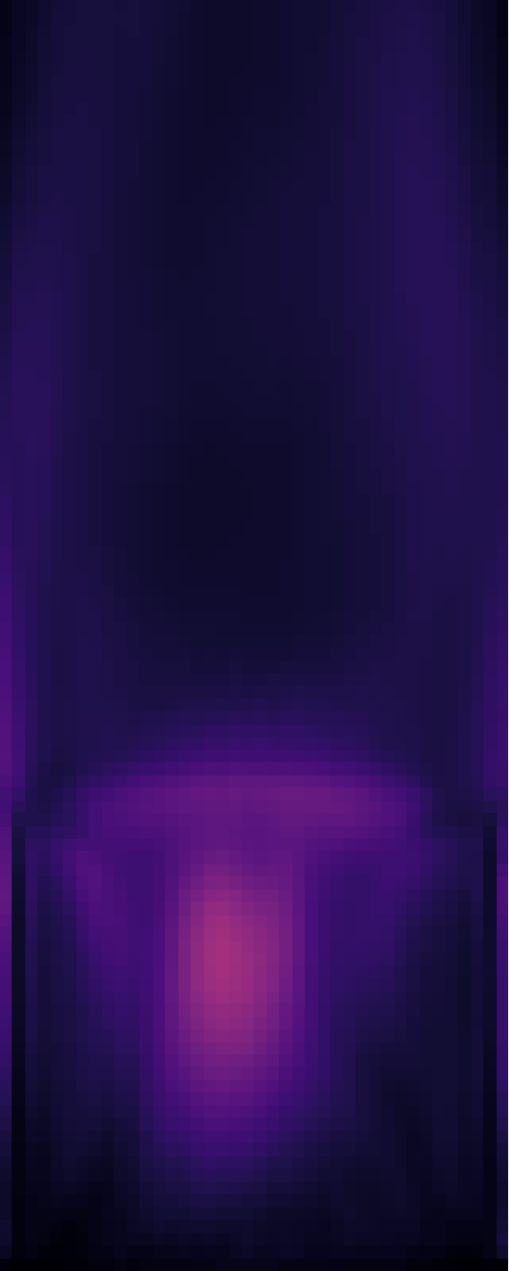}\\
\multicolumn{2}{c}{\centering \includegraphics[height=0.45\linewidth ,trim={2cm 2.5cm 5cm 2cm},clip,angle=-90]{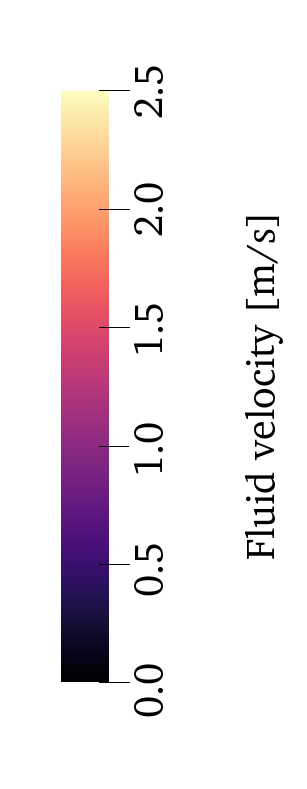}}&
\multicolumn{2}{c}{\centering \includegraphics[height=0.45\linewidth ,trim={2cm 2cm 5cm 2cm},clip,angle=-90]{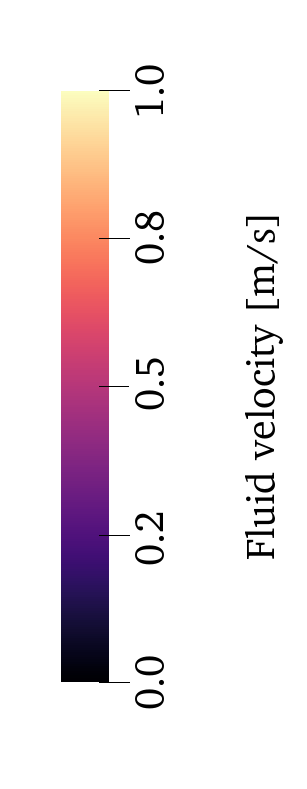}}\\
\multicolumn{2}{c}{\centering \scriptsize Mean fluid velocity}&\multicolumn{2}{c}{\centering \scriptsize Std dev. fluid velocity}\\
\vspace*{0.5cm}\\
\scriptsize \parbox{\linewidth}{\centering CFD-DEM}& \scriptsize \parbox{\linewidth}{\centering NeuralDEM} & \scriptsize \parbox[c]{\linewidth}{\centering CFD-DEM} & \scriptsize \parbox[c]{\linewidth}{\centering NeuralDEM}\\
\includegraphics[width=0.95\linewidth ,trim={0cm 0cm 0cm 0cm},clip]{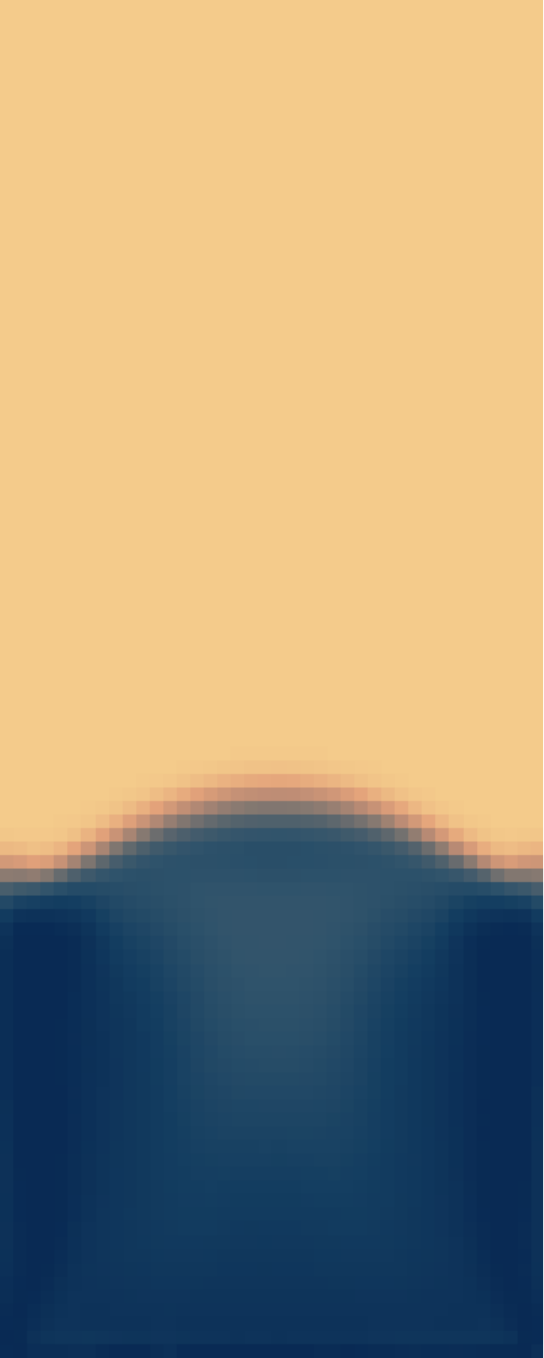} &
\includegraphics[width=0.95\linewidth ,trim={0cm 0cm 0cm 0cm},clip]{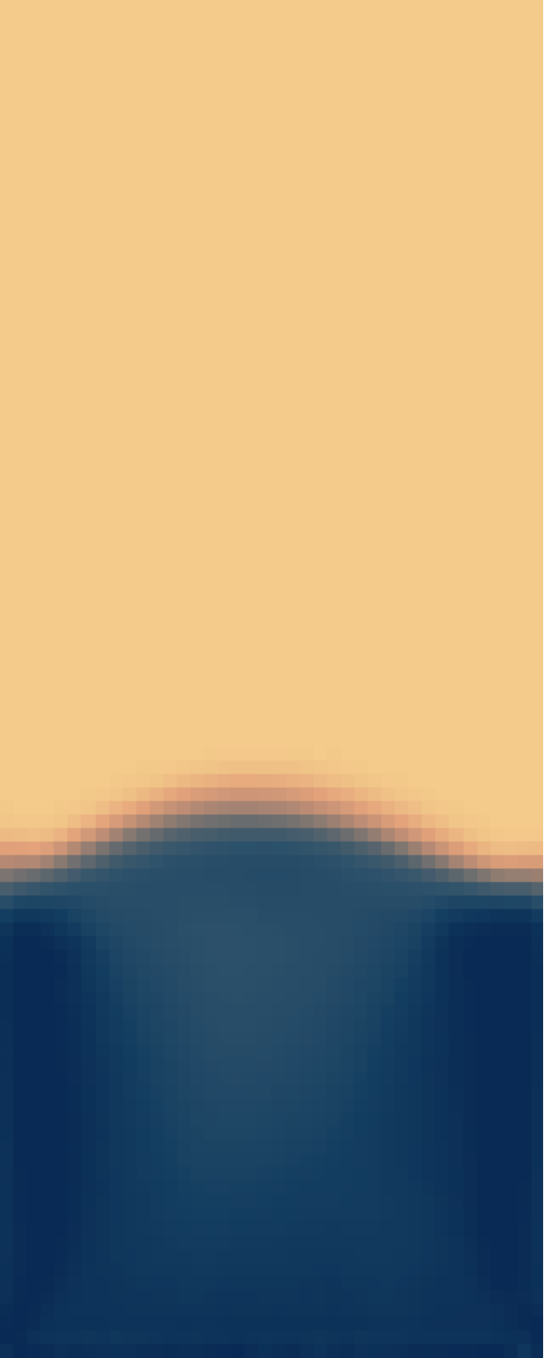}&
\includegraphics[width=0.95\linewidth ,trim={0cm 0cm 0cm 0cm},clip]{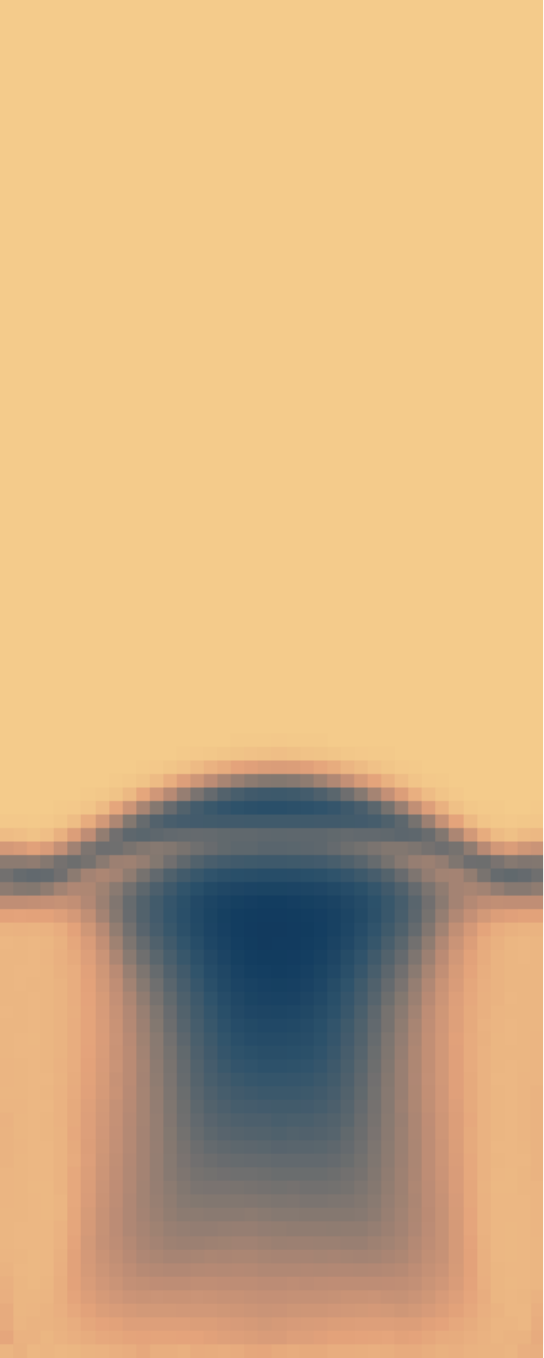} & 
\includegraphics[width=0.95\linewidth ,trim={0cm 0cm 0cm 0cm},clip]{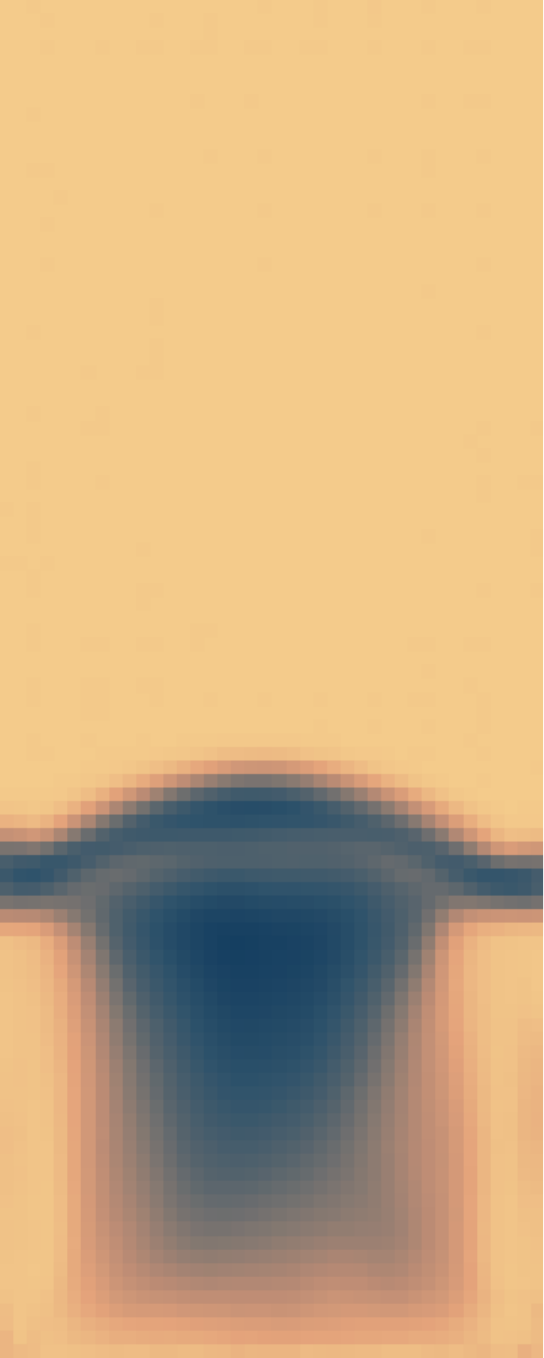}\\
\multicolumn{2}{c}{\centering \includegraphics[height=0.45\linewidth ,trim={2cm 2.5cm 5cm 0cm},clip,angle=-90]{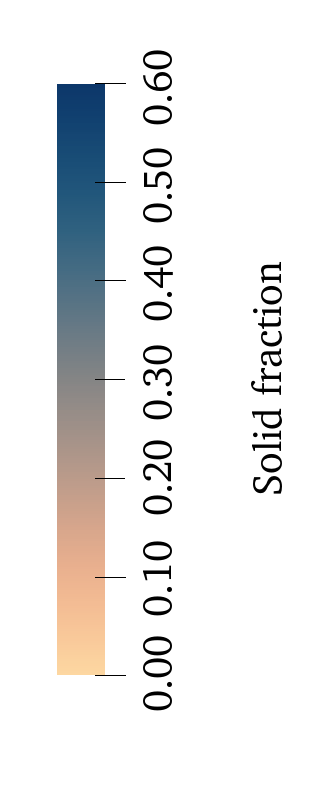}}&
\multicolumn{2}{c}{\centering \includegraphics[height=0.45\linewidth ,trim={2cm 2cm 5cm 0cm},clip,angle=-90]{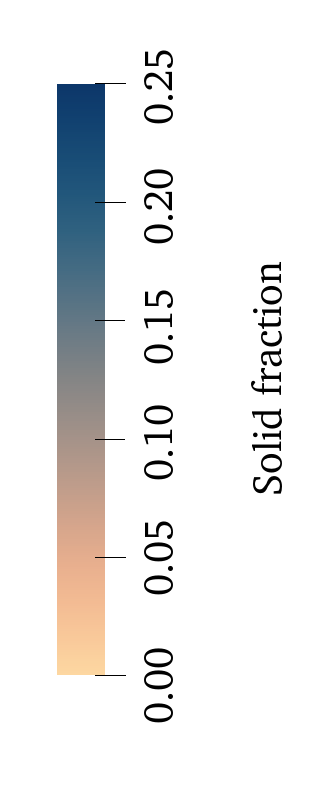}}\\
\multicolumn{2}{c}{\centering \scriptsize Mean solid fraction}&\multicolumn{2}{c}{\centering \scriptsize Std dev. solid fraction}\\
\end{tabular}
\caption{Low inlet velocity}
\end{subfigure}%
\hspace*{1.cm}
\begin{subfigure}[h]{0.46\linewidth}
\begin{tabular}{@{}m{0.24\linewidth}@{} @{}m{0.24\linewidth} m{0.24\linewidth}@{} @{}m{0.24\linewidth}@{}}
\scriptsize \parbox{\linewidth}{\centering CFD-DEM}& \scriptsize \parbox{\linewidth}{\centering NeuralDEM} & \scriptsize \parbox[c]{\linewidth}{\centering CFD-DEM} & \scriptsize \parbox[c]{\linewidth}{\centering NeuralDEM}\\
\includegraphics[width=0.95\linewidth ,trim={0cm 0cm 0cm 0cm},clip]{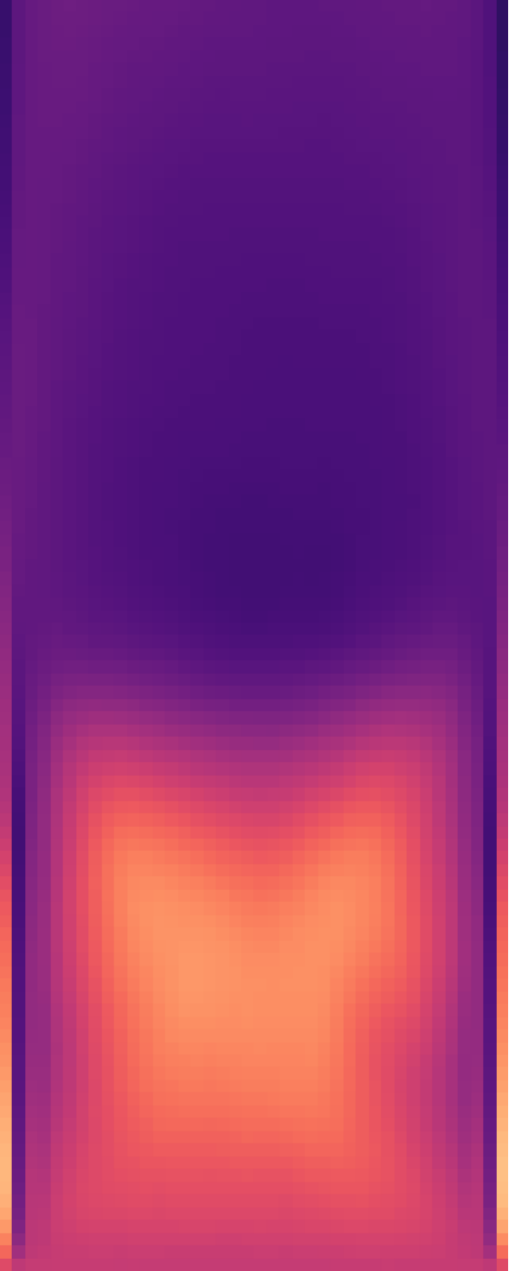} &
\includegraphics[width=0.95\linewidth ,trim={0cm 0cm 0cm 0cm},clip]{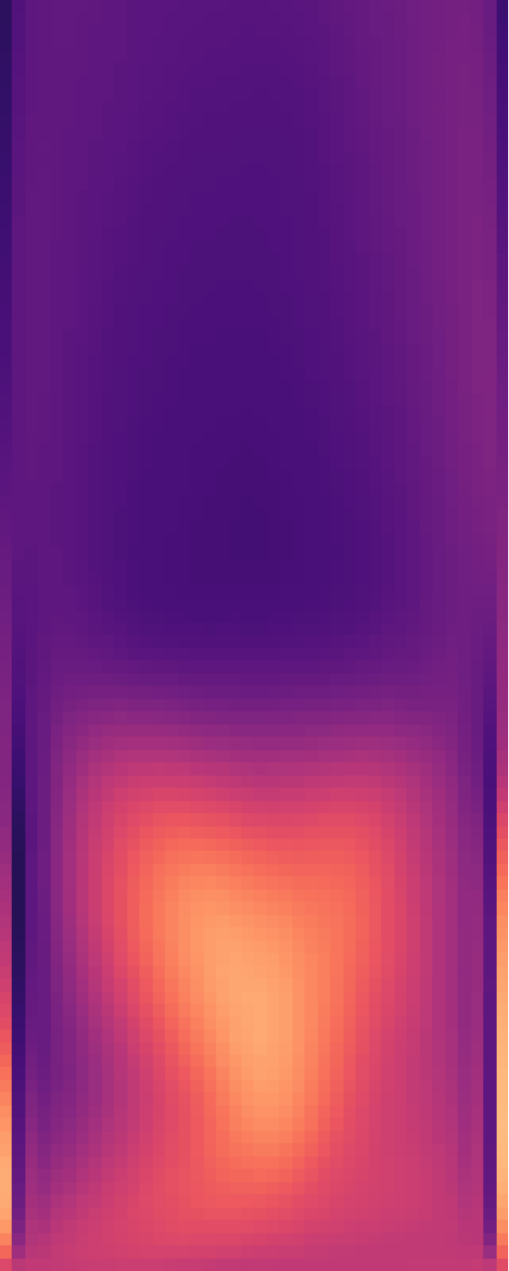}&
\includegraphics[width=0.95\linewidth ,trim={0cm 0cm 0cm 0cm},clip]{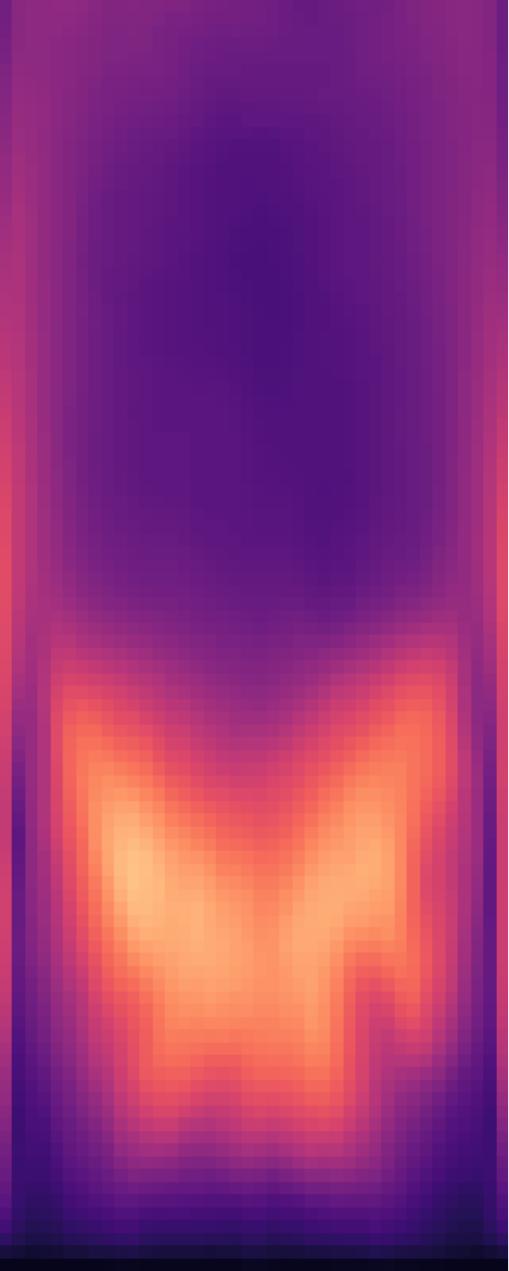} & 
\includegraphics[width=0.95\linewidth ,trim={0cm 0cm 0cm 0cm},clip]{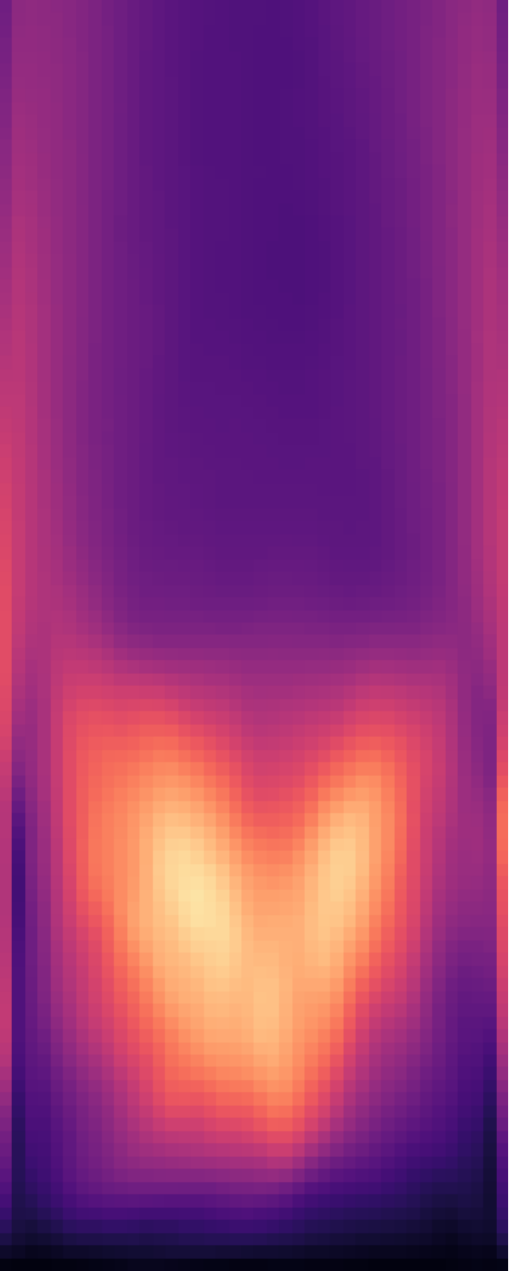}\\
\multicolumn{2}{c}{\centering \includegraphics[height=0.45\linewidth ,trim={2cm 2.5cm 5cm 0cm},clip,angle=-90]{images/FB/time_avg_statistics/colorbar/vel_25.png}}&
\multicolumn{2}{c}{\centering \includegraphics[height=0.45\linewidth ,trim={2cm 2cm 5cm 0cm},clip,angle=-90]{images/FB/time_avg_statistics/colorbar/vel_1.png}}\\
\multicolumn{2}{c}{\centering \scriptsize Mean fluid velocity}&\multicolumn{2}{c}{\centering \scriptsize Std dev. fluid velocity}\\
\vspace*{0.5cm}\\
\scriptsize \parbox{\linewidth}{\centering CFD-DEM}& \scriptsize \parbox{\linewidth}{\centering NeuralDEM} & \scriptsize \parbox[c]{\linewidth}{\centering CFD-DEM} & \scriptsize \parbox[c]{\linewidth}{\centering NeuralDEM}\\
\includegraphics[width=0.95\linewidth ,trim={0cm 0cm 0cm 0cm},clip]{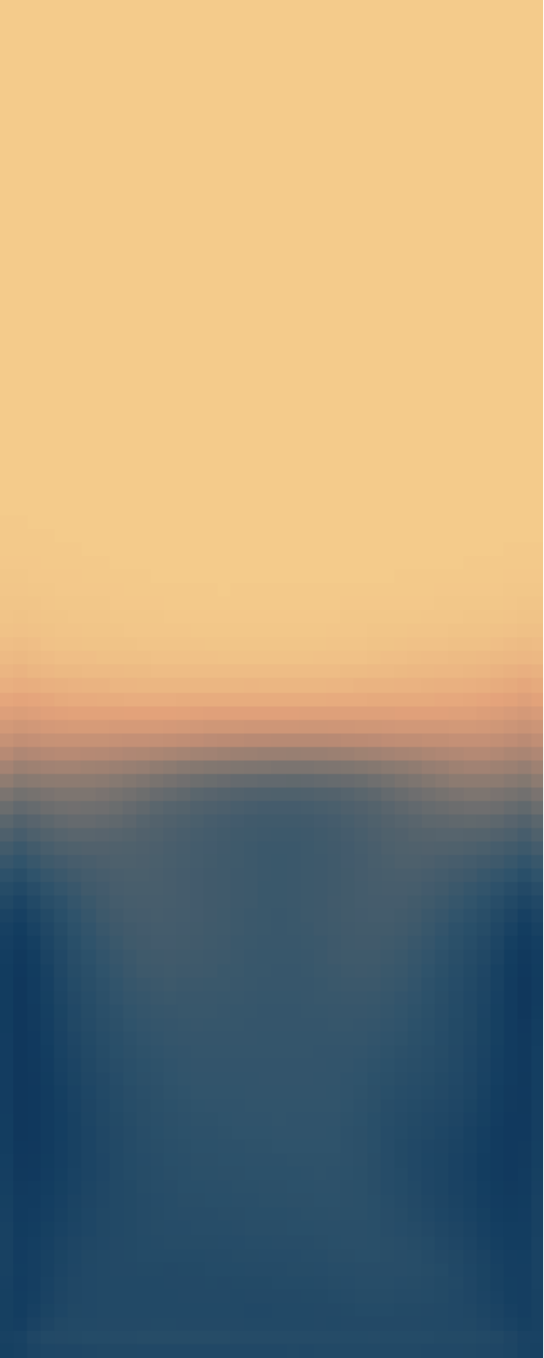} &
\includegraphics[width=0.95\linewidth ,trim={0cm 0cm 0cm 0cm},clip]{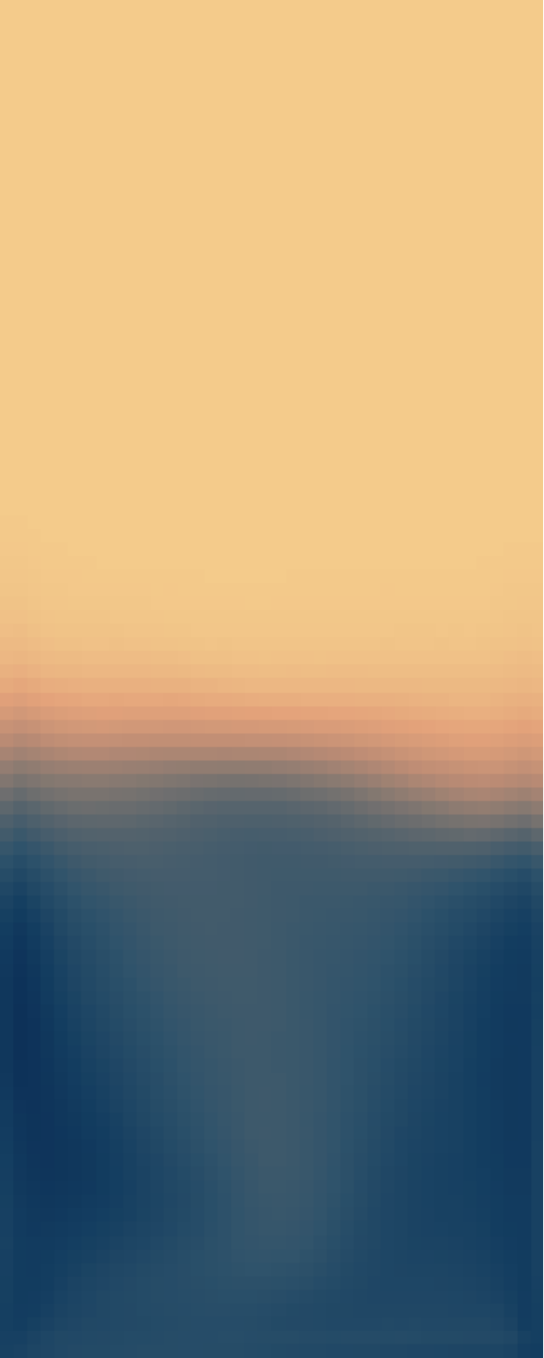}&
\includegraphics[width=0.95\linewidth ,trim={0cm 0cm 0cm 0cm},clip]{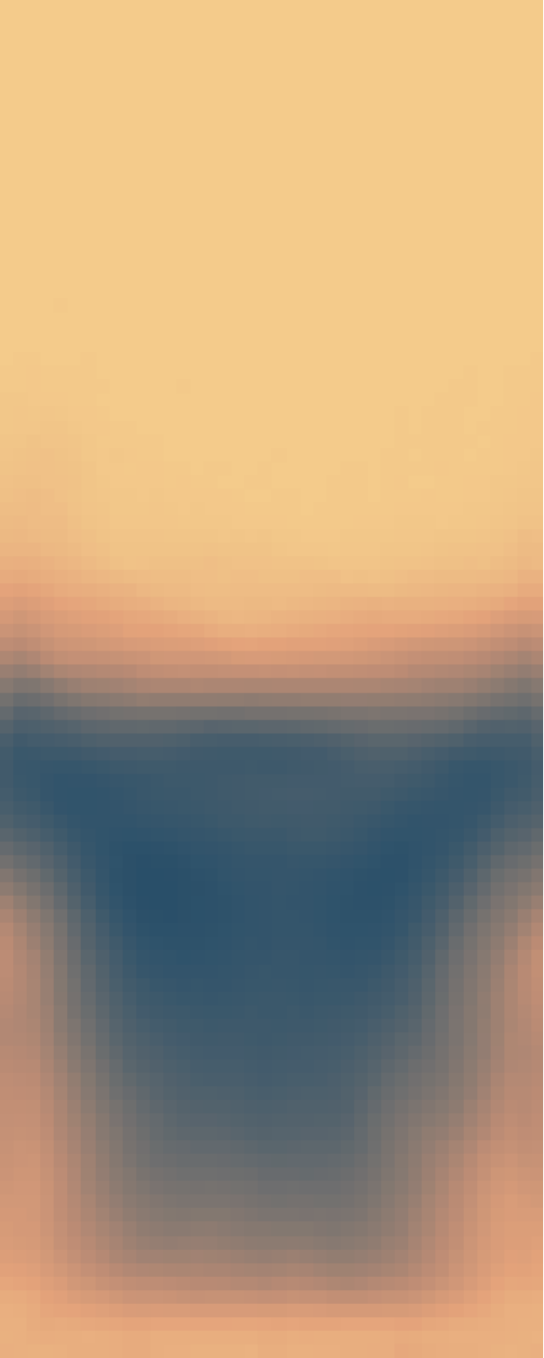} & 
\includegraphics[width=0.95\linewidth ,trim={0cm 0cm 0cm 0cm},clip]{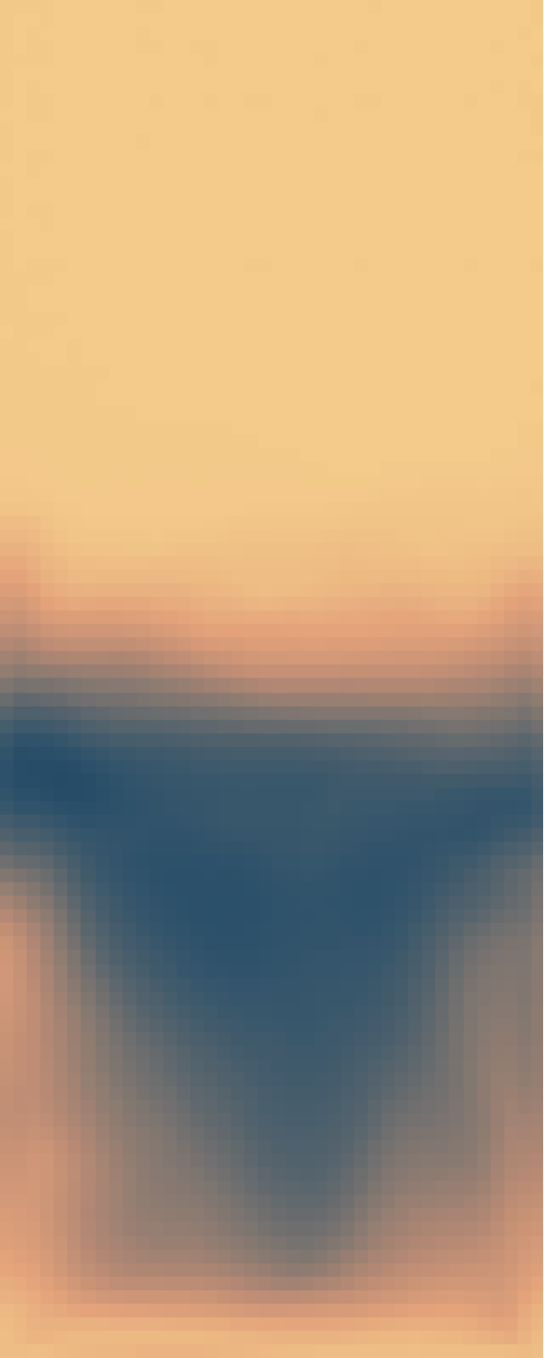}\\
\multicolumn{2}{c}{\centering \includegraphics[height=0.45\linewidth ,trim={2cm 2.5cm 5cm 0cm},clip,angle=-90]{images/FB/time_avg_statistics/colorbar/solid_0.6.png}}&
\multicolumn{2}{c}{\centering \includegraphics[height=0.45\linewidth ,trim={2cm 1cm 5cm 0cm},clip,angle=-90]{images/FB/time_avg_statistics/colorbar/solid_0.25.png}}\\
\multicolumn{2}{c}{\centering \scriptsize Mean solid fraction}&\multicolumn{2}{c}{\centering \scriptsize Std dev. solid fraction}\\
\end{tabular}
\caption{High inlet velocity}
\end{subfigure}%
\caption{Comparison of long term temporal averaging statistics for two fluid inlet velocities. Slices along the $y$-axis of the mean and the standard deviation of the magnitude of the velocity field and the solid fraction field shown. Each pairs shows the CFD-DEM simulation and the NeuralDEM model, respectively. NeuralDEM predictions closely match the long-term statistics of both slow and fast inlet velocity regimes, where spatial gradients vary from small to very high. 
}
    \label{fig:fb_visualization_stat}
\end{figure}

\begin{figure}[!htb]
    \centering
    \begin{subfigure}{0.48\textwidth}
        \centering
        \resizebox{\linewidth}{!}{
        \includegraphics[width=0.49\linewidth]{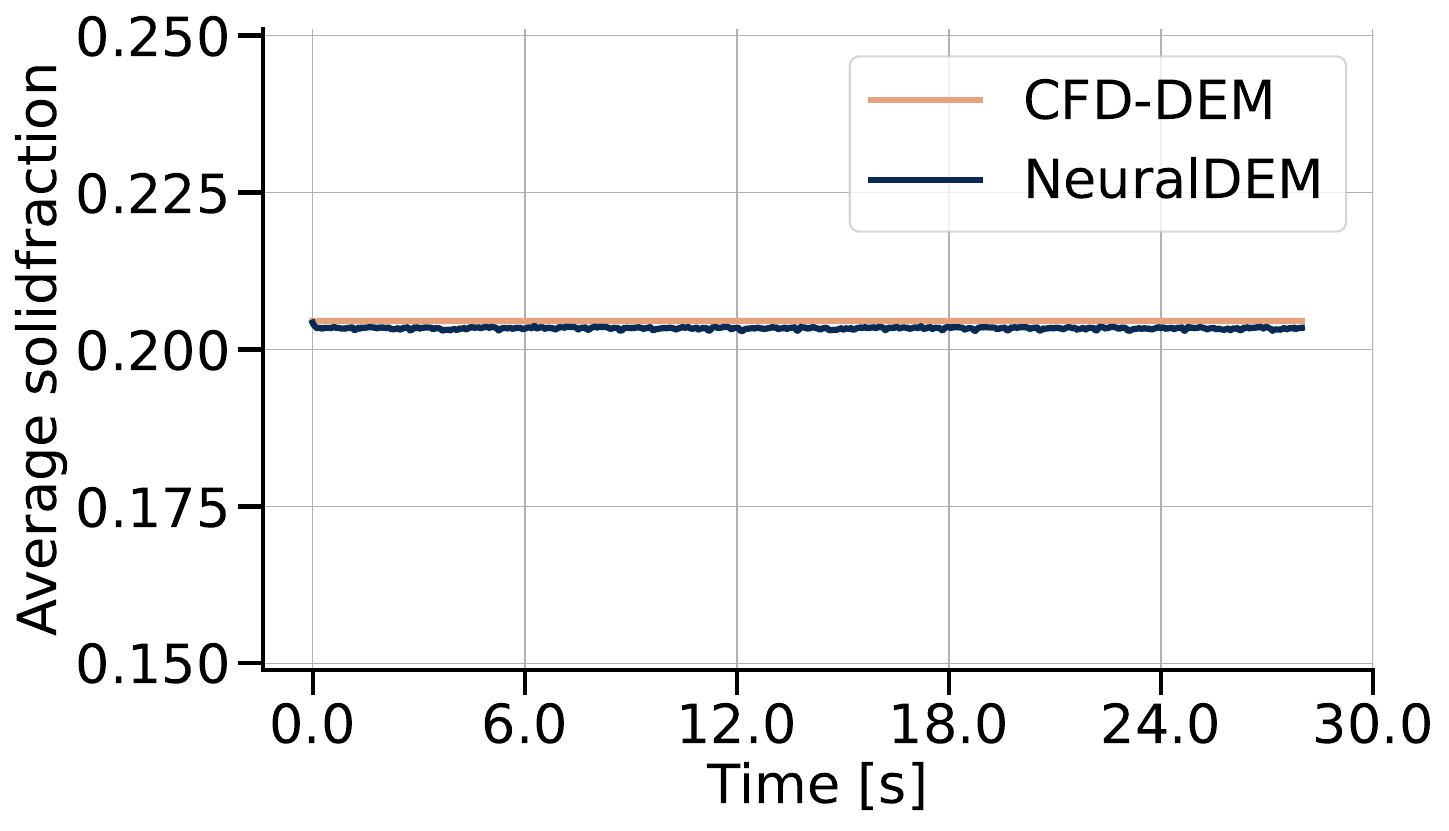}
        }
        \caption{Low inlet velocity}
    \end{subfigure}%
    \begin{subfigure}{0.48\textwidth}
        \centering
        \resizebox{\linewidth}{!}{
        \includegraphics[width=0.49\linewidth]{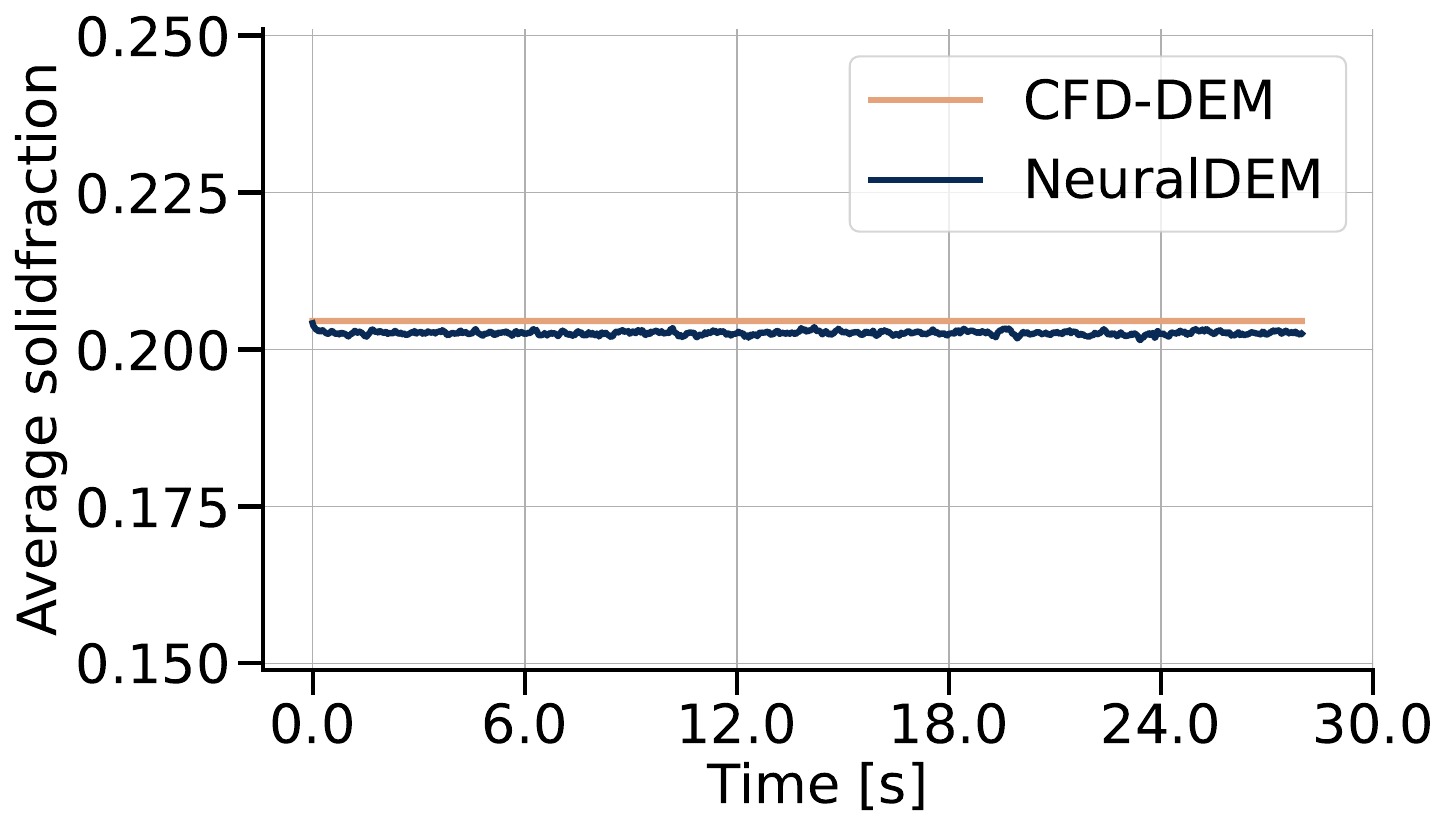}
        }
        \caption{High inlet velocity}
    \end{subfigure}
    \caption{The average solid fraction predicted by the NeuralDEM model is stable over extremely long rollouts that are almost $10\times$ the length of the $\SI{3}{\second}$ sequences used for training. Notably, although the output of NeuralDEM models a field, the mass of the entire system is preserved almost perfectly. The y-axis ranges from $0.15$ to $0.25$ to show more detail but the solid fraction can range from $0$ to $1$.}
    \label{fig:average_voidfraction}
\end{figure}

A fluidized bed reactor simulation exhibits fast and transient dynamics with many physically possible trajectories, i.e., fluidized bed reactor trajectories are chaotic. This means that -- also for numerical solvers -- starting a fluidized bed simulation with different initial particle packings will yield different trajectories. Still, in the limit of long simulation trajectories, different initial packings do not affect the temporal statistics. Due to this phenomenon, we use time-averaged statistics for quantitative comparisons, see Figure~\ref{fig:fb_visualization_stat} and Figure~\ref{fig:average_voidfraction}. Precise step-by-step comparison is not feasible due to numerical differences that naturally arise during a rollout. In Figure~\ref{fig:draining_hopper_transport_visualization}, we show a visual comparison of the iso-surface of solid fraction, solid fraction field, and fluid velocity for two different inlet velocities. 
When the two time series are compared to the ground truth snapshots they look very similar, especially the bubble structure visualized by the iso surface on the solid fraction field.

Notice how, with low inlet velocity, a single bubble of air forms inside the dense particle bed. When the bubble reaches the surface, a ripple-like structure is formed by the particles, clearly displaying their fluidization. With high inlet velocity, the behavior is much more unorganized and a complicated bubble structure inside the particle bed arises. NeuralDEM handles both regimes successfully, modeling the organized and structured low-velocity bubble dynamics as well as the chaotic, high-velocity particle interactions, accurately capturing the transitions in flow patterns and complex fluidization behavior across different regimes.

The fluid velocity within the particle bed shows distinct patterns influenced by particle motion and fluid-particle interactions. At low inlet velocity, the fluid flow is relatively stable, with modest spatial gradients around the rising air bubble. In contrast, with high inlet velocity, the fluid velocity across the bed is higher compared to the previous case, not only due to the high inlet velocity but also because of the dynamic movement of particles within the bed. These variations result in complex flow fields, with strong fluctuations that are challenging to model. NeuralDEM accurately captures these dynamics and replicates the different velocity profiles for all inlet velocity regimes between the scenarios shown here.


\subsubsection{Physics evaluation}
NeuralDEM's capability to maintain stability and accuracy over extended simulation periods is essential for realistic modeling of particle-fluid systems.
We demonstrate the long-term rollout stability by analyzing the time average solid fraction and fluid velocity field for 2.8k steps, representing $\SI{28}{\second}$ of physical simulation. 
The length of the trajectory resulted from a $\SI{30}{\second}$ CFD-DEM simulation, sufficient to determine the long-term statistics. NeuralDEM is not limited to this length and did not show any stability problem up to $\SI{100}{\second}$.
To the best of our knowledge, these are the longest reported stable rollouts in the deep learning-based 3D physical simulation field~\citep{lippe:2024,Kochkov:21,Sanchez:20}. 

In Figure~\ref{fig:fb_visualization_stat} we show the mean and variance of the magnitude of the velocity and the solid fraction over time for a given mesh cell. In the figure, we display the central slice along the $y$-axis (aka the depth of the reactor). NeuralDEM successfully captures the different long-term behavior of the particles and the fluid when different inlet velocities are used. 

Possibly the most crucial property of a physics simulation is mass conservation. In a CFD-DEM simulation, as long as the particles stay within the domain, the overall mass will not change. In Figure~\ref{fig:average_voidfraction} we show how NeuralDEM, although it does not model each particle independently, maintains very good mass conservation for all timesteps, showing no drift throughout the predicted trajectory.

\begin{wrapfigure}[21]{r}{0.35\linewidth}
\vspace{-3em}
        \centering
        \begin{subfigure}{0.5\linewidth}
            \centering
            \includegraphics[width=\linewidth]{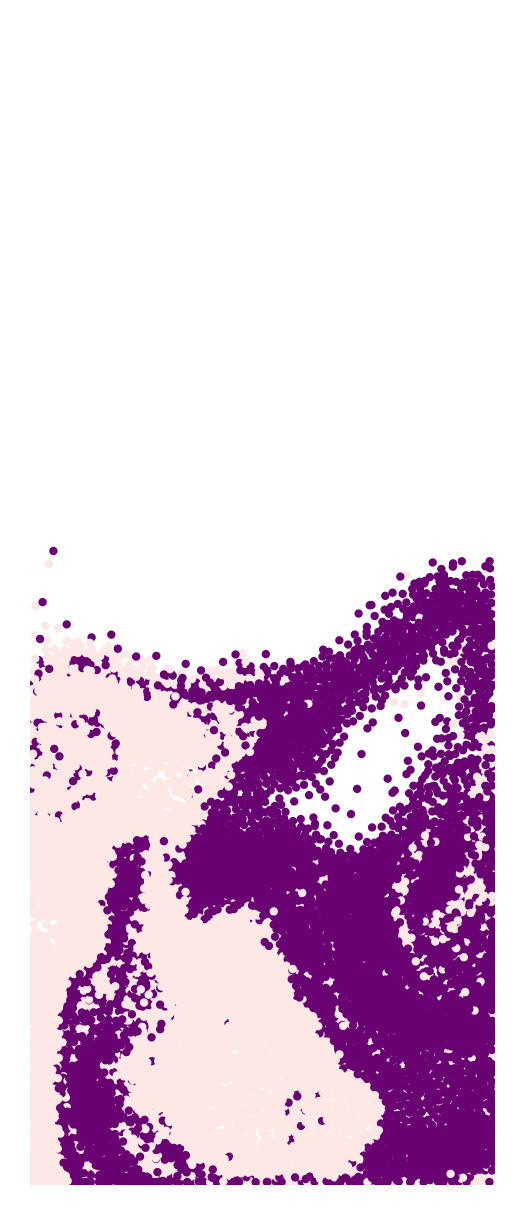}
            \caption{}
        \end{subfigure}%
        \begin{subfigure}{0.5\linewidth}
            \centering
            \includegraphics[width=\linewidth]{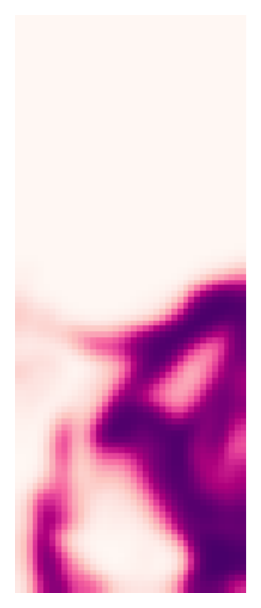}
            \caption{}
        \end{subfigure}
        \caption{(a) \textit{Particle-based mixing} where the dark particles started from the right half of the reactor. (b) \textit{Field-based concentration} obtained using a Gaussian kernel on a mesh.}
        \label{fig:mix_mapping}
\end{wrapfigure}

\subsubsection{Mixing behavior}

Particle mixing by definition is a particle-associated quantity where each particle either belongs to group A or B. In our numerical experiments, we label particles belonging to group A those that start in the left half of the reactor, and particles belonging to group B those that start in the right half. However, when using field-based representations, a faithful modeling of particle mixing is not feasible. Therefore, we introduce the particle mixing concentration field, which defines -- for a given spatial location and time -- the concentration of particles that belong to group B. Consequently, we discretize the domain using a hexahedron mesh and map the particle information using a Gaussian kernel on that mesh, as shown in Figure \ref{fig:mix_mapping}.

\begin{figure}[!htb]
        \begin{subfigure}{0.32\textwidth}
        \centering
        \includegraphics[width=\linewidth]{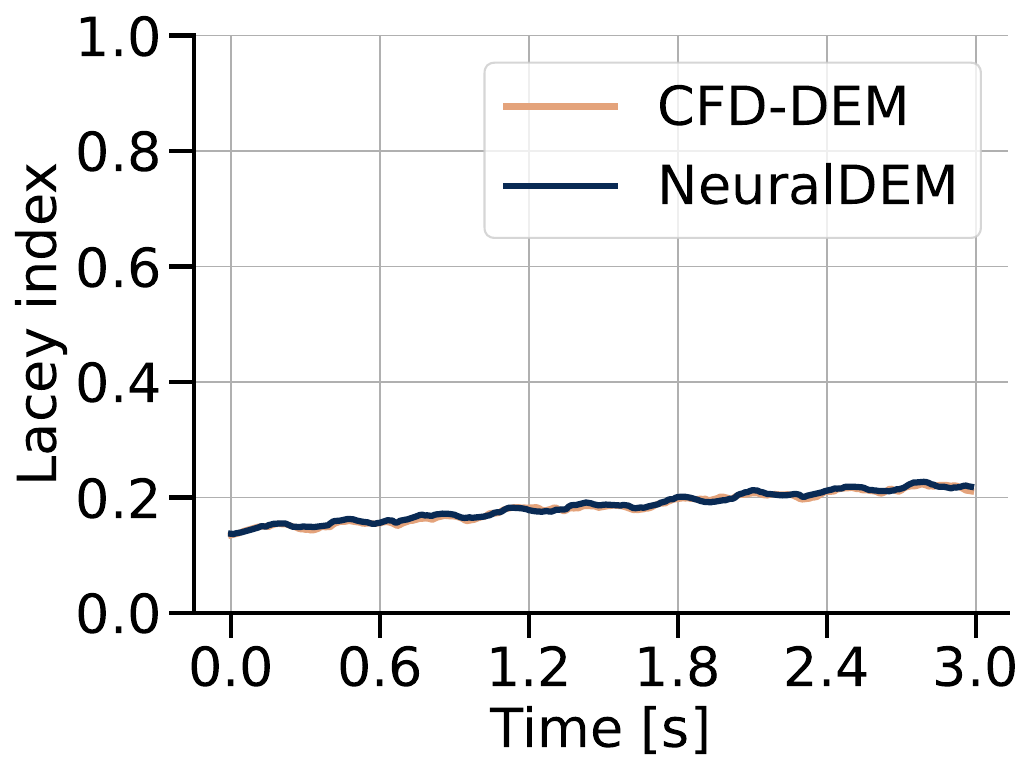}
        \caption{Low inlet velocity}
    \end{subfigure}%
    \begin{subfigure}{0.32\textwidth}
        \centering
        \includegraphics[width=\linewidth]{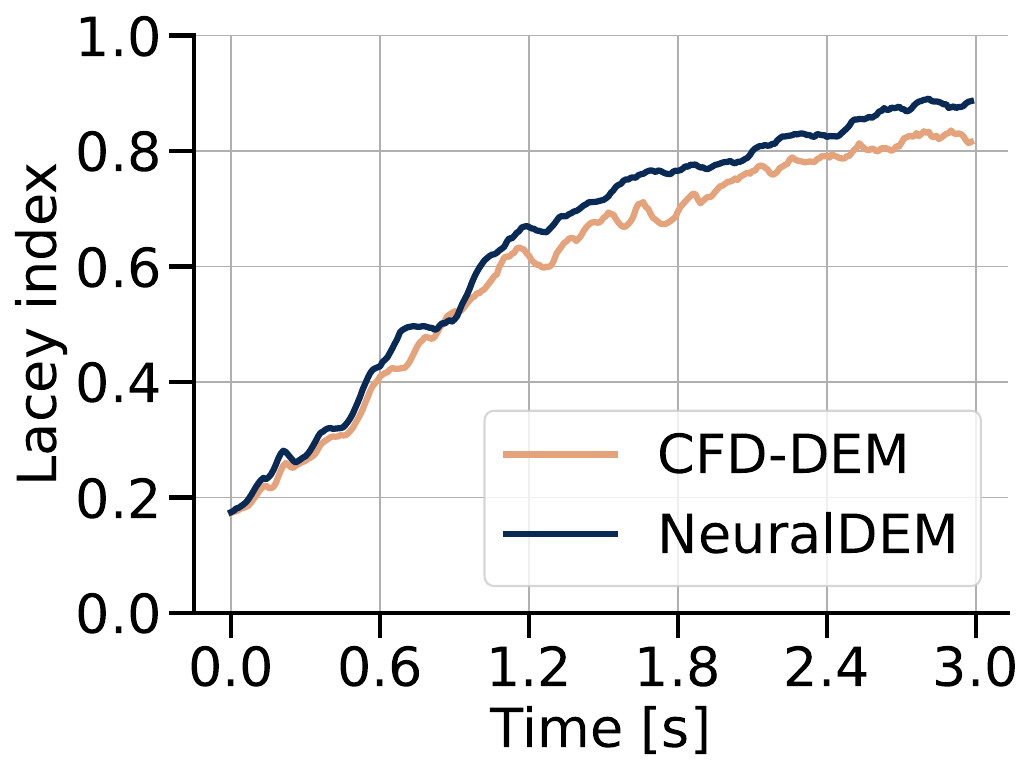}
        \caption{Medium inlet velocity}
    \end{subfigure}%
    \begin{subfigure}{0.32\textwidth}
        \centering
        \includegraphics[width=\linewidth]{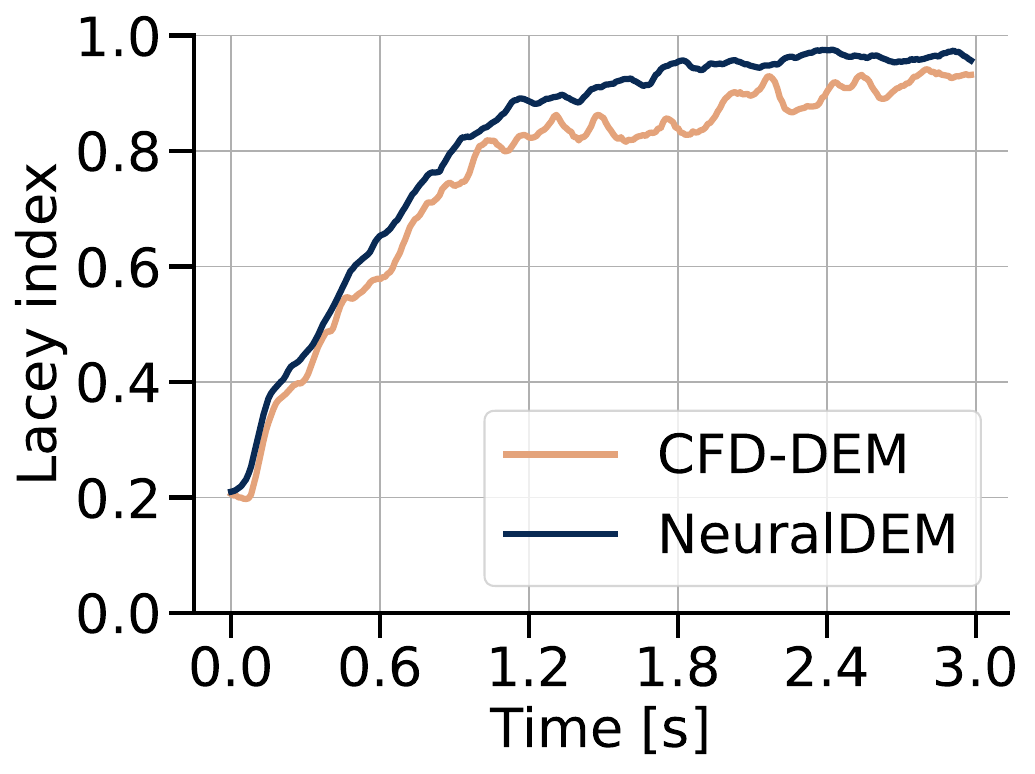}
        \caption{High inlet velocity}
    \end{subfigure}
    \caption{ 
    Comparison of the temporal evolution of the Lacey mixing index for three different inlet velocities: (a) $\SI{0.36}{\meter\per\second}$, (b) $\SI{0.55}{\meter\per\second}$, (c) $\SI{0.84}{\meter\per\second}$. NeuralDEM predictions (blue) match the characteristics of the ground truth DEM-CFD trajectories (orange) over long time horizons. In the low inlet velocity simulation, mixing is slow and the model can very accurately predict the mixing rate. For higher inlet velocities mixing is faster and the exact behavior looks more random because the behavior is more chaotic. However, the model correctly captures the exact trend in the mixing, starting fast and later slowing down.}
    \label{fig:mixing_temporal}
\end{figure}

We use the Lacey mixing index \cite{lacey1954mixingindex} to allow for a quantitative comparison between the model prediction for the mixing concentration field and ground truth simulation data. The Lacey mixing index represents the current state of mixing and can be plotted over time to compare the time evolution of the process. As the particles get more mixed, the Lacey mixing index goes towards $1$. As visualized in Figure~\ref{fig:mixing_temporal}, NeuralDEM predictions match the characteristics of the ground truth DEM-CFD trajectories over long time horizons. It is of special note how NeuralDEM can accurately predict the mixing rate for both slow-mixing systems and fast-mixing ones by simply conditioning on the appropriate inlet velocity.

\subsubsection{Runtime: NeuralDEM enables real-time simulations}

Similar to Section~\ref{sec:runtime_hopper}, we evaluate NeuralDEM's potential for real-time simulations. In contrast to the hopper simulations of Section~\ref{sec:hopper}, the fluidized bed reactors are larger and more complex, requiring roughly 20 times larger neural network architectures. Numerically, when using traditional CFD and DEM methods, the simulation of a fluidized bed reactor of 500k particles, with a trajectory spanning $\SI{3}{s}$ amounts to 12k CFD and 1.2M DEM timesteps. This requires 6 hours on 64 cores of high-performance CPUs. In contrast, on a single state-of-the-art GPU, the fastest NeuralDEM inference model faithfully reproduces the same physical behavior in just $ \SI{11}{\s}$. With further speedups via, e.g., model parallelization~\citep{shoeybi:2019megatron, xu:2021gspmd}, or model quantization~\citep{sung:2015resiliency,liu:2021post}, real-time inference is within reach.

\section{Discussion}

\subsection{Conclusion}
The present work has introduced NeuralDEM, a multi-branch neural operator framework that can learn the complex behavior of particulate systems over a wide range of dynamic regimes: from dense, pseudo-steady motion in hoppers to dilute, highly unsteady flow in fluidized bed reactors. 
NeuralDEM treats the Lagrangian discretization of DEM as an underlying continuous field while, simultaneously, modeling macroscopic behavior directly as additional auxiliary fields.
Long-term rollouts of our data-driven model show a high degree of stability and lead to accurate predictions regarding various target quantities such as residence times or mixing indices even for unseen conditions like different wall geometries, material properties, or boundary values. Evaluation times are several orders of magnitude faster than the underlying (CFD-)DEM simulations and demonstrate the real-time capability of NeuralDEM.
This remarkable speedup stems from NeuralDEM's field-based representation and the forward propagation thereof which is computationally much more efficient than the coupled solution of a huge number of real-space EOMs and lends itself to massive parallelization. As a side effect, NeuralDEM does not require the specification of microscopic DEM parameters. Instead, it can directly operate on macroscopic properties like angle of repose and shear cell characterization, which allows for a simple integration into engineering workflows.

\subsection{Existing modeling limitations}
A fundamental pillar of NeuralDEM is the new concept of modeling physics via field-based representations. This approach allows us to model macroscopic quantities extremely efficiently, thus scaling to large systems. However, the field-based representation comes with the limitation that properties attached to individual particles which cannot inherently be approximated spatially, are hard to model. To overcome this limitation, these quantities have to be replaced by an approximate field-based counterpart. A requirement for the demonstrated success of the autoregressive rollout is that the output distribution during rollout should match the input distribution observed during training. For example, particle-based quantities will be approximated with their field counterpart, which does not necessarily match the distribution of the input data, stifling the ability of the model to perform long-horizon predictions. Additionally, the autoregressive rollout is compute-intensive but could be improved using time propagation in the latent space, as introduced by~\citet{alkin2024upt}. Finally, NeuralDEM works best within the data distribution it was trained on. Therefore, NeuralDEM currently needs quite a lot of data to generalize across different geometries, operation parameters, and material properties.

\subsection{Future work}
While NeuralDEM clearly demonstrates the merits and massive potential of deep learning surrogates, we have applied our methodology only to test cases which are smaller and simpler than real industrial processes. For this reason, future work will address the issues of larger scales and more complex physics. With regard to the spatial scope, we will investigate if it is preferable to use data from detailed, fine-grained simulations in small domains and devise strategies for NeuralDEM to upscale this information, or if large-scale data from more approximate, coarse-grained simulations without subsequent upscaling steps lead to more reliable results. Concerning real multi-physics predictions, it is desirable to include heat transport and transfer as well as chemical reactions into NeuralDEM. Besides the conceptual challenge that a larger number of branches needs to interact with each other in the neural operator -- dynamics, heat transfer, and chemistry can be tightly coupled -- and give a detailed description of how grains of different temperatures and compositions mix and interact, we will also have to face more practical obstacles. Such systems often come with additional time scales because certain chemical reactions might be very slow compared to, e.g., the rapid particle and bubble motion in a fluidized bed. In these cases, data generation with conventional numerical methods could become unfeasible, and it might be necessary to include an intermediate, data-assisted step~\cite{Lichtenegger2017b} to first obtain long-term training data from short, high-fidelity time series.

\section{Acknowledgements}
We would like to sincerely thank Dennis Just, Miks Mikelsons, Robert Weber, Bastian Best, the whole NXAI team, and the whole Emmi AI team for ongoing help and support. We are grateful to Sepp Hochreiter, Phillip Lippe, Maurits Bleeker, Patrick Blies, Andreas Fürst, Andreas Mayr, Andreas Radler, Behrad Esgandari, Daniel Queteschiner for valuable inputs. Samuele Papa and Johannes Brandstetter thank Efstratios Gavves and Jan-Jakob Sonke for the help to make Samuele's research exchange smooth and mutual beneficial.

We acknowledge EuroHPC Joint Undertaking for awarding us access to Karolina at IT4Innovations, Czech Republic, MeluXina at LuxProvide, Luxembourg, LUMI at CSC, Finland and Leonardo at CINECA, Italy. The ELLIS Unit Linz, the LIT AI Lab, the Institute for Machine Learning, are supported by the Federal State Upper Austria. 

\bibliography{dem}
\bibliographystyle{dem_cit}

\end{document}